\documentclass[12pt,reqno]{article}

\usepackage[hmarginratio=1:1,margin=2.5cm,top=30mm,bottom=30mm]{geometry}
\usepackage{latexsym}
\usepackage{amscd}
\usepackage{amsmath,amssymb,bbm,bm}
\usepackage{amsfonts}
\usepackage{mathrsfs}
\usepackage{comment}
\usepackage{color,framed}
\usepackage{enumerate}
\usepackage{soul}
\usepackage{stmaryrd}
\usepackage{tikz-cd} 
\usepackage{tikz}
\usepackage{graphicx, caption, subcaption}
\usepackage{multirow}
\usepackage{adjustbox}
\usepackage[colorlinks]{hyperref}
\usepackage{url}
\usepackage{tcolorbox}
\usepackage{csvsimple}
\usepackage{algorithm}
\usepackage[noend]{algpseudocode}
\usepackage[space]{grffile}
\usepackage{marginnote}
\DeclareSymbolFont{sfletters}{OML}{cmbrm}{m}{it}
\DeclareMathSymbol{\svarphi}{\mathord}{sfletters}{"27}

\newcounter{mypbm}

\newcommand{\idop}{\textbf{id}}

\newcommand{\diff}{\,\text{d}}
\newcommand{\half}{{\frac 12}}
\newcommand{\Rd}{{\mathbb{R}^d}}
\newcommand{\transp}{^\top}

\newcommand{\inv}{^{-1}}

\newcommand{\pz}{\mathbf{p}_0}
\newcommand{\Pkj}{P^{k,j}}

\newcommand{\Ne}{{N_E}}
\newcommand{\nNe}{\frac{1}{\Ne}}
\newcommand{\elltwonorm}[1]{\| #1 \|_{Q}}
\newcommand{\gammainvnorm}[1]{\elltwonorm{#1}}
\newcommand{\kalmanp}{{K}_{\mathbf p}}

\newcommand{\cov}{\text{Cov}}

\newcommand {\ipRd}[2]{\langle #1,\, #2\rangle_{Q}}

\newcommand{\DiffV}{{\text{Diff}_{V}}(\Omega)}

\newcommand{\NNEdM}{\mathcal{N}(0, 1)^{d\times M}}
\newcommand{\Uniformdm}{U[-1, 1]^{d\times M}}

\newcommand{\RdM}{{\mathbb{R}^{d\times M}}}

\newcommand{\norm}[2]{\| #1 \|_{ #2 }}
\newcommand{\vnorm}[1]{\norm{ #1 }{V}}

\newcommand{\Vsf}{\textsf{V}}

\newcommand{\qzlm}{\mathbf{q}_{0}}
\newcommand{\qolm}{\mathbf{q}_{1}}
\newcommand{\dbfqt}{\dot{\mathbf{q}}_t}
\newcommand{\bfqt}{\mathbf{q}_t}
\newcommand{\bfpt}{\mathbf{p}_t}

\usepackage[citestyle=alphabetic,bibstyle=authortitle,backend=biber]{biblatex}
\begin{document}

\title{Learning landmark geodesics using Kalman ensembles}

\author{Andreas Bock \and Colin Cotter}

\maketitle

\begin{abstract}
We study the problem of diffeomorphometric geodesic landmark matching where the objective is to find a diffeomorphism that via its group action maps between two sets of landmarks. It is well-known that the motion of the landmarks, and thereby the diffeomorphism, can be encoded by an initial momentum leading to a formulation where the landmark matching problem can be solved as an optimisation problem over such momenta. The novelty of our work lies in the application of a derivative-free Bayesian inverse method for learning the optimal momentum encoding the diffeomorphic mapping between the template and the target. The method we apply is the ensemble Kalman filter, an extension of the Kalman filter to nonlinear observation operators. We describe an efficient implementation of the algorithm and show several numerical results for various target shapes.
\end{abstract}

\tableofcontents

\section{Introduction}

A central matching problem in shape analysis is to find a diffeomorphism that via its group action brings into alignment two sets of so-called \emph{landmarks} (point clouds in our domain $\Omega:=\Rd$) called the \emph{template} and the \emph{target} - a problem in line with the metric pattern theory \cite{mumford2002pattern} framework of Grenander \cite{grenander1994representations}. For applications in computational anatomy \cite{grenander1998computational} it is often convenient to choose diffeomorphometric methods for matching since this class provides transformations that can represent smooth biological growth. A popular method is the \emph{large deformation diffeomorphic metric mapping} (LDDMM) \cite{younesshapes}. In LDDMM we study a curve $t\mapsto \varphi_t \in \DiffV$, $t\in [0,1]$ on the group of diffeomorphisms $\DiffV$ induced by a curve $t\mapsto u_t$ on the smooth vector space $V\hookrightarrow\textsf{C}_0^1(\Omega, \Rd)$ (the space of continuously differentiable functions on $\Omega$ vanishing at infinity and taking values in $\Rd$) via the following equation, see e.g. \cite{dupuis1998variational}:
\begin{equation}\label{diffeo}
\dot{\varphi_t} = u_t \circ \varphi_t\, , \qquad  \varphi_0 = \idop.
\end{equation}
Letting $\varphi_1 . q$ denote the group action of $\varphi_1$ on $q$, the LDDMM matching problem can be solved by finding the one-parameter family of velocity fields $t\mapsto u_t \in V$ such that the distance between a target $q_1$ and $\varphi_1 . q_0$ is minimised. The candidate curve of velocities in $V$ is the one whose kinetic energy $\int_0^1\|u_t\|^2_V \diff t$ is minimised subject to \eqref{diffeo}. See \cite{trouve1995infinite,trouve1998diffeomorphisms,trouve2005metamorphoses} for the technical development of the family of methods using this approach and \cite{holm2009euler} for an extension of LDDMM called metamorphosis.
A particular strength of this framework is that the geodesic motion of the shape can be encoded by an initial momentum conjugate to the template. This means that the full curve $\varphi$ can be described by a single initial momentum rather than a curve of velocities in $V$. The technical details are described in section \ref{sec:lddmm}. LDDMM can therefore be viewed as an \emph{inverse problem} in the sense that we want to find the momentum parameter leading to the observation given by the evolution of the template shape at time $t=0$ to fit the target shape at time $t=1$. A
popular method for solving such inverse problems is called
\emph{shooting} and we highlight here some central references explaining this
approach. Shooting typically employs a Newton method to explore the space of
momentum provided some initial guess; the gradient-based optimisation scheme of Beg \cite{beg2005computing}
popularised this for LDDMM using the so-called EPDiff equation \cite{holm1998euler}, see also
\cite{younes2007jacobi,younes2009evolutions,vialard2012diffeomorphic,miller2006geodesic}. See \cite{kuhnel2017differential} for an introduction to implementation landmark geodesic equations.\\

In this paper we build a surrogate Bayesian model on the tangent space
of the landmark manifold and treat the landmark matching problem as a Bayesian
inverse problem to arrive at a derivative-free matching algorithm.
An inverse problem in its most abstract form seeks to recover the input
parameter $p\in P$ that is mapped to a known state $q\in Q$ by some typically
known \emph{observation operator} $G: P\rightarrow Q$ e.g. via the solution of a
differential equation such as the geodesic equations for landmarks. 
When $G$ is the forward operator mapping from initial momentum $p$ to a candidate target $G(p)$ we can also view $G(p) = q_1$ as a \emph{Bayesian inverse problem} \cite{stuart2010inverse,dashti2017bayesian} where the aim is to find a
\emph{distribution} of candidate functions $p$ that minimise the kinetic
energy of a transformation
\cite{cotter2013bayesian,ma2008bayesian,ma2010bayesian}.
This means inverting, in some sense, $G$ so that $p = G\inv(q)$. In practice, we
are often simply looking for an approximation: $p \approx G\inv(q)$, and the
inverse may not be unique or classically defined. We can view shape matching
as the inverse problem of finding the velocity $p$ such that an initial shape
$q_0$ is mapped to $q_1\approx G(p)$ where $G$ is the forward geodesic motion of the template landmarks. In a crude sense we wish to bound $\|q_1 - G(p)\|$ by the following triangle inequality:
\begin{align*}
\| q_1 - G(p)\|_Q
& \leq \| q_1 - G_h(p)\|_Q + \| G_h(p) - G(p)\|_Q\\
& \leq \| q_1 - G_h(p)\|_Q + \| G_h - G\| \|p\|_P.
\end{align*}
where $G_h$ is a numerical approximation of $G$ and $\|\cdot\|$ is a formal dual
norm. In the present context we control $\| G_h - G\|$ by numerical discretisation of the landmark geodesics. The aim is to
\emph{learn} the optimal momentum $p$ encoding the forward operator that allows
us to take $\| q_1 - G_h(p)\|_Q$ to zero, and we do so by applying a nonlinear filtering method called the ensemble Kalman filter (enKF). As for geodesic shooting  we exploit the linearity of
the space of momentum to define an iterative Bayesian method. In this setting, a
collection - or \emph{ensemble} - of initial momentum is drawn from a proposed
prior distribution and is iteratively updated by measuring its likelihood as a
function of how close the \emph{average} template landmarks are from the target
under the flow of the diffeomorphism generated by the initial momentum. 
At the enKF level the algorithm is \emph{embarassingly parallel} in the ensemble
dimension and we present several numerical results using a parallel implementation.
The enKF algorithm is also entirely derivative-free, paving the way for
researchers to use other forward operators instead of those for LDDMM that are
used in this paper.



\subsection{Outline}

A mathematical treatment of LDDMM using a reproducing kernel Hilbert space framework is provided in section \ref{sec:lddmm} along with derivations of Hamilton's equations for landmarks. Section \ref{sec:enkf} describes the enKF in detail and its application to landmark matching. Next, section \ref{numerical_results} shows several numerical examples for various settings and robustness of the algorithm shows clear promise of our approach. Section \ref{sec:conclusion} summarises this paper.

\section{Large Deformation Diffeomorphic Metric Mapping}\label{sec:lddmm}

To set up the notation we first present some preliminaries in section  \ref{sec:math} before describing the classic LDDMM framework in section \ref{sec:geodesics}.

\subsection{Mathematical Preliminaries}\label{sec:math}

In this paper, $\|\cdot\|_\Rd$ denotes the standard Euclidean norm and we let $\langle h,g\rangle_Q:= \langle h,g\rangle_\RdM:= \sum_{i=1}^M \langle h^i, g^i\rangle_\Rd$ for $h,g\in\RdM$. Futher, let $\langle \cdot, \cdot\rangle_{0^d, \Omega}$ be the $L^2(\Omega)$ norm of vector $d$-valued functions defined over $\Omega$.\\

We use $\mathbf q_t \in Q:=\RdM$ to denote a vector of $M$ landmarks $\mathbf q_t =\{ q_t^i \}_{i=1}^M$ at time $t\in [0,1]$, with $q_t^i\in\Rd$ for $d=2$. When the time index is omitted $\mathbf q$ refers to the one-parameter family of landmark positions and we say $\mathbf{q}\in \mathbf Q:=\{ \mathbf q \,|\, t\mapsto \mathbf q_t \in Q,\, t\in [0,1]\}$. $V \hookrightarrow \mathsf{C}_0^2(\Omega)^d$ denotes a vector space to be specified later on, and $\Vsf=L^2([0,1], V)$ the space of square-integrable curves taking values in the sufficiently smooth space $V$ which we define later. Further, when $\varphi_t\in\DiffV$ is a diffeomorphism we understand the action of $\varphi_t$ on $\mathbf{q}_t\in Q$ as:
\[
\varphi_t . \mathbf{q}_t = \{ \varphi_t . q^i_t \}_{i=1}^M,
\]
where $\varphi_t . q^i = q^i \circ \varphi_t\inv$. For brevity we shall use the notation:
\[
\dbfqt = u_t \circ \bfqt
\]
to describe the evolution of the collection of landmarks whereby each landmark
$i=1,\ldots M$ is governed by:
\[
\dot{q}^i_t = u_t \circ q^i_t.
\]
In this paper the space $V$ is a reproducing kernel Hilbert space \cite[Chapter 9]{younesshapes} with kernel $K_V:\Omega\times\Omega\rightarrow\mathbb{R}$. We denote by $L_V: V\rightarrow L^2(\Omega,\Rd)$ the symmetric operator generating $V$ which for all $u \in V$ satisfies the following:
\begin{align*}
& \|u\|_V^2 = \langle L_V u, u\rangle_{0^d, \Omega},\\
& \langle u, u\rangle_{0^d, \Omega}\leq  C\langle L_V u, u\rangle_{0^d, \Omega},\quad \text{for some constant}\quad C>0,
\end{align*}
with an associated inverse described by $K_V$. We assume that the kernel $K_V$ is Gaussian:
\begin{equation}\label{KVdef}
K_V(x, y) = e^{-\frac{\|x - y\|_\Rd^2}{2\tau^2}},
\end{equation}
where $\tau>0$ is a kernel parameter that determines the interaction of the
landmarks which we refer to as the \emph{size} of the landmarks. This is easily seen: as $\tau\rightarrow 0$, $x$ and $y$ may be closer in the plane before $K_V(x, y)$ takes values away from zero.

\subsection{Geodesics}\label{sec:geodesics}

The LDDMM matching problem between two configurations of landmarks $\qzlm$ and
$\qolm$ seeks to minimise the following functional as a function of $u \in \Vsf$:
\begin{equation}\label{fnl:lddmm}
    \half\int_0^1 \vnorm{u_t}^2 \diff{t},
\end{equation}
subject to the evolution equation $\dbfqt = u_t \circ \bfqt$ and the boundary conditions $\qzlm$. We address the end-point condition momentarily. Letting $\mathbf p$ denote the conjugate momentum to $\mathbf q$ we can write the Lagrangian associated to \eqref{fnl:lddmm}:
\[
\mathfrak{L} = \int_0^1(\bfpt\,|\, \dbfqt - u_t \circ \bfqt)_{Q,Q^*} -
\half\vnorm{u_t}^2\diff{t}.
\]
Differentiating $\mathfrak L$ with respect to arbitrary variations $\delta u$, $\delta \mathbf p$ and $\delta \mathbf q$ in $u$, $\mathbf p$ and $\mathbf q$, respectively, gives us the equations:
\begin{subequations}
\begin{align}
& \langle L_V u_t, \delta u\rangle_{0^d,\Omega} = (\bfpt\,|\,\delta u \circ
\bfqt)_{Q,Q^*},& \quad\forall v \in V,\\
& ( \mathbf p\,|\, \dot{\delta\mathbf q} - \nabla u_t \circ \mathbf q \delta \mathbf q)_{Q,Q^*} = 0,& \quad\forall \delta \mathbf q \in \mathbf Q,\label{pmomentum}\\
& (\delta \mathbf p\,|\, \dbfqt - u_t \circ \bfqt)_{Q,Q^*} = 0,& \quad\forall \delta \mathbf p \in \mathbf Q^*.
\end{align}
\end{subequations}
Using the properties of the RKHS this variational system has an explicit
solution in terms of $\bfpt$ and $\bfqt$. Landmarks can be viewed as
measures (see e.g. \cite[Section 6.1]{holm2009euler})
since landmarks can be lifted to $V^*$ by the delta functional $\delta_y(x)$ which equals $1$ when $x=y$ and is otherwise 0, and we write the
right-hand side $\langle \bfpt, \delta u \circ \bfqt\rangle_\RdM$ as a function:
\[
x\mapsto \sum_{i=1}^M p_t^i \cdot \delta u(x) \delta_{q_t^i} (x) ,
\]
Integrating by parts in \eqref{pmomentum} and using $L_V\inv=K_V$ we can write Hamilton's equations as follows where we seek
$u\in\Vsf,\,\mathbf{p}\in \mathbf Q$ (since $\mathbf{Q}\simeq \mathbf Q^*$) and $\mathbf{q}\in \mathbf Q$:
\begin{subequations}\label{eq:ldm}
\begin{align}
&u_t(q_t^j) = \sum_{i=1}^M K_V(q_t^i, q_t^j)p_t^i, \quad
j=1,\ldots,M,\label{eq:ldm:u}\\ 
&\dot{\mathbf p_t} = - \nabla u_t\transp\circ\bfqt \bfpt,\label{eq:ldm:p}\\ 
&\dot{\mathbf q_t} = u_t\circ\bfqt,\label{eq:ldm:q}
\end{align}
\end{subequations}
subject to the boundary conditions $\qzlm$. When we do not enforce an end-point condition, \eqref{eq:ldm:p} and \eqref{eq:ldm:q} are both simple ODEs.
Note that $u$ is fully described by $K_V$, $\mathbf p$ and $\mathbf{q}$ in \eqref{eq:ldm:p} and \eqref{eq:ldm:q}. Since $K_V$ is provided as a parameter and $\mathbf q_0$ is known the system \eqref{eq:ldm} is fully described by $\mathbf p_0$ and its norm can be written as:
\begin{equation}\label{unormrkhs}
\vnorm{u_t}^2 = \sum_{i,j=1}^M p_t^j K_V(q_t^i, q_t^j)p_t^i,
\end{equation}
The initial momentum encodes the forward geodesic motion to provide destination shapes $q_1$ at $t=1$ by integration in time. For a template $\mathbf{q}_0$ we can define the \emph{forward map} $f_{\mathbf{q}_0}:Q\rightarrow Q$ by:
\begin{align}\label{forward}
& \mathbf{p}_0 = \{p_0^i\}_{i=1}^M \mapsto f_{\mathbf{q}_0}[\mathbf{p}_0].
\end{align}
To avoid confusion we let $\mathbf{q}_1$ denote the desired target and $f_{\mathbf{q}_0}[\mathbf{p}_0]$ the landmarks at time $t=1$ in \eqref{eq:ldm}, defining the misfit function by:
\[
\mathbf{p}_0 \mapsto \mathbf q_1 -  f_{\mathbf{q}_0}[\mathbf{p}_0],
\]
which we aim to minimise in the coming sections. Note that in LDDMM it is the sum of this mismatch and the regularisation term \eqref{fnl:lddmm} that is minimised, while in this paper regularisation is introduced in the enKf.\\

For $T$ timesteps $0,\ldots, T-1$ of size $\Delta t$ we choose a forward Euler scheme to discretise the time derivative in these last two equations. This leads to a discrete forward operator depending on $\Delta t$ which, with a slight abuse of notation, shall also be denoted $f$. All of our simulations are implemented in Python using Pytorch \cite{paszke2019pytorch} and KeOps (\url{kernel-operations.io}, see also \cite{charlier2020kernel}). Appendix \ref{appendix:a} contains details on how to obtain and run our code.
\section{Bayesian Inverse Problem}\label{sec:enkf}

\subsection{The Ensemble Kalman Filter}

The enKf is a Monte Carlo data assimilation \cite{reich2015probabilistic} algorithm dating back to 1994 \cite{evensen1994sequential} where the objective is to estimate the state $x$ in some space $\mathcal{X}$ of the system at future times via Bayes' rule:
\begin{equation}\label{bayesrule}
\rho(x|y) \propto \rho(x)\rho(y|x),
\end{equation}
for some prior information $\rho(x)$ about a state $x$ and likelihood $\rho(y|x)$ of a prediction $y|x\in\mathcal{Y}$. In this setting we assume that the prior is Gaussian, but whereas the covariance is prescribed in the standard Kalman filter \cite{kalman1960new}, we compute sample statistics in the enKf from a collection, or \emph{ensemble}, $X = \{X_i\}_{i=1}^{N_E}$, of $N_E$ state vectors taking values in $\mathcal{X}$ equipped with an inner product $\langle\cdot, \cdot\rangle_\mathcal{X}$. The trade-off here is that we must evolve a system of state equations. Moreover, the enKf update is in fact also a nonlinear system owing to the dependence of the Kalman gain on the ensemble itself via the sample statistics, breaking with the Gaussian assumption of the ensemble. In the limit of large ensembles the enKf can be shown to converge to the Kalman filter \cite{mandel2011convergence}.\\

We recall the Gaussian assumptions on the prior and the posterior for the classical Kalman filter, namely $\mu$ and $C$ being the mean and covariance of the prior, and, for an observation operator $H:\mathcal{X}\rightarrow\mathcal{Y}$, $Hx$ and $R$ represents that of the data $y|x$:
\begin{subequations}
\begin{align}
\rho(x) &\propto\exp \Big(-\frac 12 (x-\mu)\transp C\inv(x-\mu) \Big),\label{eq:prior}\\
\rho(y|x)& \propto\exp \Big(-\frac 12 (y-Hx)\transp R\inv(y-Hx) \Big),\label{eq:posterior}
\end{align}
\end{subequations}
from which we can derive an expression for the posterior state $x|y$ once $y$ becomes known. Using the enKf approximations of sample mean $\bar{X}$ and the action of the covariance $C_E$:
\begin{subequations}\label{enkf_approximations}
\begin{align}
& \bar{X} := \frac{1}{N_E}\sum_{i=1}^{N_E} X_i,\\
& C_E[\cdot] := \frac{1}{N_E-1}\sum_{i=1}^{N_E} (X_i-\bar{X})\langle X_i-\bar{X},\cdot\rangle_\mathcal{X},
\end{align}
\end{subequations}
we write the \emph{ensemble} Kalman gain as follows:
\begin{align}\label{kalmanupdate}
& K_E := C_EH\transp(HC_EH\transp + R)\inv,
\end{align}
so that we can form samples of the posterior distribution as follows:
\begin{align}\label{ensembleupdate}
& X_i' := X_i + K_E(y - HX_i),\qquad i=1,\ldots,N_E.
\end{align}
Note that \eqref{ensembleupdate} parallelises across the ensemble members.\\

While the standard enKf is typically used as an inference tool
it has also been proposed as a way to solve general class inverse problems \cite{iglesias2013ensemble,iglesias2016regularizing,li2007iterative}, supported by a wide array of numerical evidence in particular for data assimilation in atmospheric science (see \cite{schneider2017earth} and its bibliography). We briefly outline the steps of an iterative enKf method for a Bayesian inverse problem roughly on the form \emph{given a $y$, find the $x$ such that $y \approx H[x]$}, where $H$ now is a specified observation operator. The key in this approach is that the term $y - H X_i$ in \eqref{ensembleupdate} now represents the misfit that we want to minimise (modulo possible added noise). 
\begin{enumerate}
    \item Let $k=0$ and denote by $X^k = \{X_i^k\}_{i=1}^{N_E}$ the initial ensemble, and compute the statistics in \eqref{enkf_approximations}.
    \item Propagate the ensemble through the observation operator:
      \[
        Y^k := \{Y_i^k\}_{i=1}^{N_E} = \{H[X_i^k]\}_{i=1}^{N_E}.
      \]
    \item Update the ensemble: $X^{k+1} = X^k + K_E (y - Y^k)$, where these operations are understood element-wise across the ensemble.
    \item Verify convergence of $y - \bar{Y^k}$, otherwise increment $k$ and go to step 1.
\end{enumerate}
Under certain assumptions on the linearity of $H$ and the covariance operators $C$ and $R$ it can be shown \cite[Section 2.6]{iglesias2013ensemble} that the iterative enKf approximates, without the use of derivatives, a solution to the Tikhonov-Philips \cite{vogel2002computational} regularised functional:
\[
x\mapsto \| y - H x \|_\mathcal{Y}^2 + \| x - \bar{x} \|_\mathcal{X}^2,
\]
where in this context $\bar{x}$ is the average of $x$.\\

Next we show that the enKf provides a massively parallel and derivative-free method for applications in shape analysis. We demonstrate the utility of this algorithm in the next section where we show numerical evidence of convergence and accuracy for landmark matching problems.

\subsection{Application to Landmark Matching}

Ideally we want to solve the inverse problem of finding the initial momentum $\mathbf{p}_0$ such that $f_{\mathbf{q}_0}(\mathbf{p}_0)$ is close to a target $\mathbf q_1$:
\begin{equation}\label{problem:Bayesian}
\pz^* := \arg\inf_{\pz}
\gammainvnorm{\mathbf q_1 - f_{\mathbf{q}_0}[\pz]}^2.
\end{equation}
While this can be done by traditional shooting approaches, the aim here is to present a derivative-free and parallelisable method. We will in fact not be solving \eqref{problem:Bayesian}, but rather a version of it where we 
optimise over an \emph{ensemble} of initial momenta which, formally speaking, can be considered as a collection of different samples $\pz$ of the prior (defined over $Q$) that each map the template to different target locations. The likelihood is provided by the forward operator as it gives us information about how suitable a candidate $\mathbf p_0$ is in the sense of minimising the misfit between the proposed and the desired target. Note that the energy term in the LDDMM functional is conserved along geodesics and can be rewritten as a quadratic form on the initial momentum $\mathbf p_0$ cf. \eqref{unormrkhs}. This can be interpreted as the logarithm of the probability density for a prior Gaussian distribution on the momentum and provides the link between the LDDMM and the Bayesian formulation.\\

Our aim is then to use the enKf to generate a sequence of ensembles such that the average of the forward map applied to each of its elements converges to the desired target $\mathbf q_1$. Before we make these statements precise we introduce some notation:
\begin{itemize}
\item The space $Q_E$ of ensemble momenta is defined as
the space whose elements are collections of $N_E$ elements of $Q$. Since the enKf is an iterative method we let $P^k := \{ \Pkj \}_{j=1}^\Ne \in Q_E$, $\Pkj\in Q$ denote the state ensemble of momenta at iteration $k$. The objects $\Pkj$ represent initial momentum previously referred to as $\pz$ where the subscript denoted time but in the ensemble setting we prefer this notation. Here, $\bar{P^k}$ denotes the average across the ensemble members of $P^k$ defined for each component $i$ of $\bar{P_i^k}$ by:
\[
\bar{P_i^k} = \nNe \sum_{j=1}^\Ne \Pkj_i.
\]
\item We now define the \emph{ensemble forward map} $F:
Q_E\rightarrow Q$ via \eqref{forward} by:
\begin{equation}\label{Forward}
F_{\mathbf{q}_0}[P^k] := \overline{\{ f_{\mathbf{q}_0}[\Pkj]\}_{j=1}^\Ne} := \nNe \sum_{j=1}^\Ne f_{\mathbf{q}_0}[\Pkj].
\end{equation}
This maps each ensemble member pair to a configuration of destination landmarks and then averaging each landmark across the ensemble. In other words, the forward operator is mapped over the elements of the ensemble and then we take an average in the space $Q$.
\item We define the Kalman update operator at iteration $k$, $\kalmanp^k: Q \rightarrow Q$ be defined by:
\begin{align}\label{eqs:kalmanops}
& \kalmanp^k = \cov_{PQ}^k [\cov_{QQ}^k + \xi \Lambda]^{-1},
\end{align}
where $\xi>0$ is determined later and $\Lambda \in \mathbb{R}^{M\times M}$ is the identity matrix and the actions of the covariance matrices are given by:
\begin{align*}
& \cov_{QQ}^k[\cdot] = \frac{1}{\Ne - 1}\sum_{j=1}^\Ne (f_{\mathbf{q}_0}[\Pkj] -
F_{\mathbf{q}_0}[P^k]) \ipRd{f_{\mathbf{q}_0}[\Pkj] - F[P^k]}{\cdot},\\
& \cov_{P Q}^k[\cdot] = \frac{1}{\Ne - 1}\sum_{j=1}^\Ne (\Pkj - \bar{P^k}) \ipRd{f_{\mathbf{q}_0}[\Pkj] - F_{\mathbf{q}_0}[P^k]}{\cdot}.
\end{align*}
The multiplicative term in \eqref{eqs:kalmanops} can be seen as a countermeasure to overfitting since the matrix $\Lambda$ plays the role of the noise filter. In practice it equals the identity, and the scalar $\xi$ acts as a scaling constant tuned by experiments.
\end{itemize}

Algorithm \ref{enkfdiffeo} describes the enKf applied to shape matching and takes a similar form to the algorithm described in \cite[Section
2.2]{iglesias2016regularizing}, where $P^0$ is the initial ensemble, $\mathbf q_1$ and $\mathbf q_0$ the template and target we wish to match, $n$ is a maximum number of Kalman iterations and $\epsilon$ is a predefined error tolerance. In practice this error tolerance depends on the noise level of the measurements of $\mathbf q_1$ (if these came from some instruments, for instance), but in this paper we work only with synthetic data so we leave the investigation of a more sophisticated early termination criteria as future work. An interesting observation we draw from algorithm \ref{enkfdiffeo} is that although the Kalman gain is inherently nonlinear, the momentum ensemble at iteration $k\geq 1$ is a linear combination of the ensemble at step $k$. 
The choice of initial ensemble is therefore important as the enKf seeks the best approximation to the target in the space of shapes spanned by applying the forward operator to the ensemble momenta.
\begin{algorithm}[ht]
\begin{algorithmic}
\caption{Ensemble Kalman Filter for diffeomorphic shape matching}\label{enkfdiffeo}
\Procedure{enKfDiffeo}{$P^0$, $\mathbf q_0$, $\mathbf q_1$, $n$, $\epsilon$}
\State $k \gets 0$
\While{$k < n$}
\If {$\gammainvnorm{\mathbf q_1 - F_{\mathbf{q}_0}[P^k]} \leq \epsilon$} \Return $P^k$
\Else
    \For{$j \gets 1,\ldots, \Ne$}
      \State $P^{k+1,j} \gets P^{k,j} +\kalmanp^k\Big( \mathbf q_1 -F_{\mathbf{q}_0}[P^{k,j}]\Big) $, 
    \EndFor
    \State $k\gets k+1$
\EndIf
\EndWhile
\Return $P^k$
\EndProcedure
\end{algorithmic}
\end{algorithm}




\section{Numerical Examples}\label{numerical_results}

\begin{figure}[ht]\centering
  \includegraphics[width=.32\linewidth]{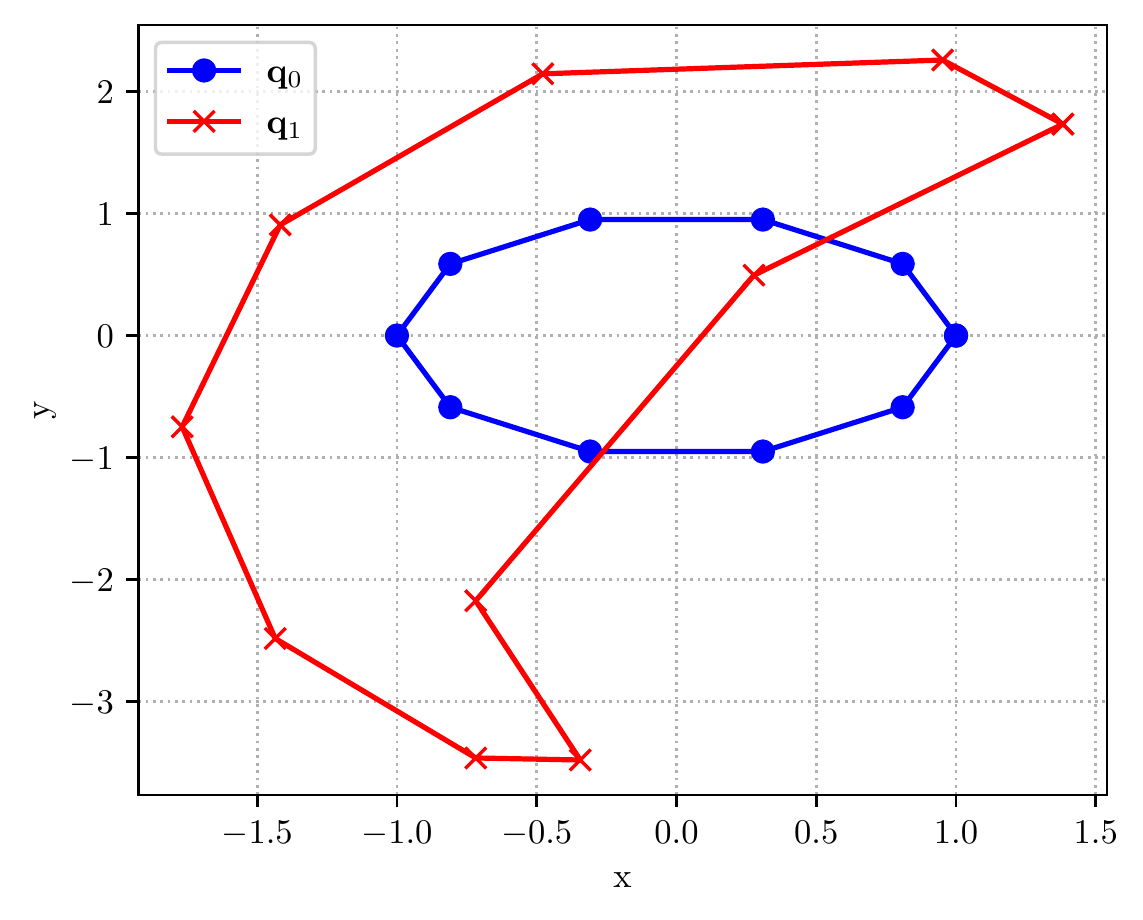}
  \includegraphics[width=.32\linewidth]{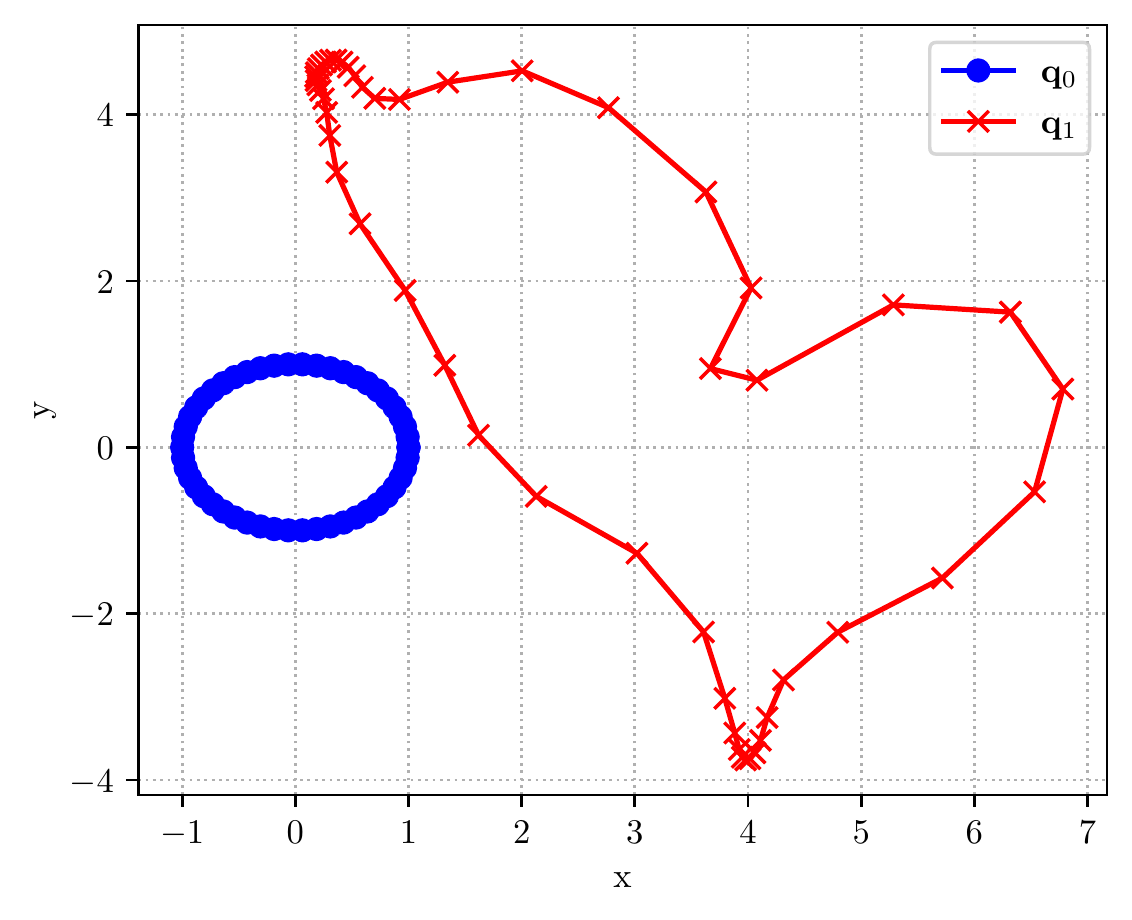}
  \includegraphics[width=.32\linewidth]{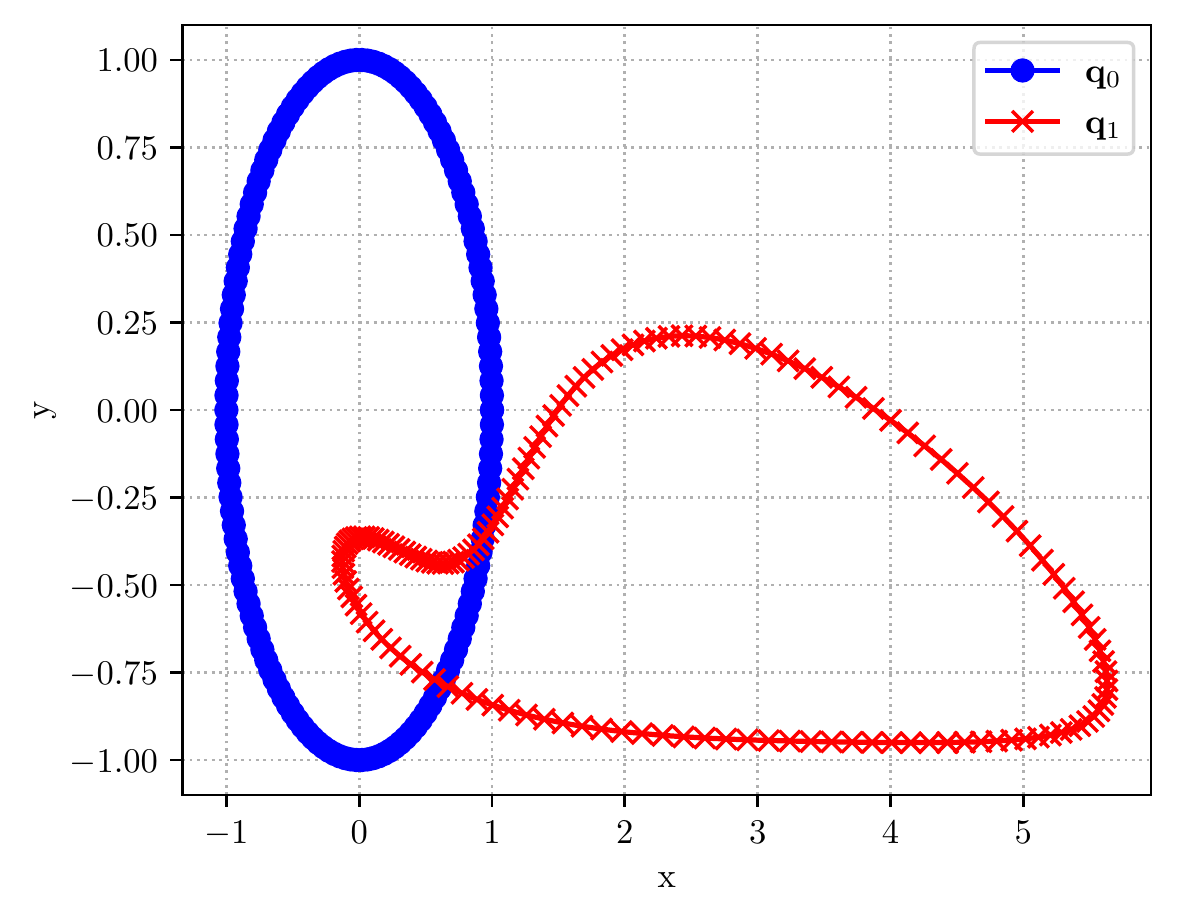}
\caption{Template-target configurations for different values of $M$. Left to right: 10, 50, 150. Linear interpolation has been used between the landmarks to improve the visualisation.}
\label{fig:templates_and_targets}
\end{figure}

In this section we show some numerical experiments using algorithm \ref{enkfdiffeo}.
We generate, for $M\in \{10, 50, 150\}$, synthetic targets by sampling normally distributed initial momenta $\mathbf{p}_0\sim\NNEdM$ and applying the forward map $f$ defined in \eqref{forward} to generate template-target configurations such as those presented in figure \ref{fig:templates_and_targets}. For convenience we always sample the template shape from the unit circle. The objective in this section is two-fold; first we study the effect of the regularisation parameter $\xi$ in section \ref{sec:regularisation} and in section \ref{sec:landmarkvsensemble} the interplay between $N_E$ (the size of the ensemble) and $M$ (the number of landmarks).

\subsection{The Effect of Regularisation}\label{sec:regularisation} 

We first investigate how the scalar $\xi$ in \eqref{eqs:kalmanops} affects the performance of algorithm \ref{enkfdiffeo}. First we define the data misfit $E^k$ at iteration $k$ as:
\begin{equation}
E^k := \gammainvnorm{\mathbf q_1 - F_{\mathbf{q}_0}[P^k]}^2.
\end{equation}

The components of our initial ensembles $P^0$ are sampled from the uniform distribution:
\begin{equation}\label{eq:initial_momentum_N01}
P^{0,j} \sim \Uniformdm, \quad j=1,\ldots, N_E.
\end{equation}
We keep the remaining parameters fixed, see table \ref{global_parameters}.
 As we are investigating the convergence of the filter we use a very low error tolerance. 
\begin{table}[ht]
\centering
\caption{Global parameters used for algorithm \eqref{enkfdiffeo}.}\label{global_parameters}
\begin{tabular}{c|r|l}
\bfseries Variable  &\bfseries Value &\bfseries Description\\\hline
$n$  & 50 & Kalman iterations\\
$T$  & 15 & time steps\\
$\xi$ & 1 & regularisation parameter\\
$\tau$ & 1 & landmark size (cf. \eqref{KVdef})\\
$\epsilon$ & 1e-05 & absolute error tolerance
\end{tabular}
\end{table}

\begin{figure}[ht]\centering    
  \includegraphics[width=.32\linewidth]{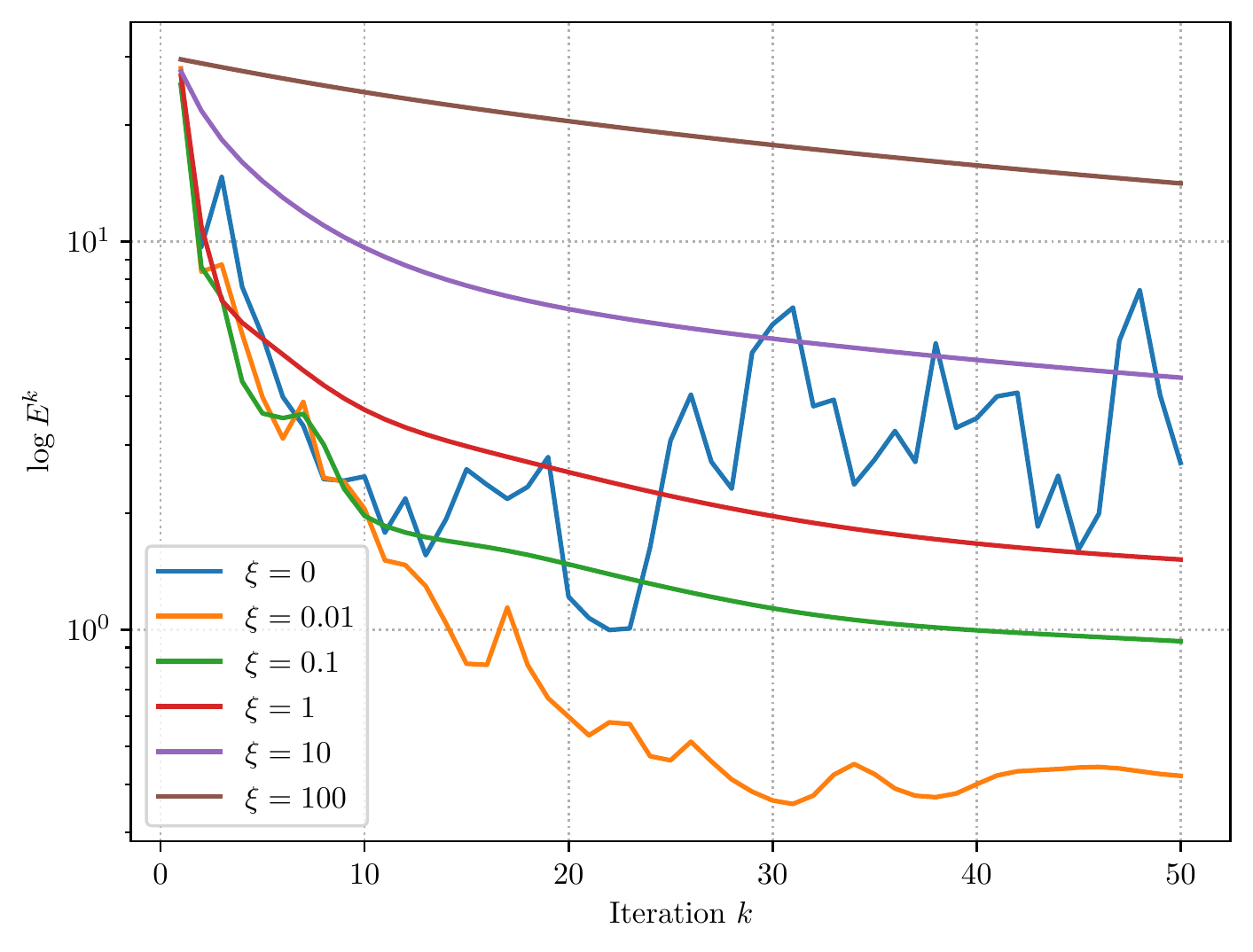} 
  \includegraphics[width=.32\linewidth]{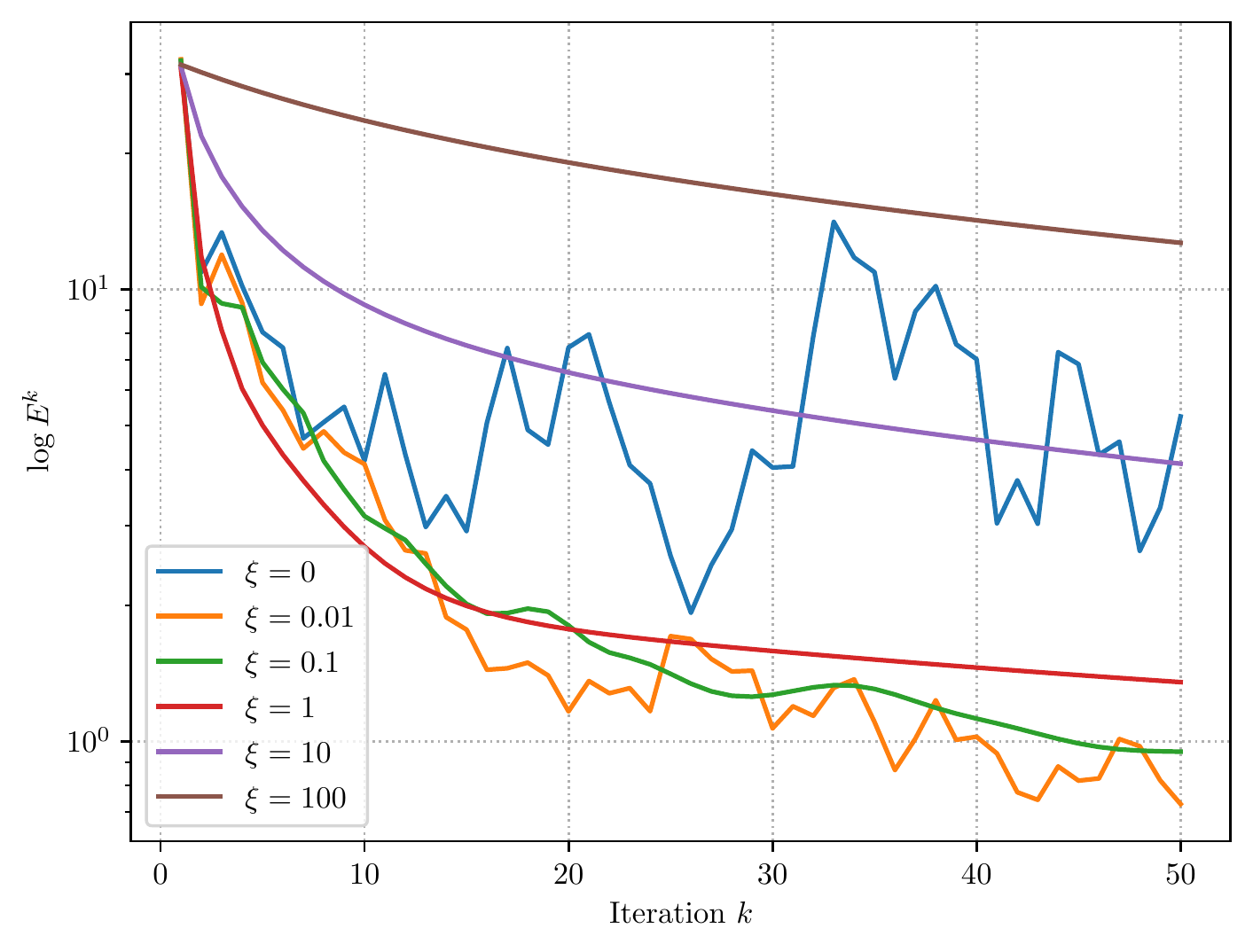}
  \includegraphics[width=.32\linewidth]{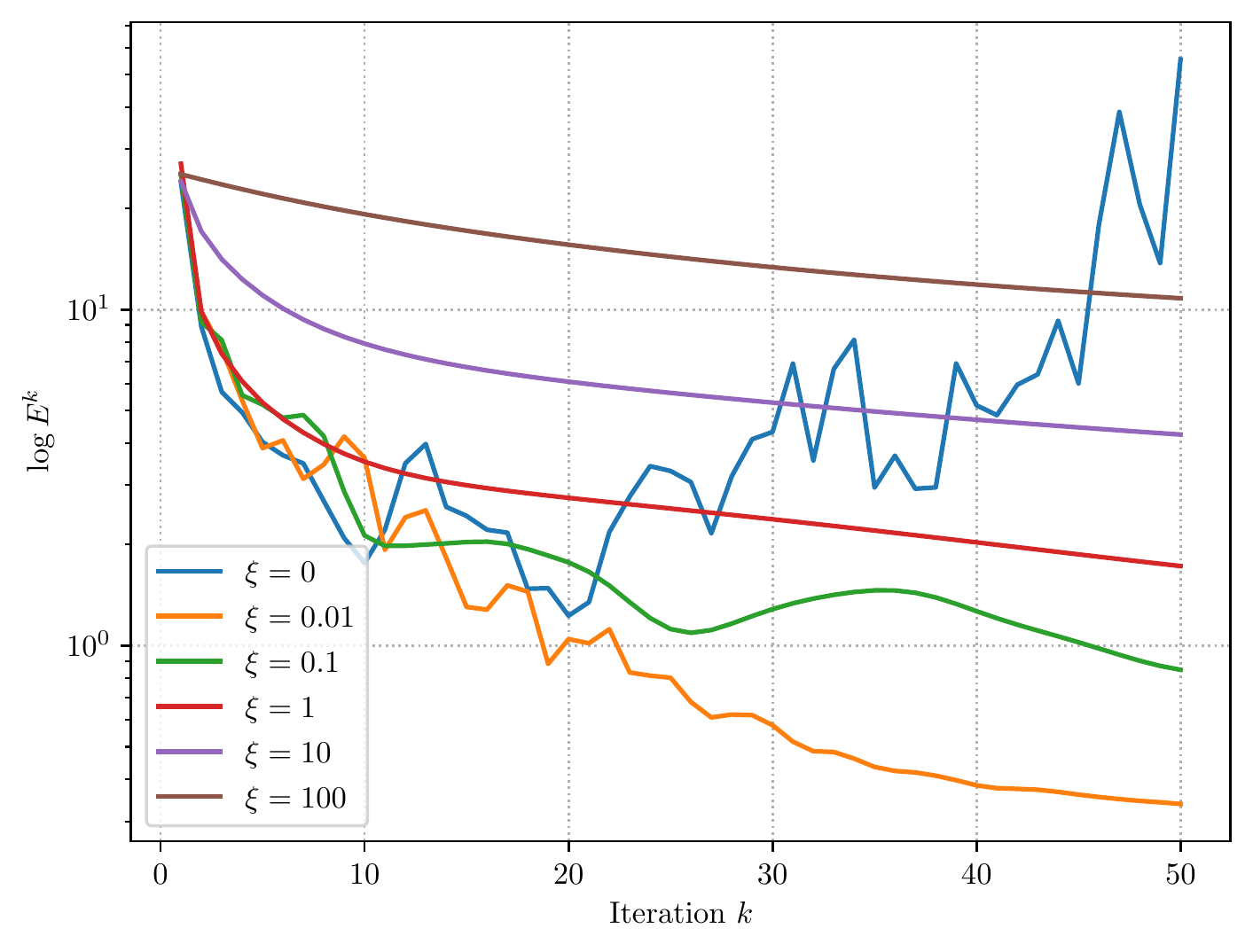}
\caption{Log data misfits for $M=N_E=50$ for different values of $\xi$ using three different targets.}
\label{fig:regularisation}
\end{figure}
For $M=N_E=50$ figure \ref{fig:regularisation} shows the log data misfits for three different targets (shown later on in figure \ref{fig:L50}) using different values of $\xi$. There is a clear trade-off between the rate of convergence and overfitting. As these figures show, a high value of $\xi$ smooths the convergence of the filter at the cost of curtailing the rate. This is expected as the covariance $\cov_{QQ}^k$ plays a smaller role in the term $[\cov_{QQ}^k + \xi \Lambda]^{-1}$ in \eqref{eqs:kalmanops} as the value of $\xi$ increases. In other words, autocorrelations are valued higher in the filter than (possibly) spurious modes that have a higher impact on the Kalman gain when $\xi$ is smaller and this helps to emphasise the way in which $\xi$ acts as a noise filter. We find that some regularisation $\xi$ is necessary in the later iterations of the filter, while setting $\xi$ too high impedes convergence in the initial iterations.

\subsection{Landmark versus Ensemble Size}\label{sec:landmarkvsensemble}

Now we look at how algorithm \ref{enkfdiffeo} performs for each value of $M$ above for different ensemble sizes $N_E\in \{10, 50, 100\}$ when provided with random ensembles whose members are normally distributed cf. \eqref{eq:initial_momentum_N01}. We let $\xi=1$ but otherwise use the parameter values described in table \ref{global_parameters}.\\

\begin{figure}[ht]\centering
  \includegraphics[width=.32\linewidth]{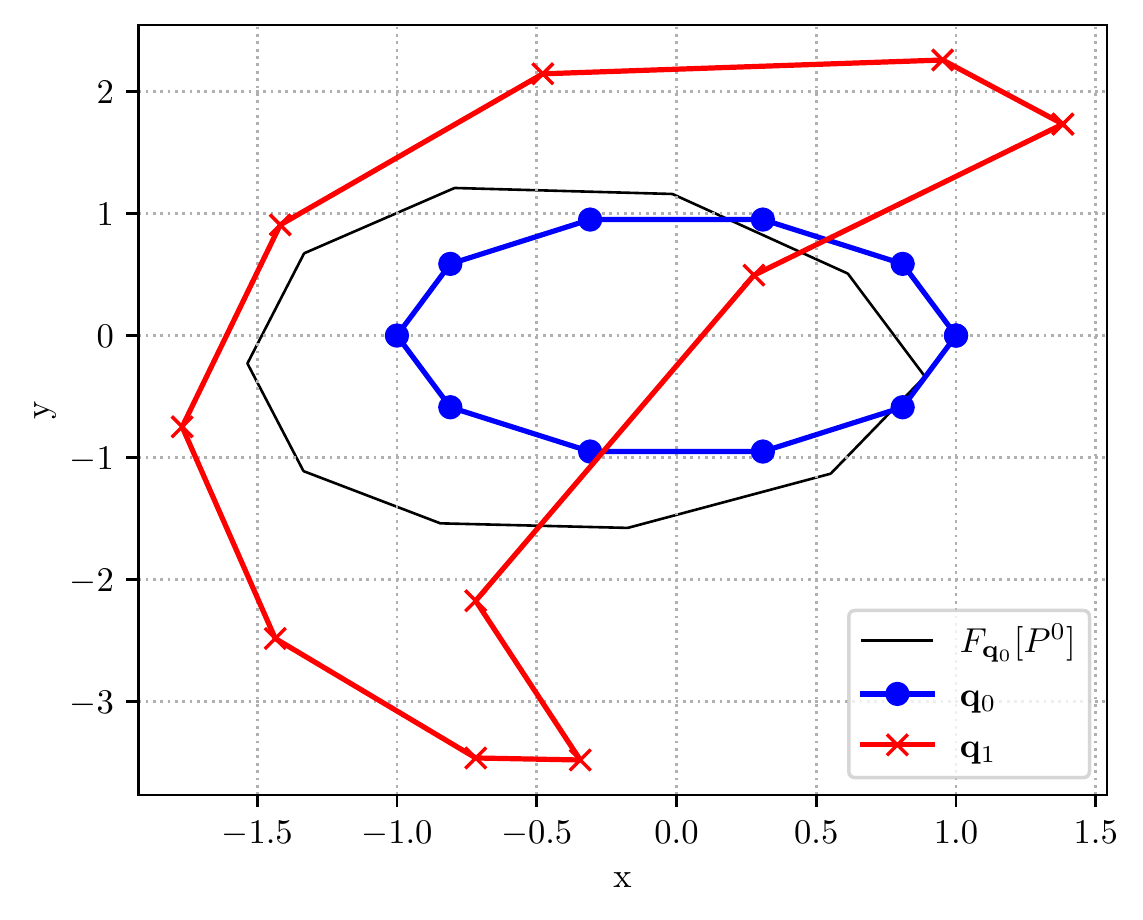}
  \includegraphics[width=.32\linewidth]{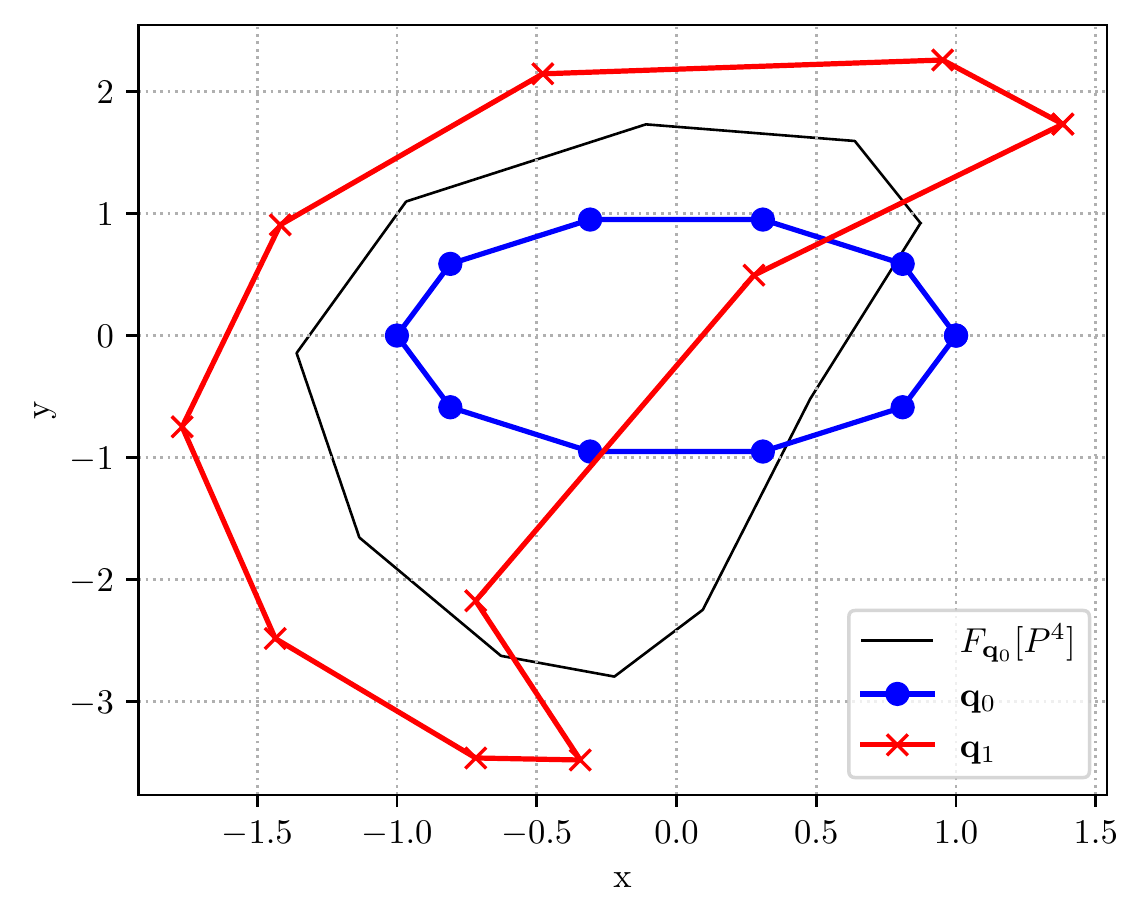}
  \includegraphics[width=.32\linewidth]{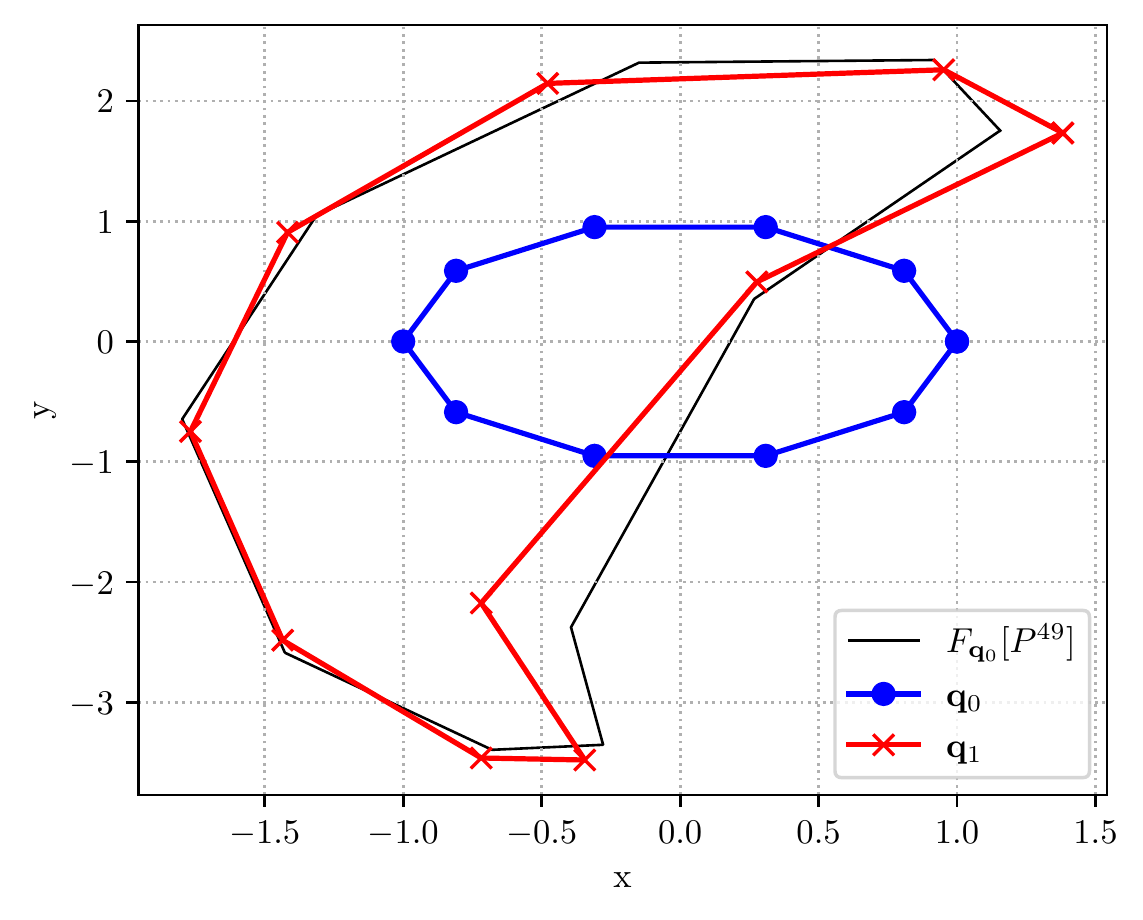}\\
  \includegraphics[width=.32\linewidth]{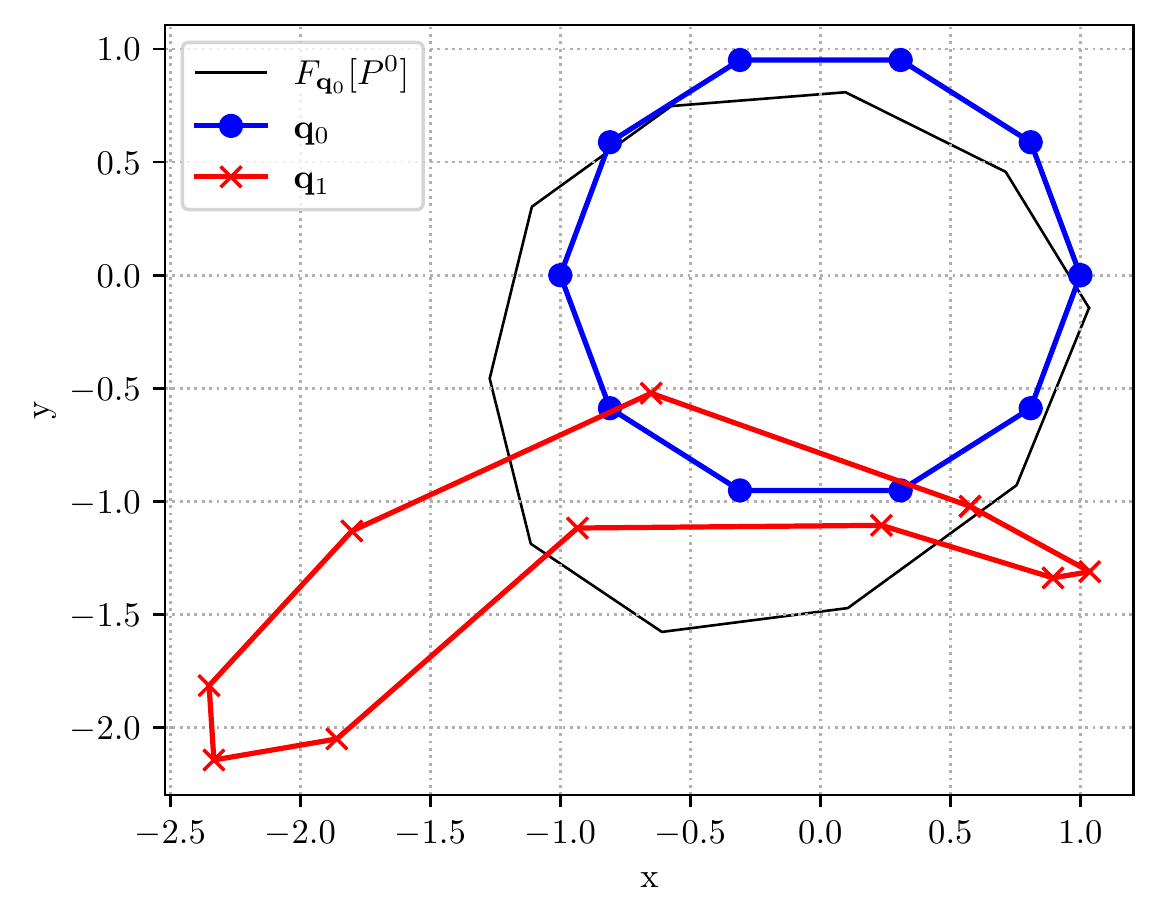}
  \includegraphics[width=.32\linewidth]{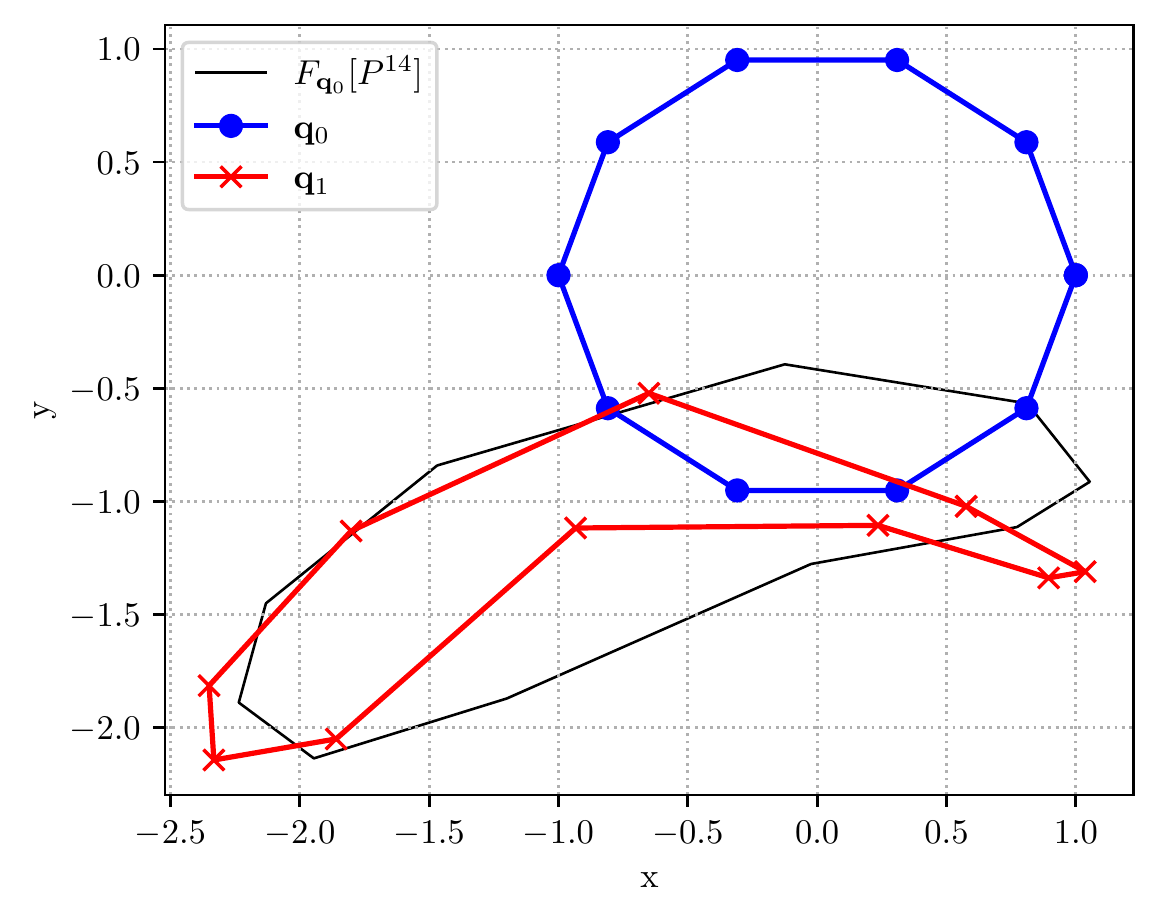}
  \includegraphics[width=.32\linewidth]{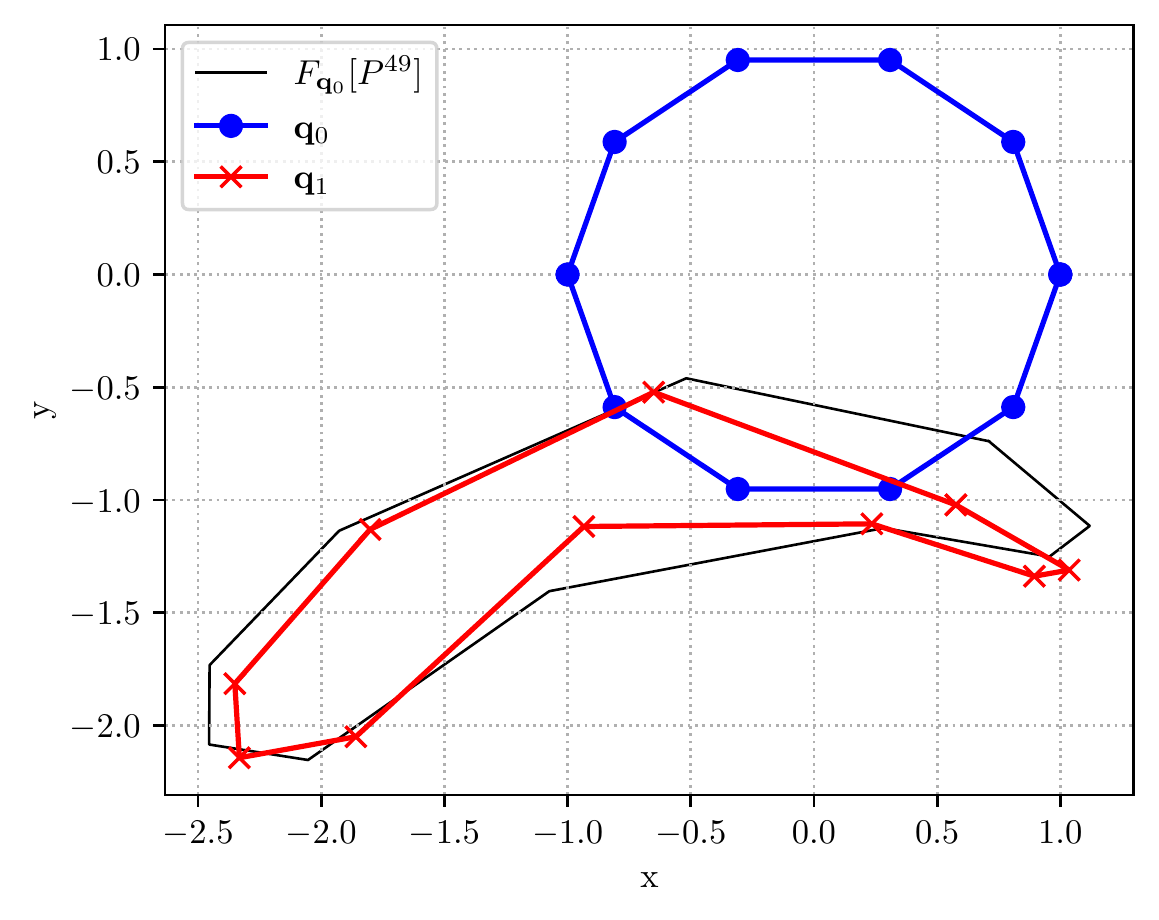}\\
  \includegraphics[width=.32\linewidth]{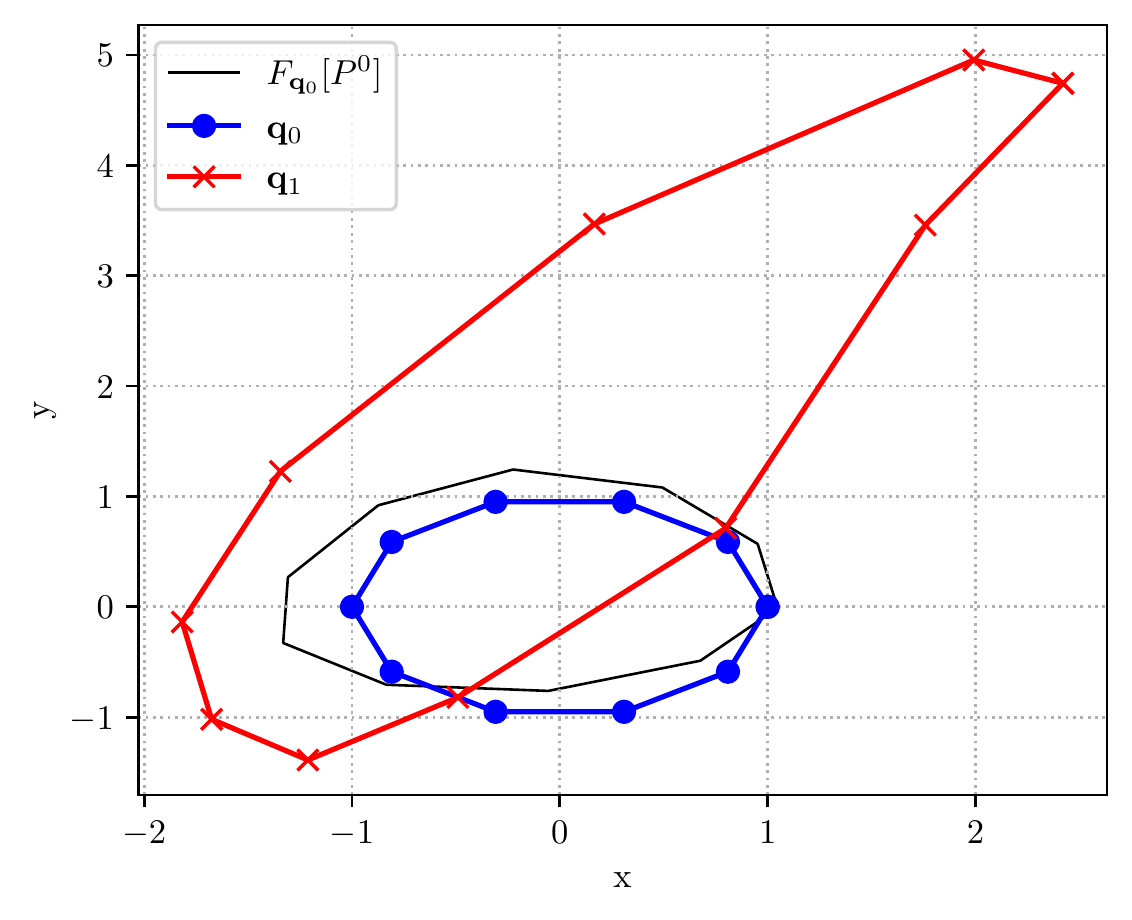}
  \includegraphics[width=.32\linewidth]{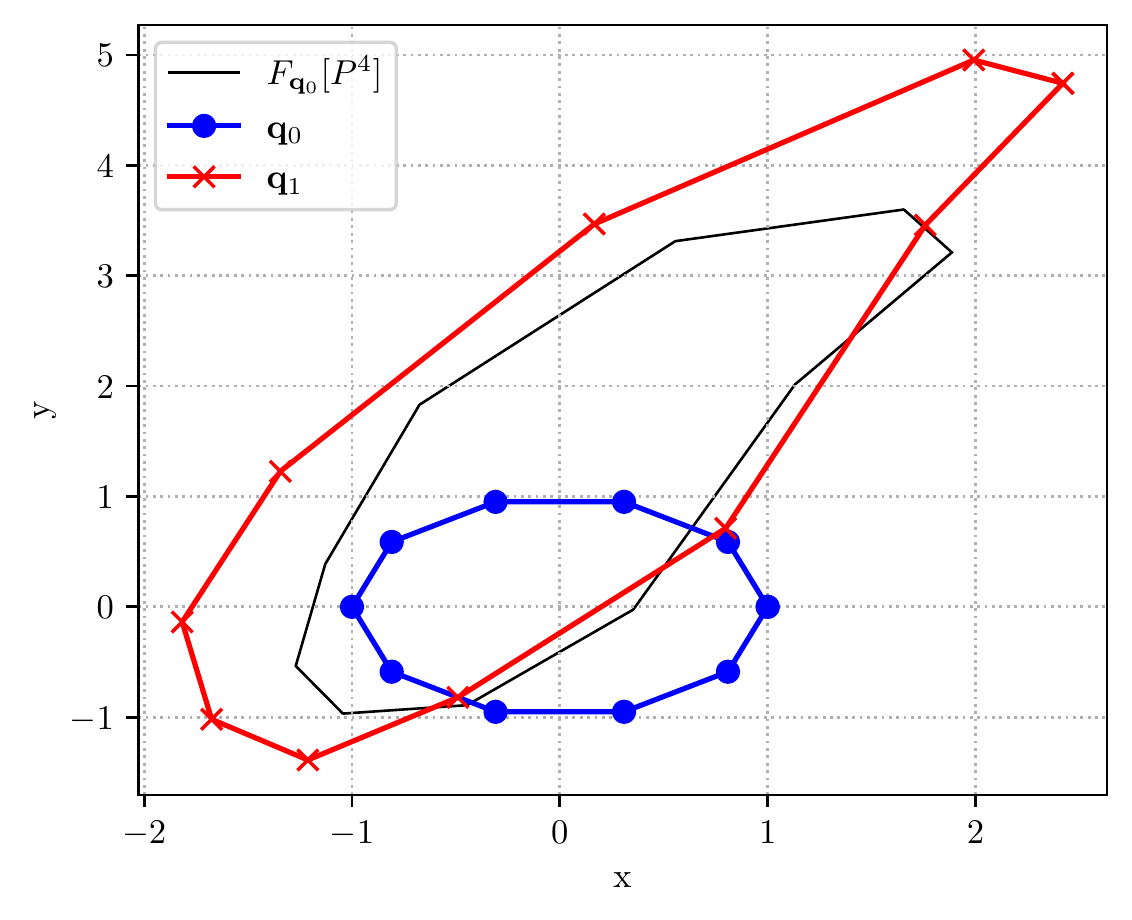}
  \includegraphics[width=.32\linewidth]{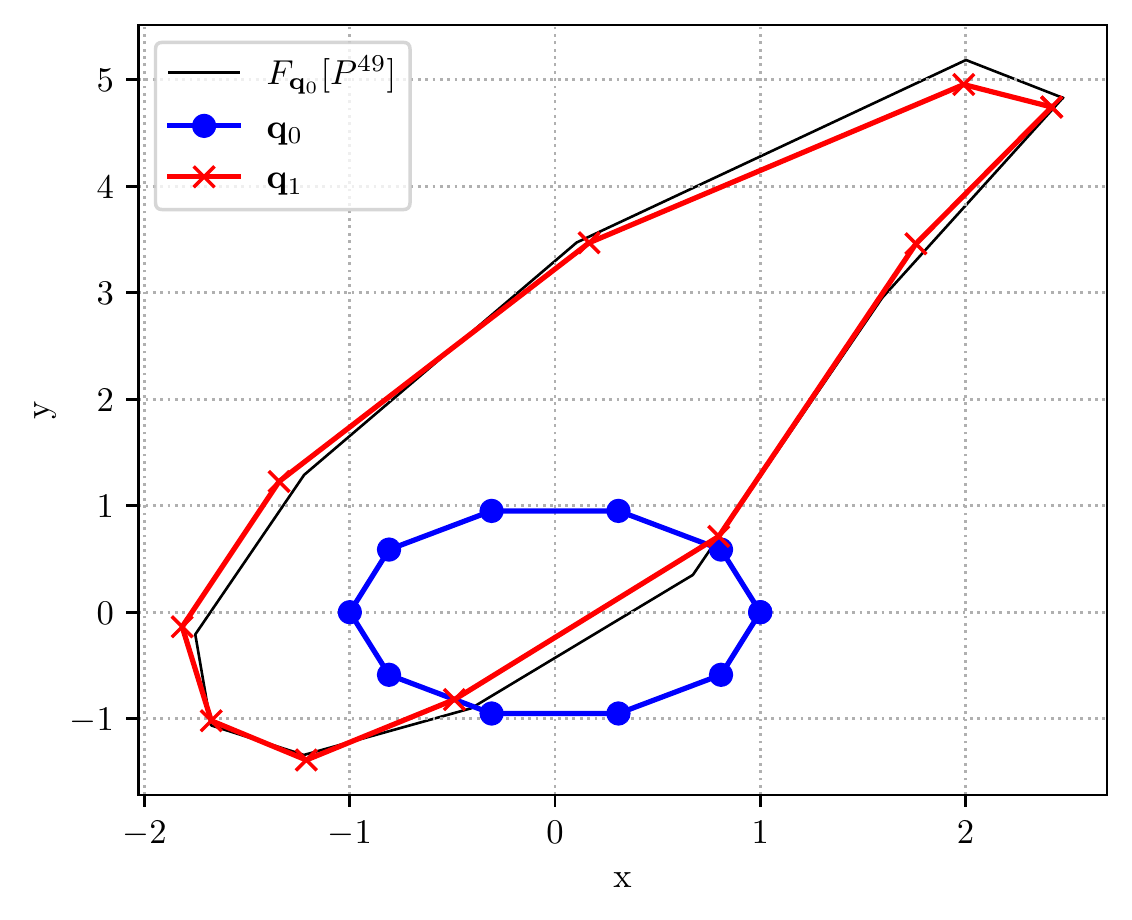}
\caption{Progression of algorithm \ref{enkfdiffeo} for various targets using $M=10$ and $N_E=10$. The left-most column shows how different the initial shape $F_{\mathbf{q}_0}[P^0]$ is from the target. Computation times for 50 iterations: 6s for each configuration.}
\label{fig:L10}
\end{figure}

\begin{figure}[ht]\centering
  \includegraphics[width=.32\linewidth]{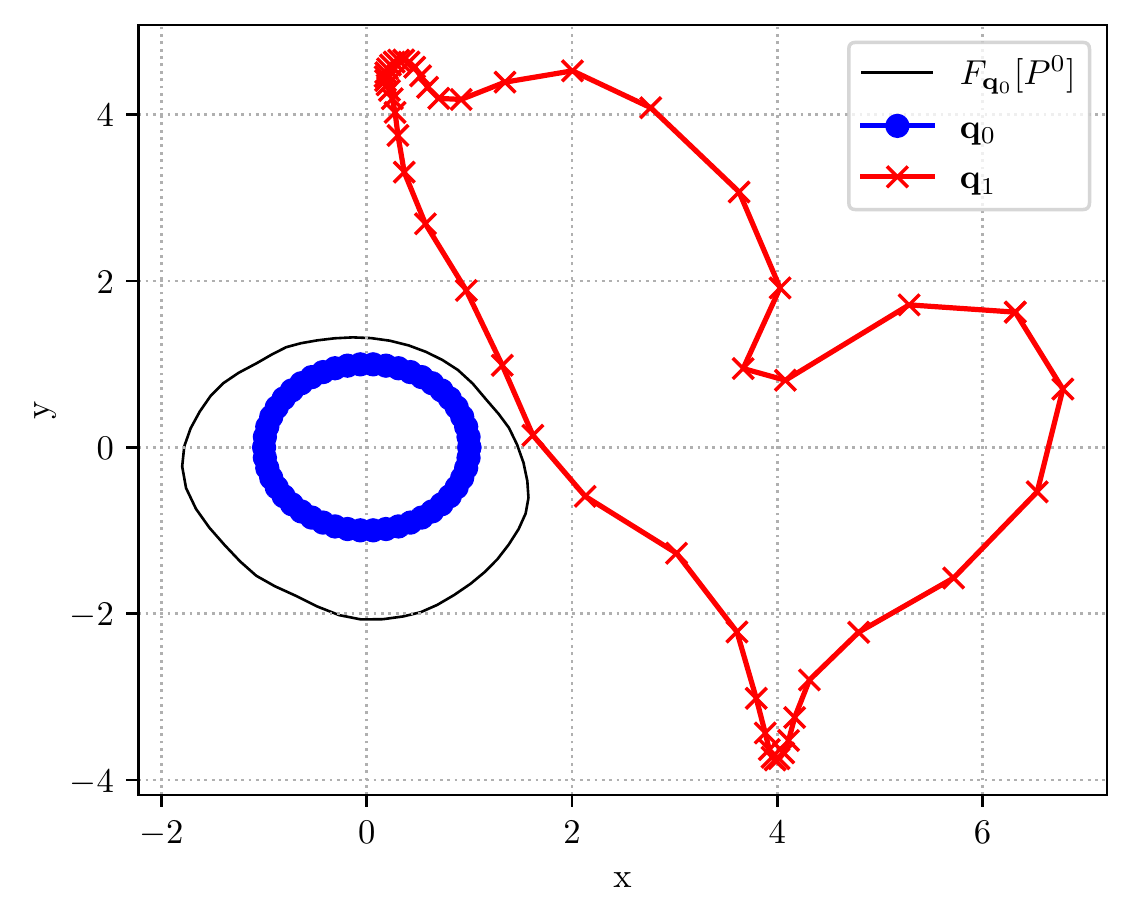}
  \includegraphics[width=.32\linewidth]{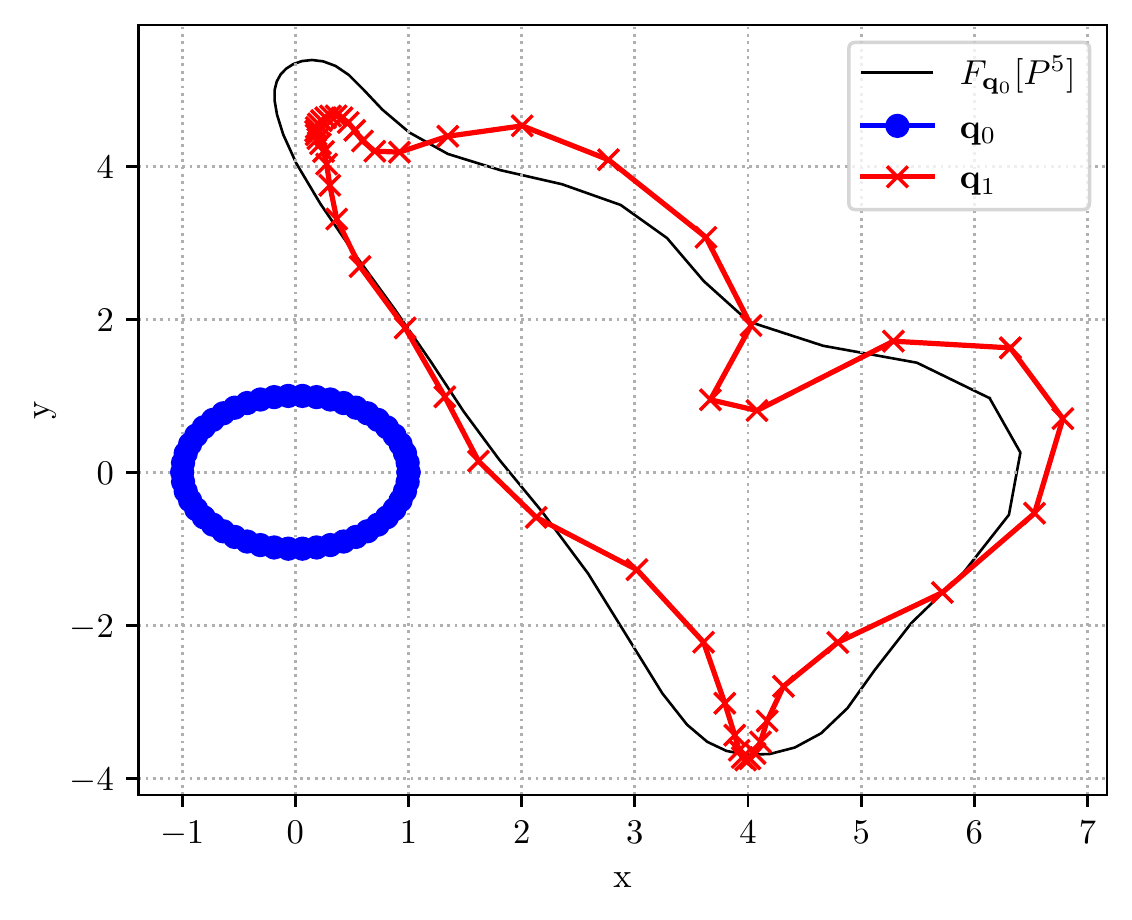}
  \includegraphics[width=.32\linewidth]{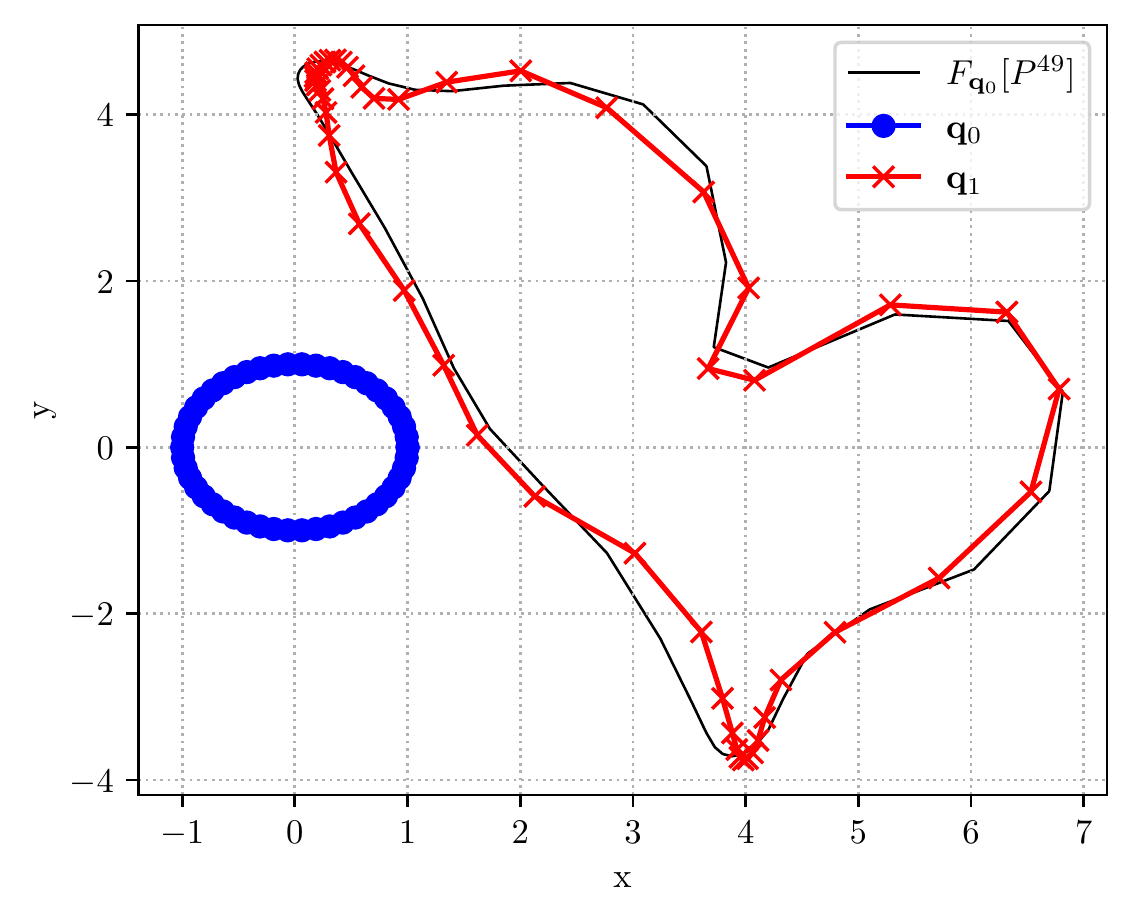}\\
  \includegraphics[width=.32\linewidth]{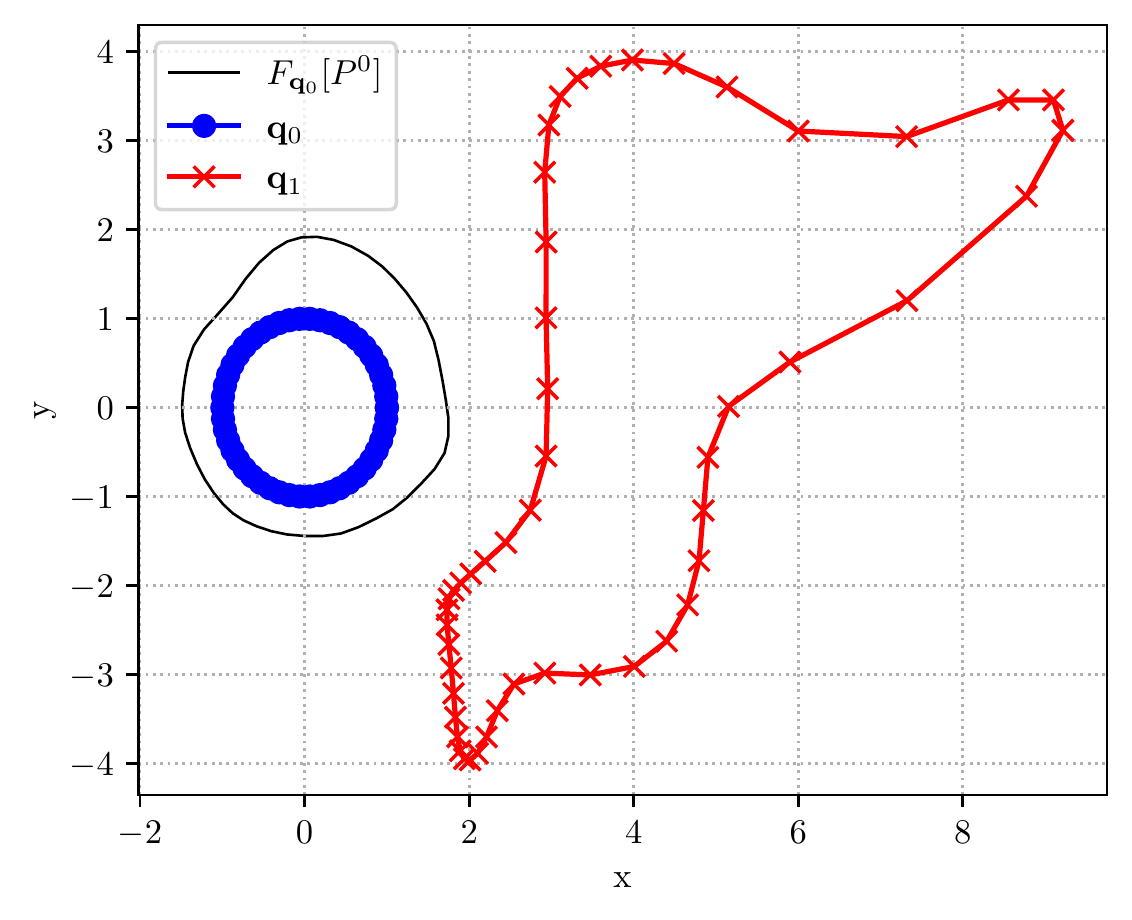}
  \includegraphics[width=.32\linewidth]{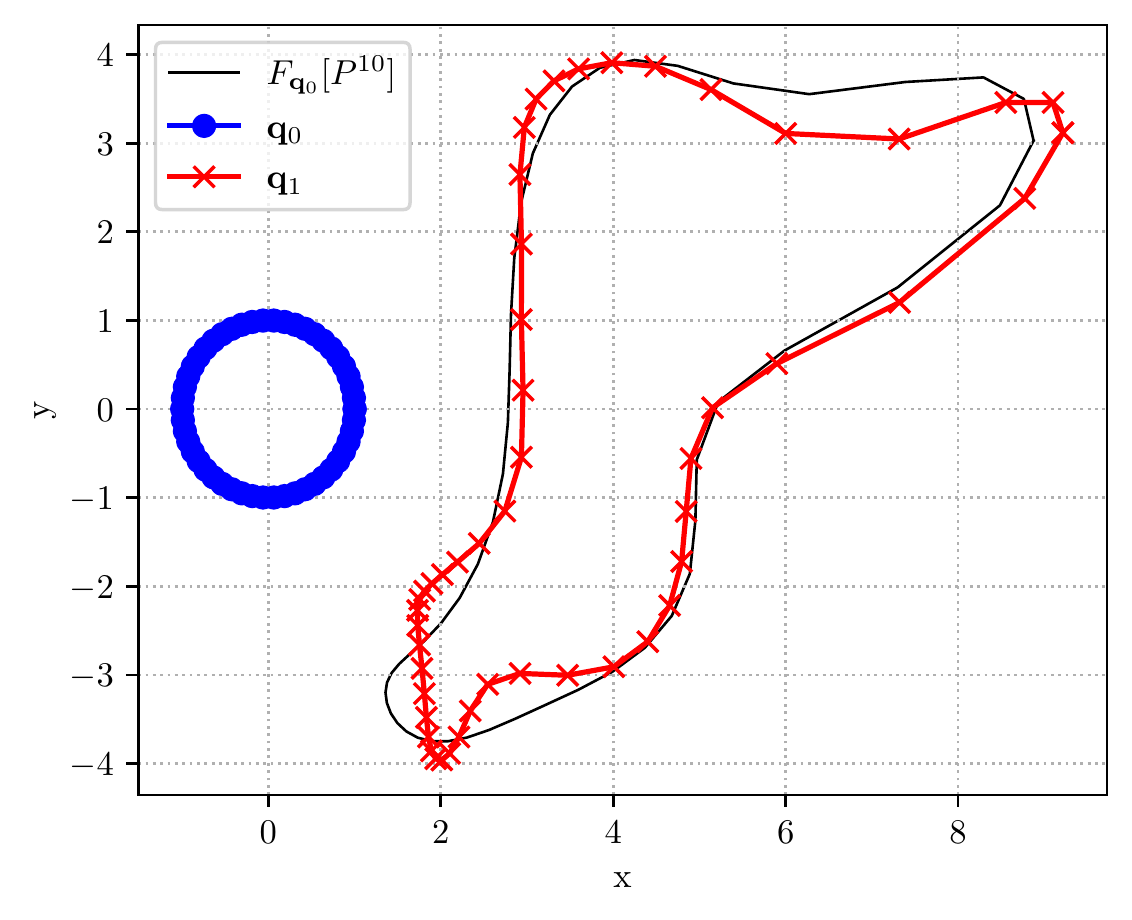}
  \includegraphics[width=.32\linewidth]{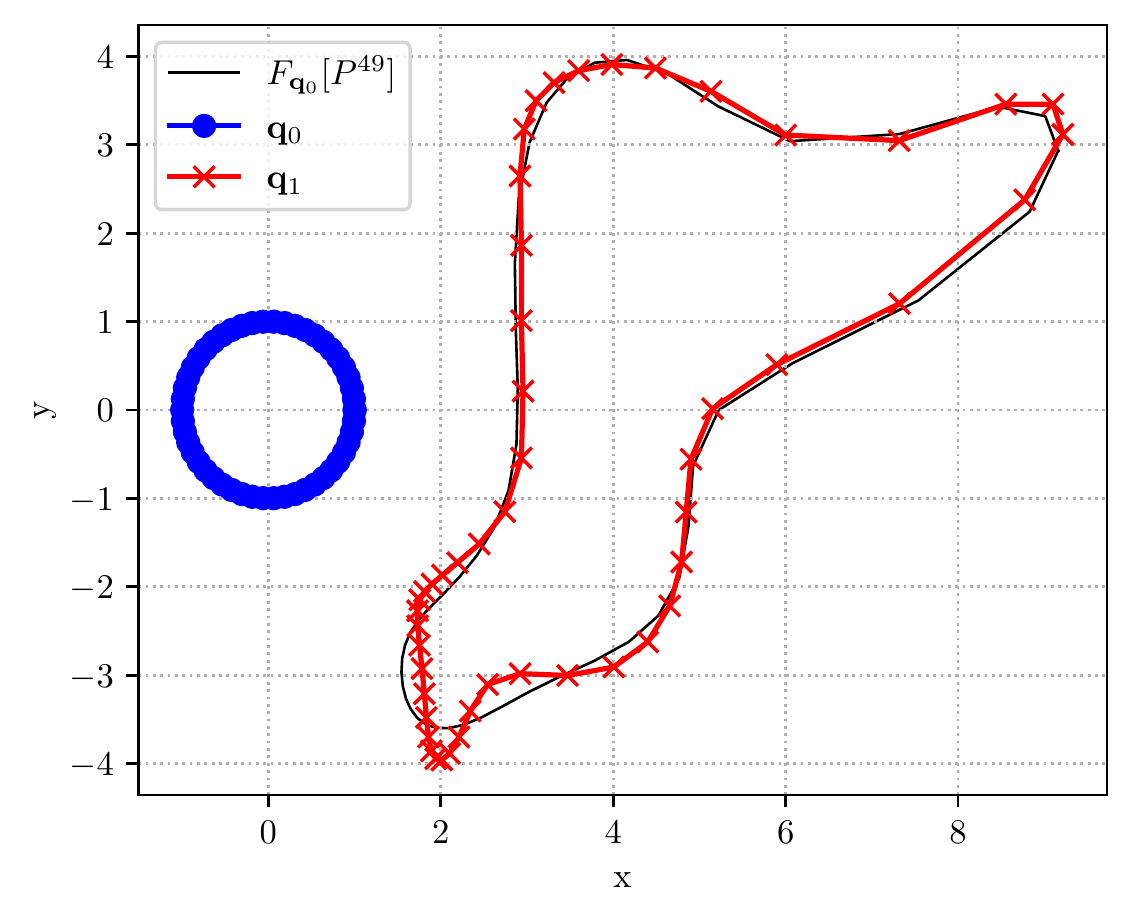}\\
  \includegraphics[width=.32\linewidth]{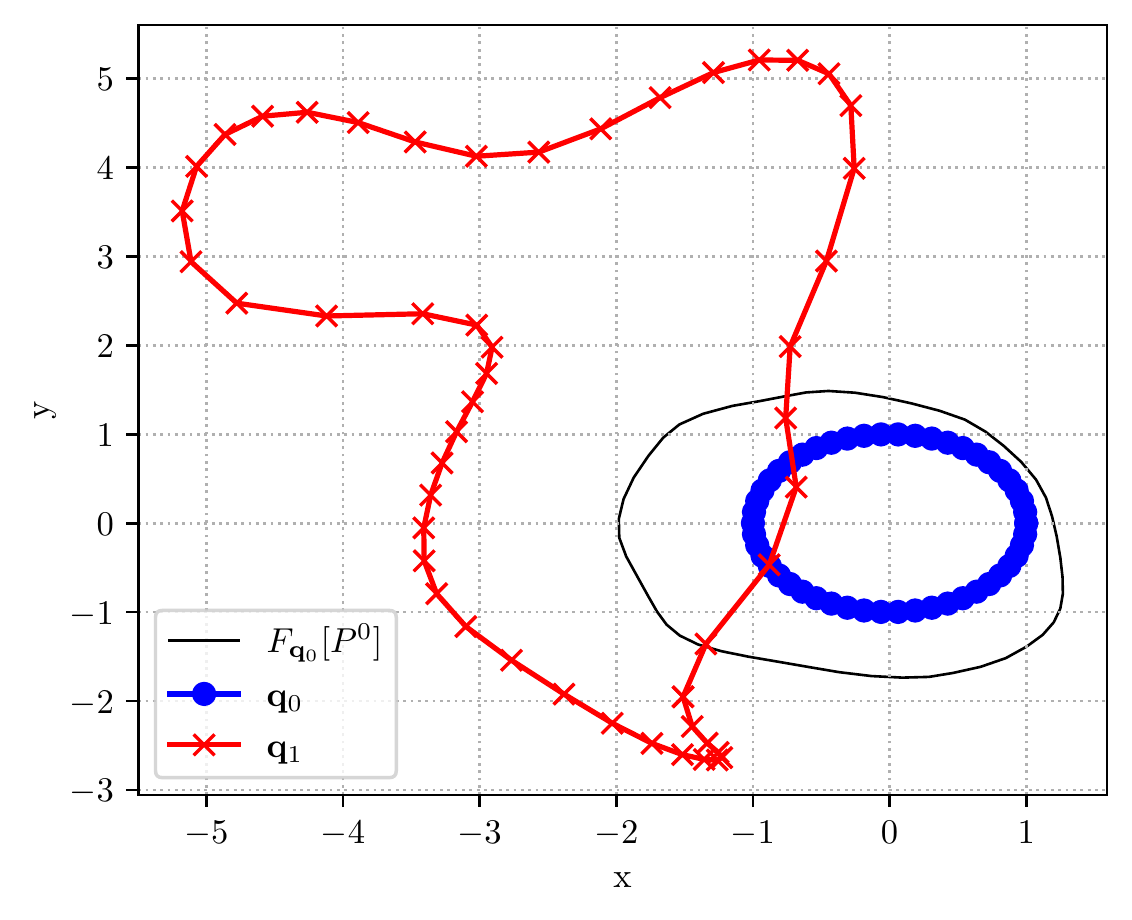}
  \includegraphics[width=.32\linewidth]{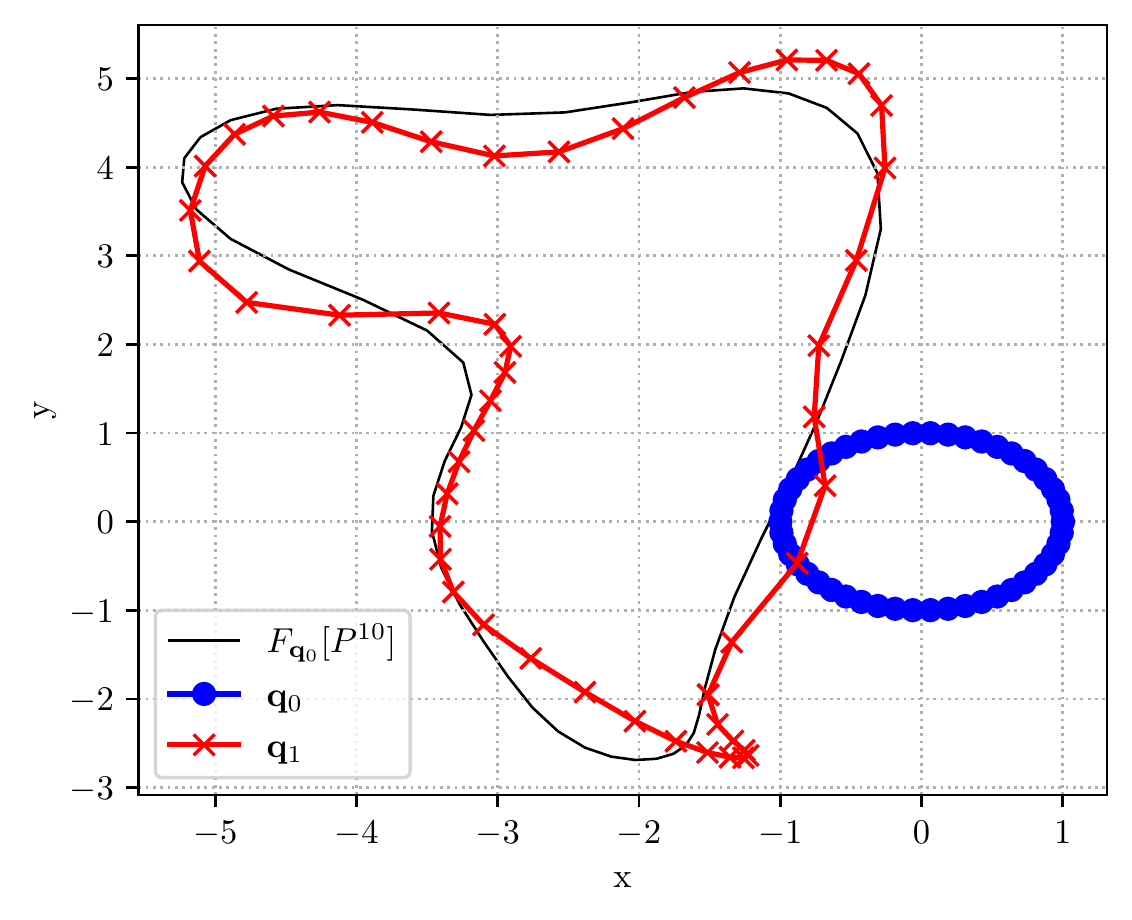}
  \includegraphics[width=.32\linewidth]{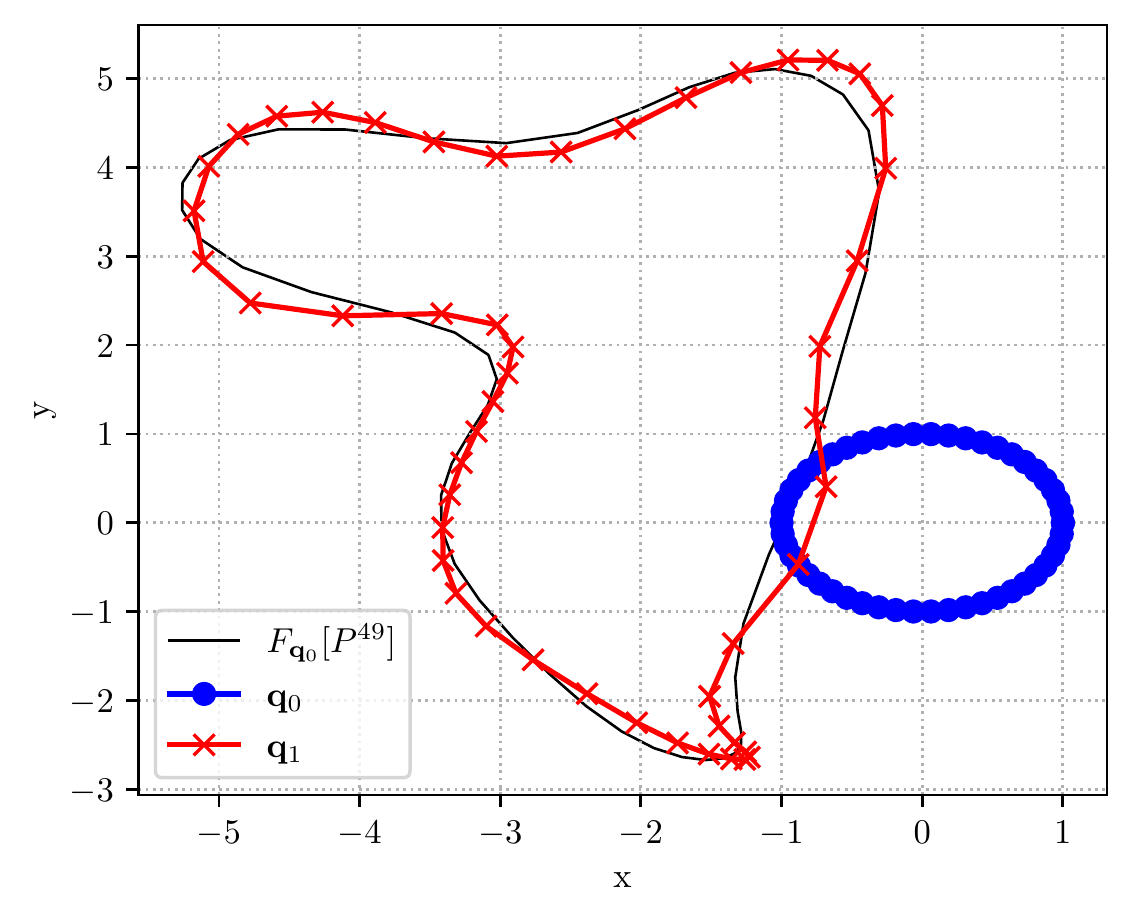}
\caption{Progression of algorithm \ref{enkfdiffeo} for various targets using $M=50$ and $N_E=50$. The left-most column shows how different the initial shape $F_{\mathbf{q}_0}[P^0]$ is from the target. Computation times for 50 iterations (top to bottom): 2m8s, 2m9s, 1m29s.}
\label{fig:L50}
\end{figure}

\begin{figure}[ht]\centering
  \includegraphics[width=.32\linewidth]{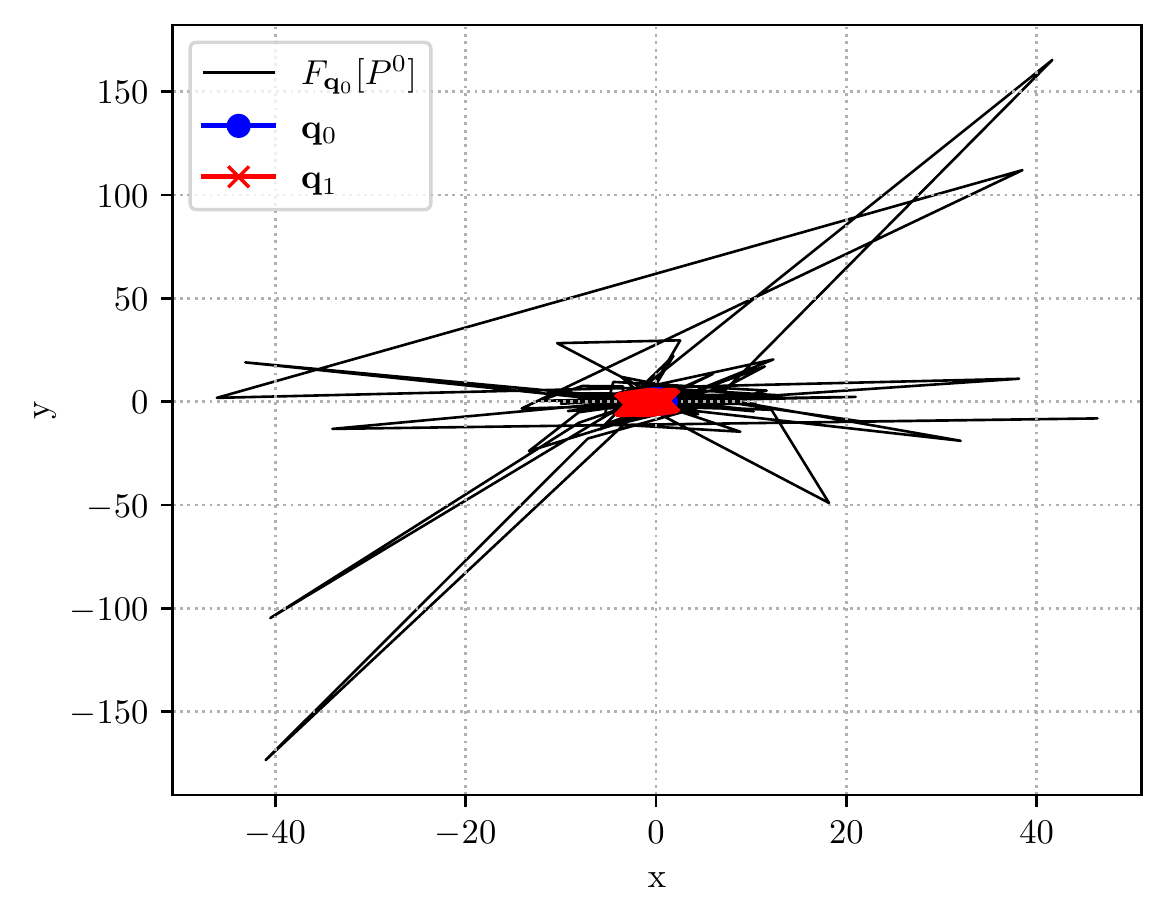}
  \includegraphics[width=.32\linewidth]{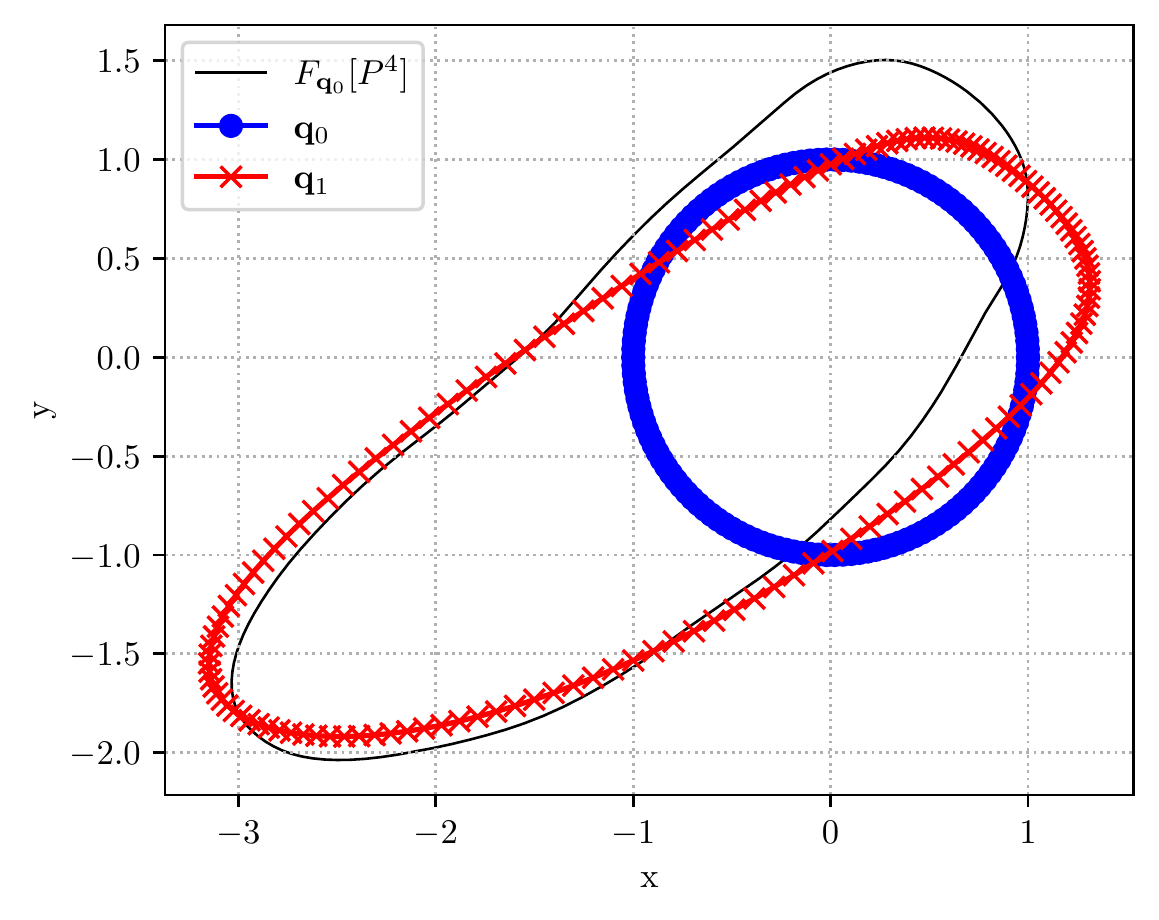}
  \includegraphics[width=.32\linewidth]{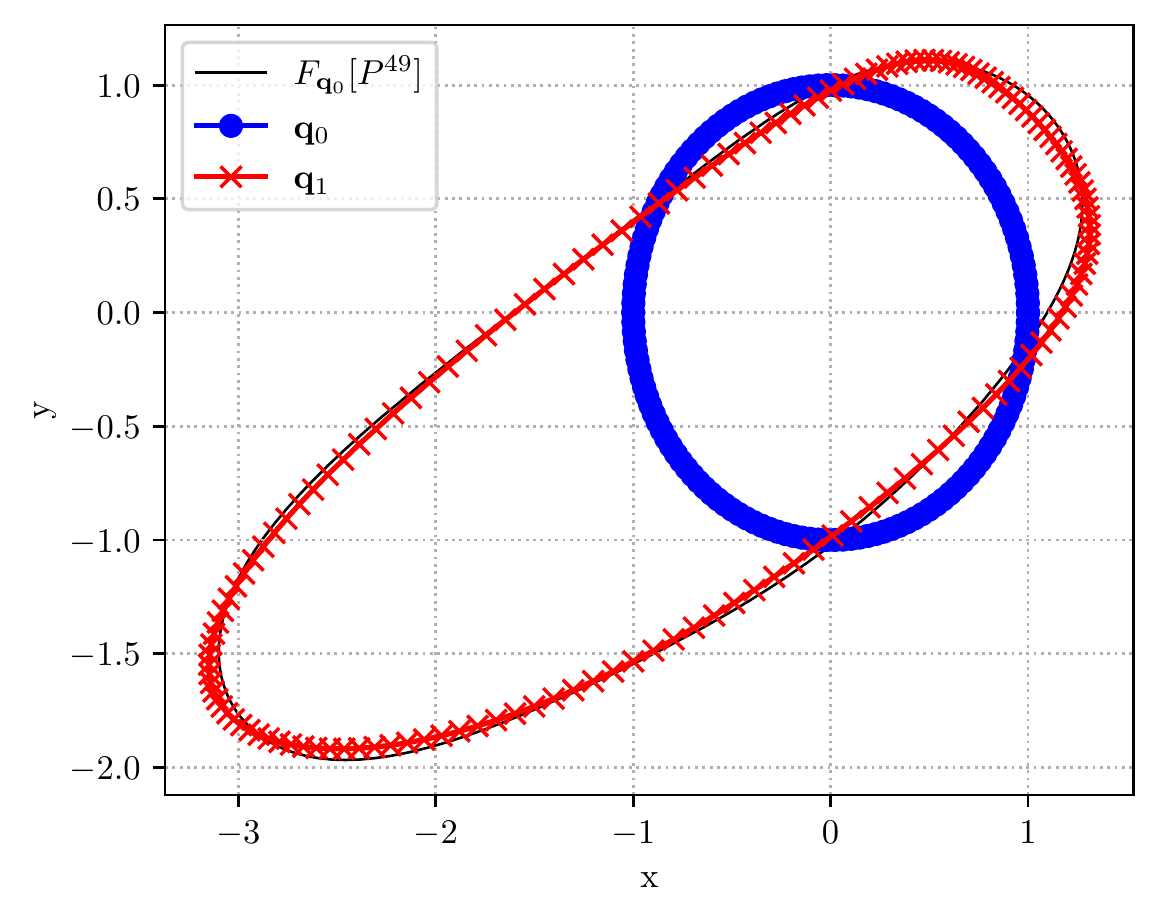}\\
  \includegraphics[width=.32\linewidth]{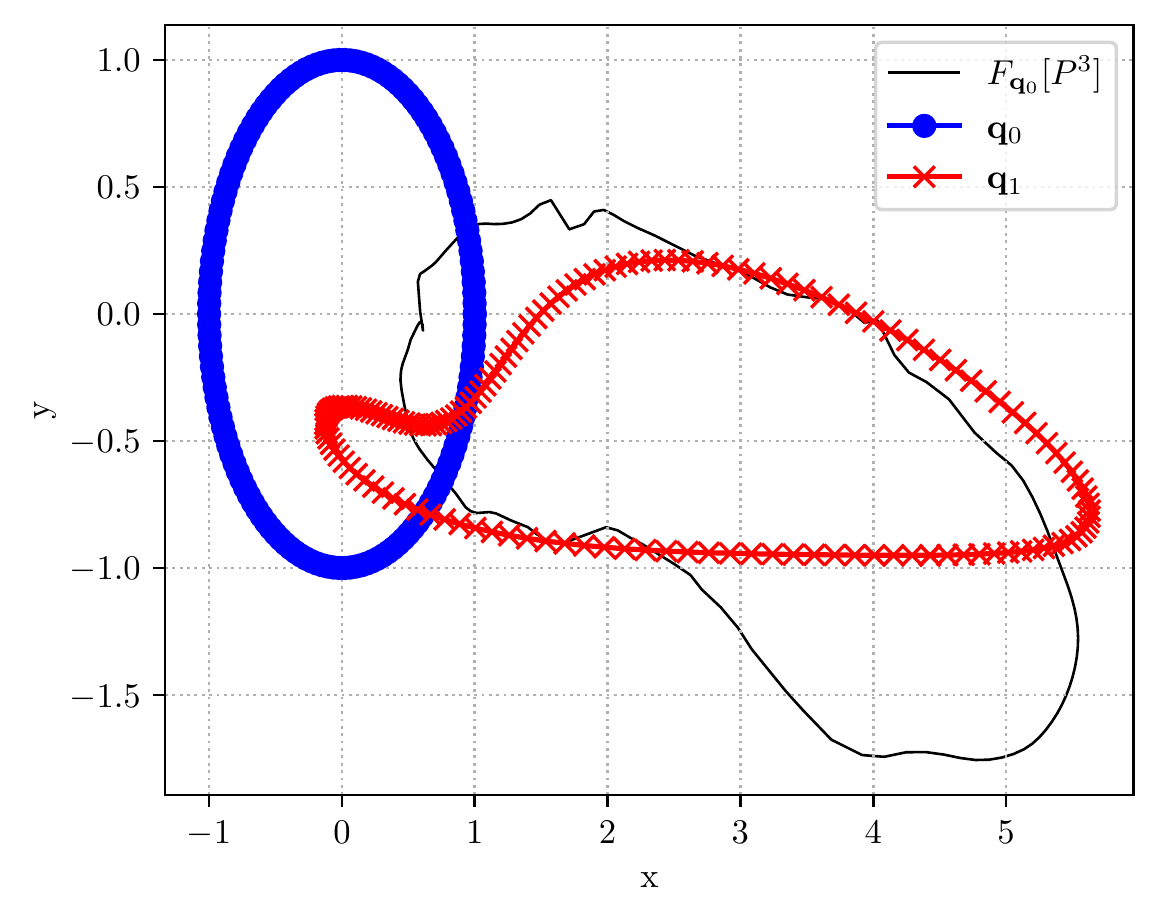}
  \includegraphics[width=.32\linewidth]{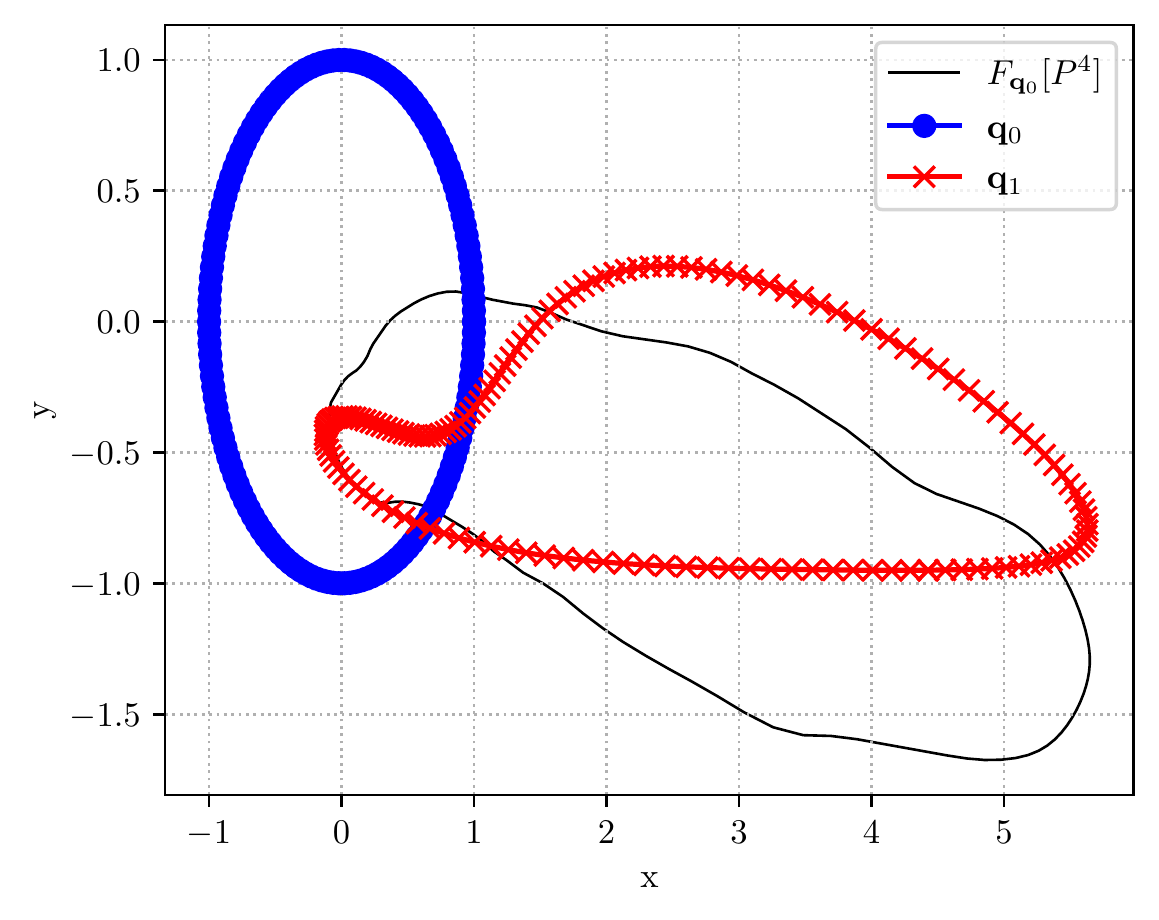}
  \includegraphics[width=.32\linewidth]{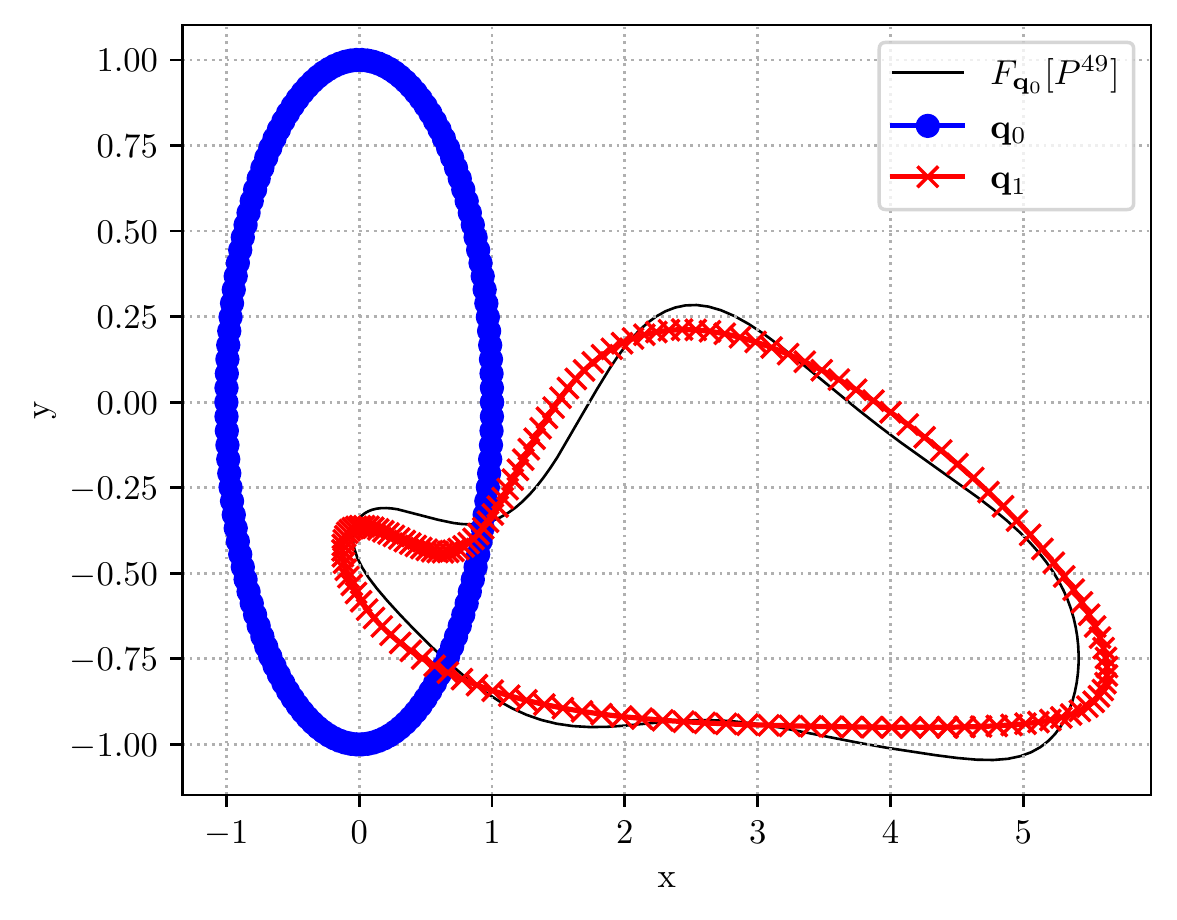}\\
  \includegraphics[width=.32\linewidth]{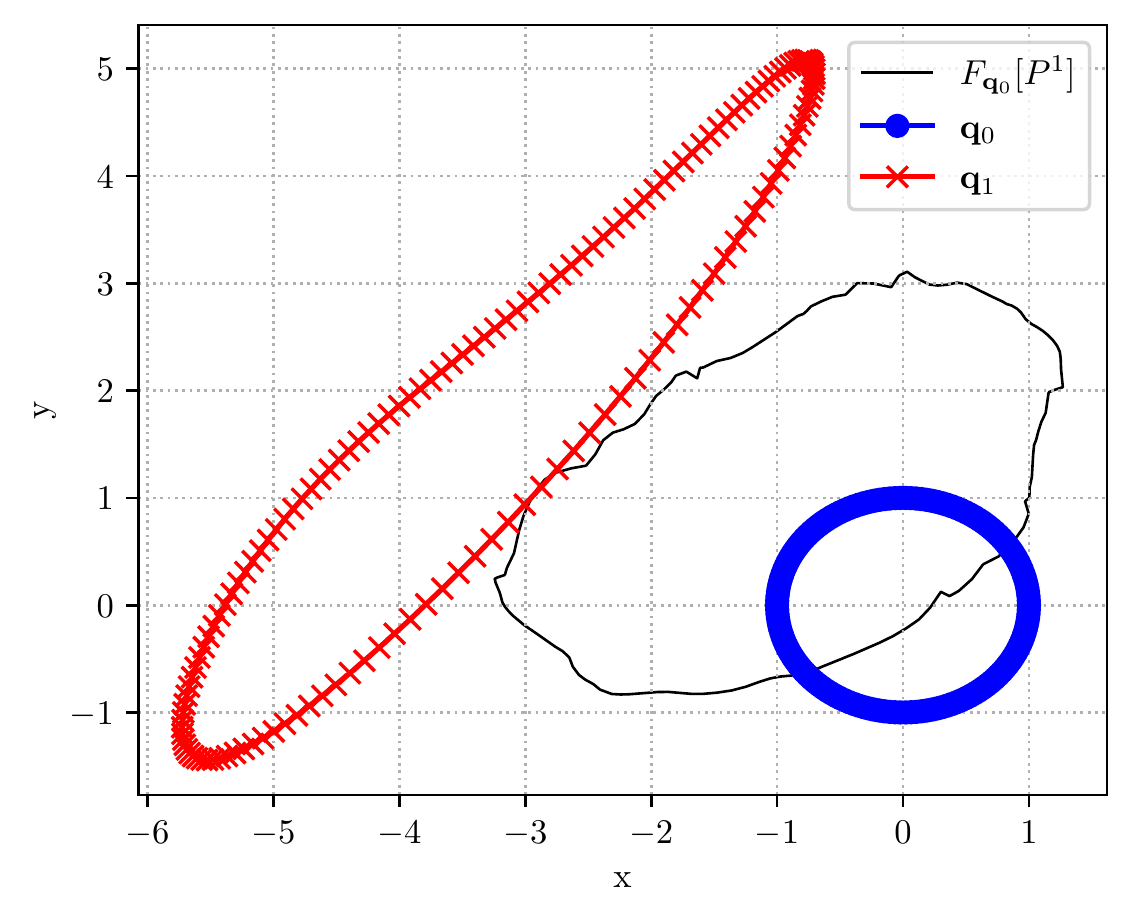}
  \includegraphics[width=.32\linewidth]{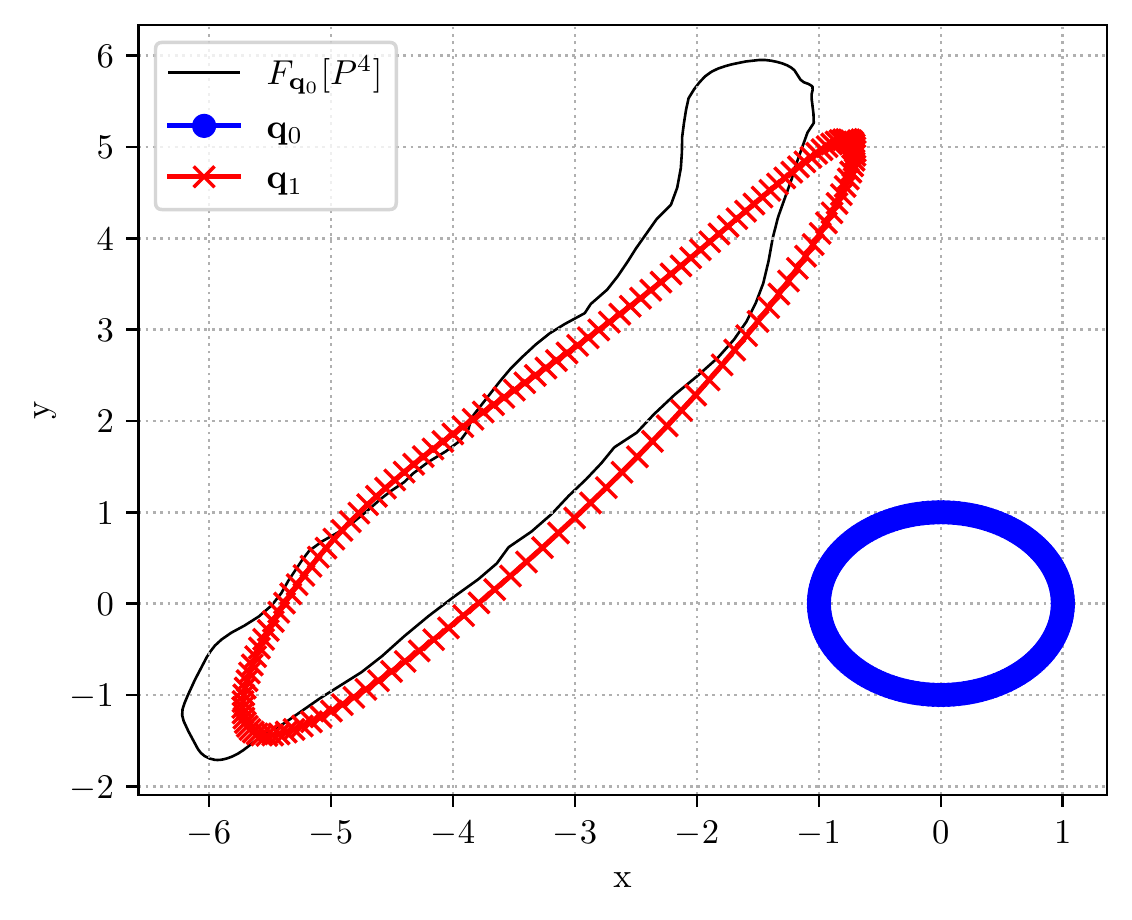}
  \includegraphics[width=.32\linewidth]{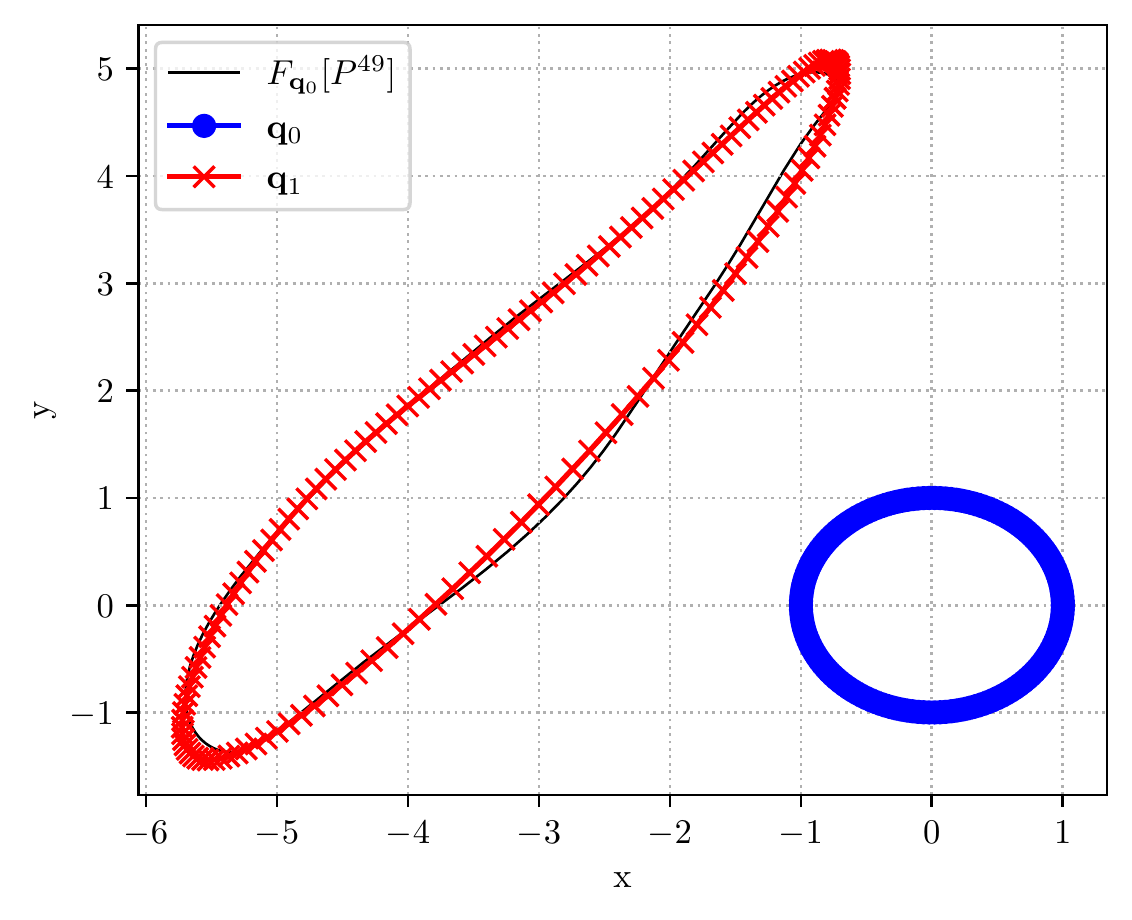}
\caption{Progression of algorithm \ref{enkfdiffeo} for various targets using $M=150$ and $N_E=100$. Computation times for 50 iterations (top to bottom): 5m22s, 5m23s, 5m23s.}\label{fig:L150}
\end{figure}

\begin{figure}[ht]\centering
  \includegraphics[width=.32\linewidth]{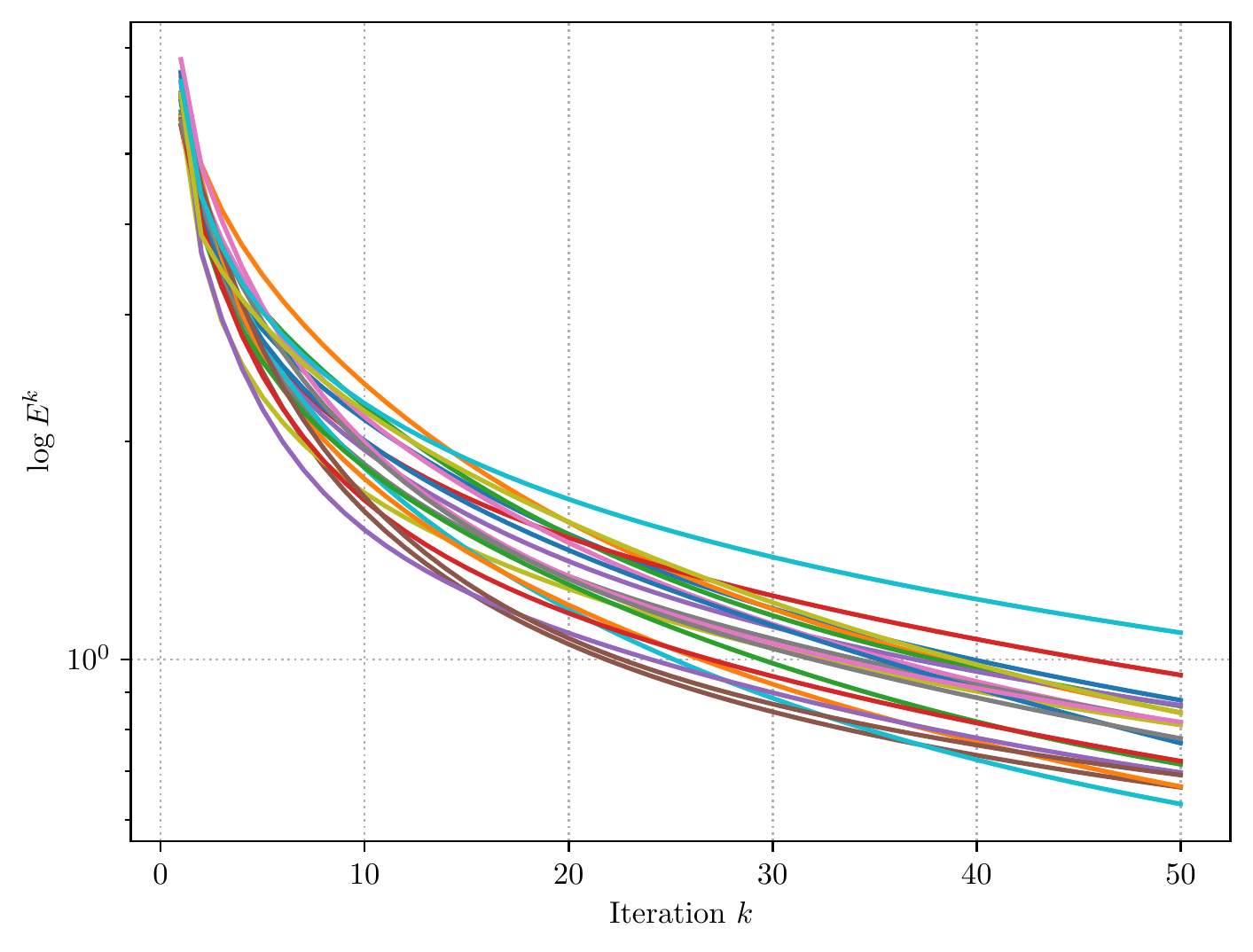}
  \includegraphics[width=.32\linewidth]{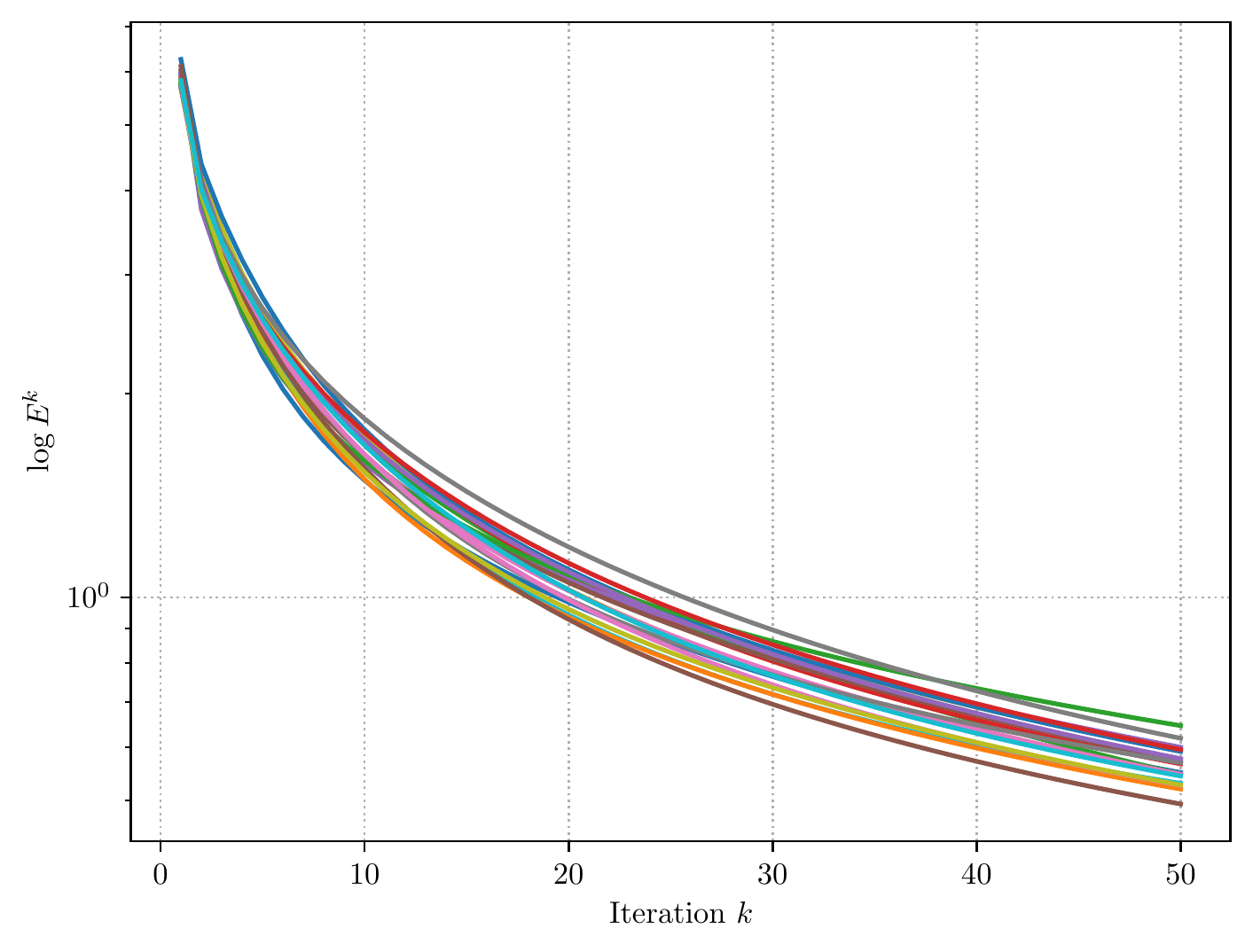}
  \includegraphics[width=.32\linewidth]{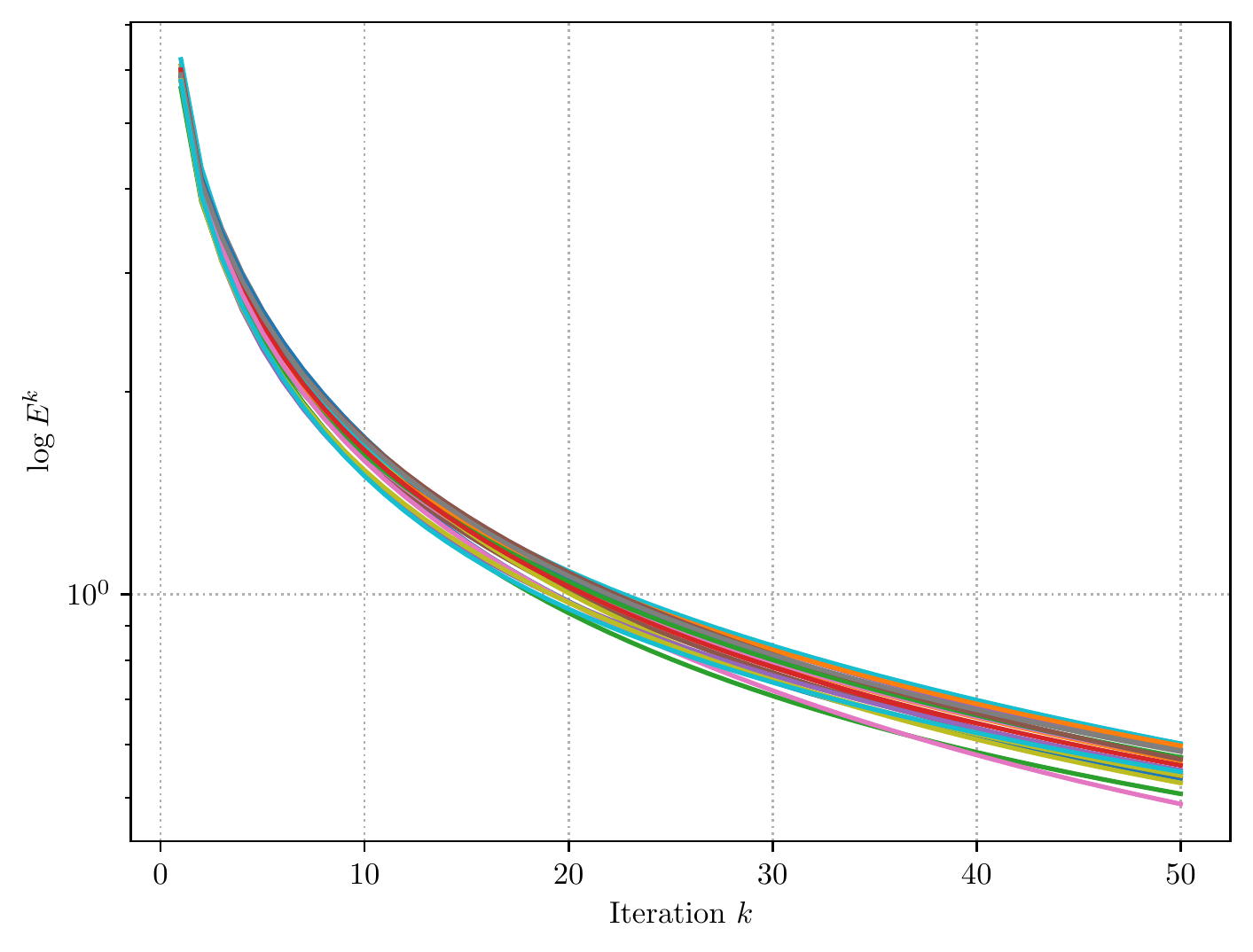}\\
  \includegraphics[width=.32\linewidth]{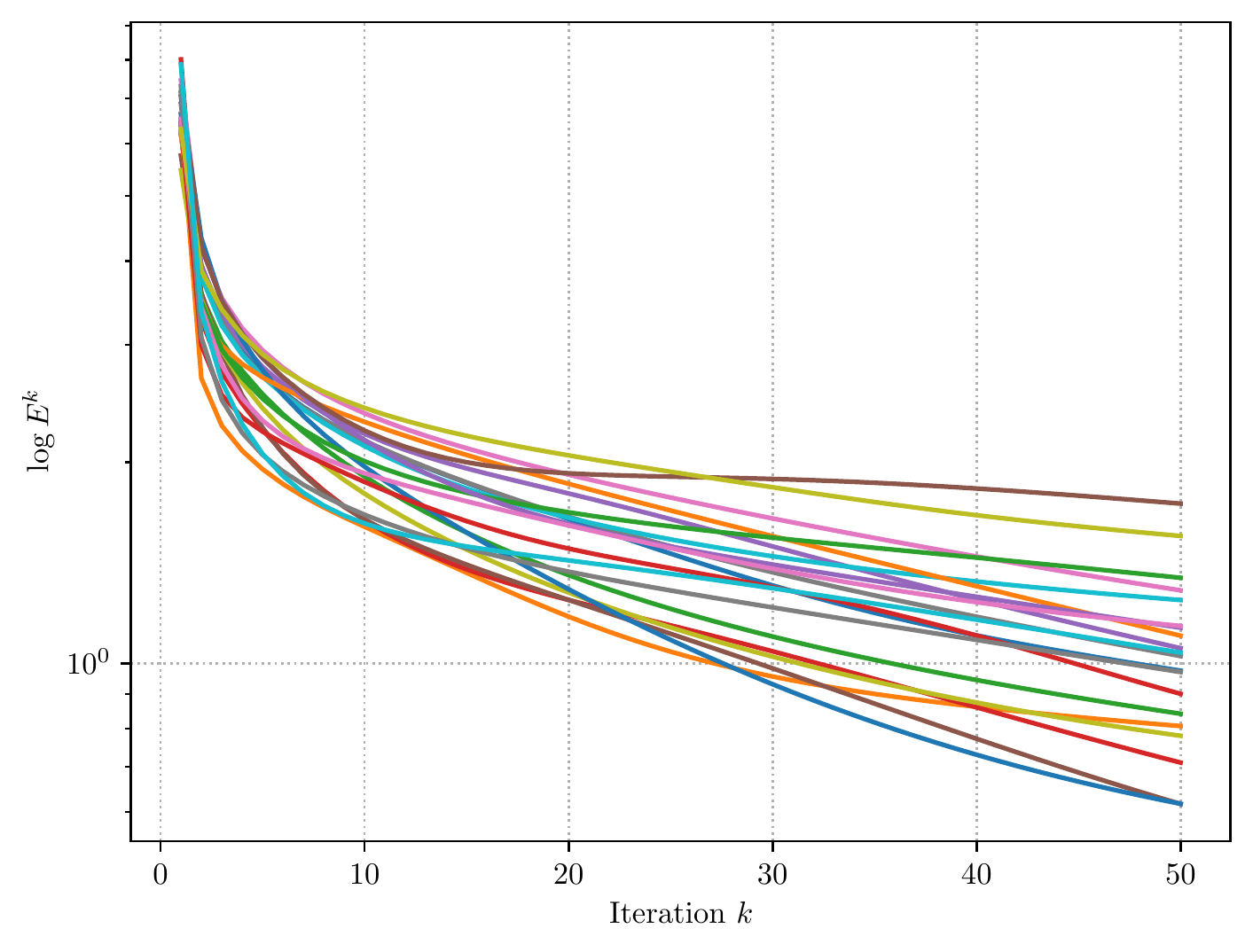}
  \includegraphics[width=.32\linewidth]{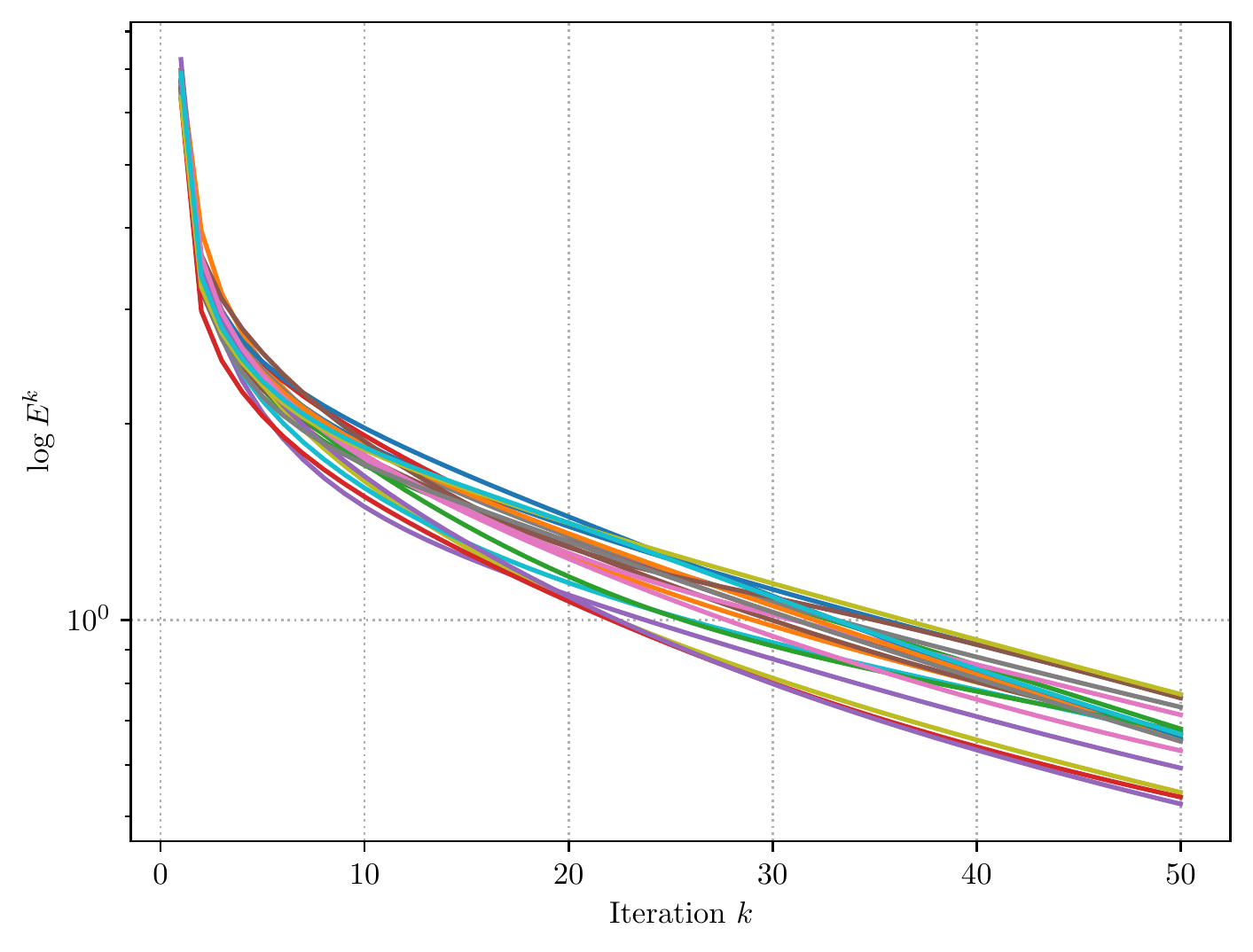}
  \includegraphics[width=.32\linewidth]{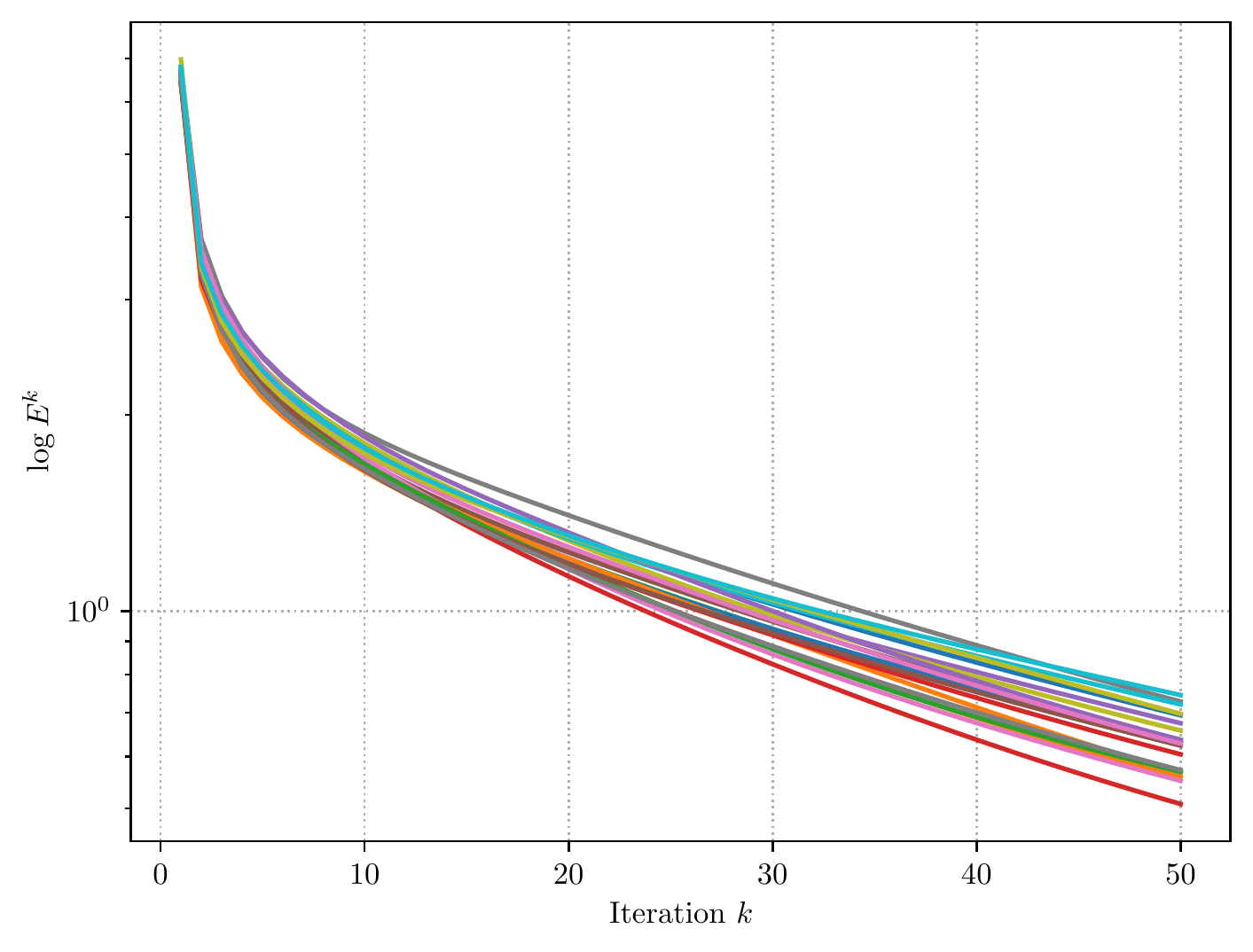}\\
  \includegraphics[width=.32\linewidth]{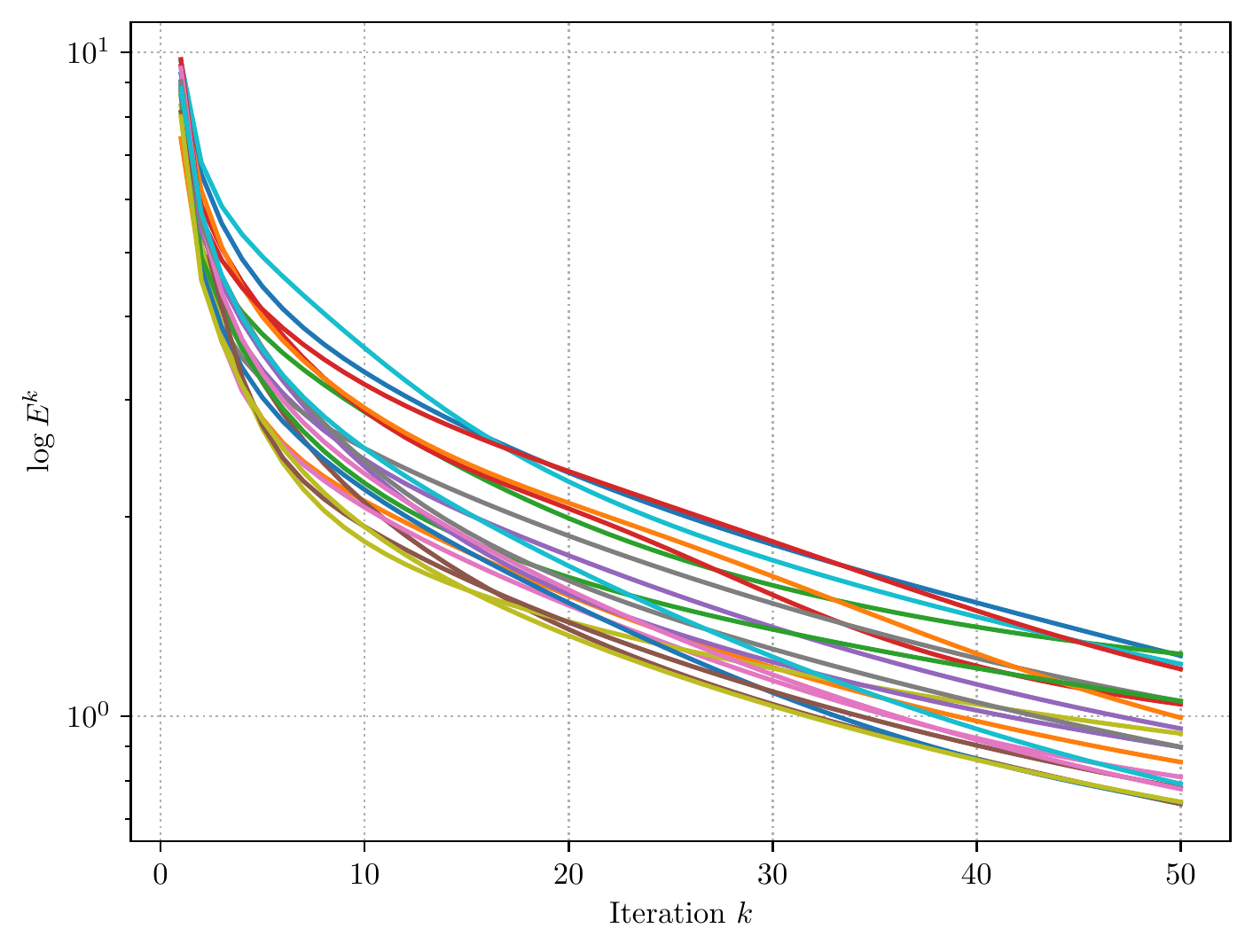}
  \includegraphics[width=.32\linewidth]{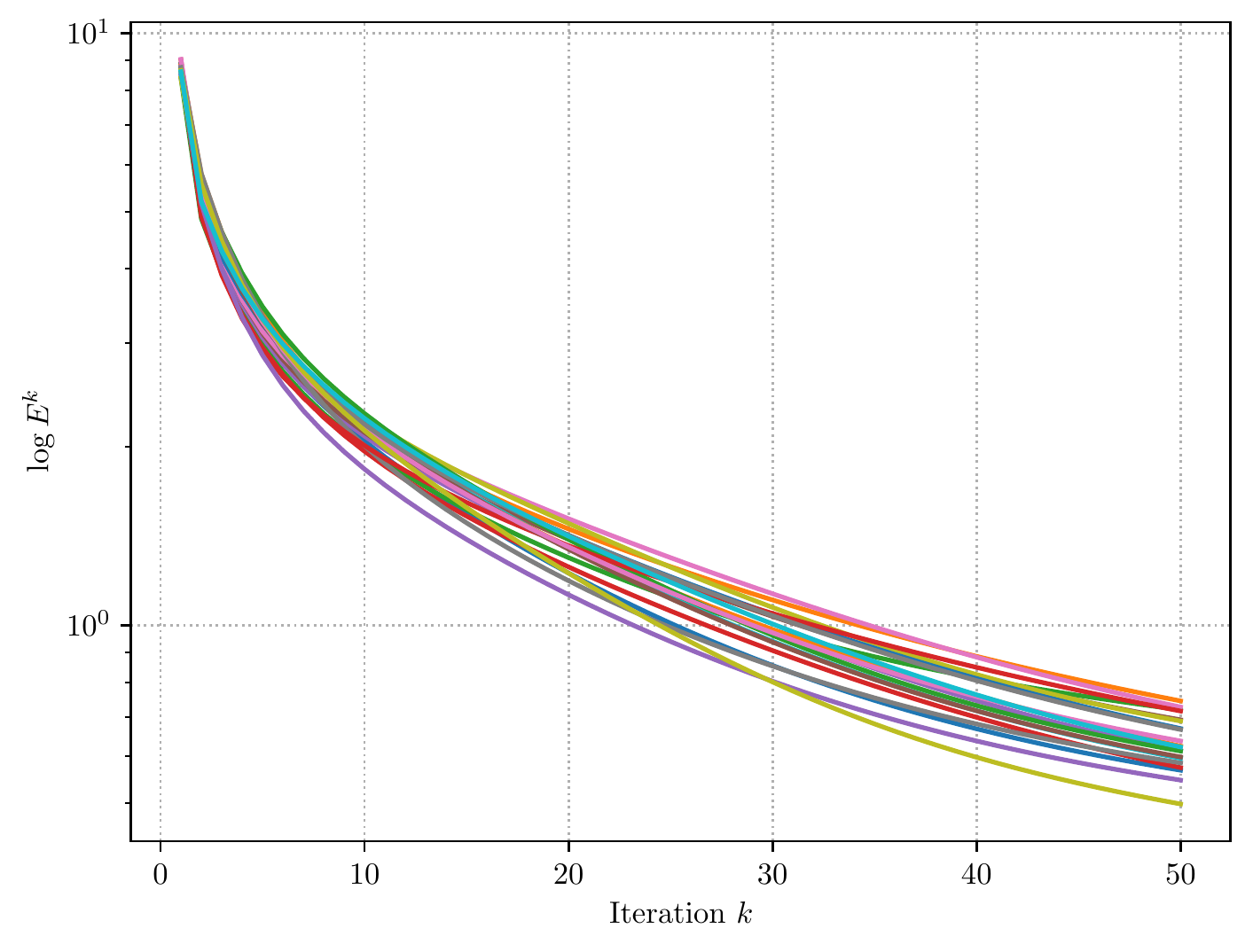}
  \includegraphics[width=.32\linewidth]{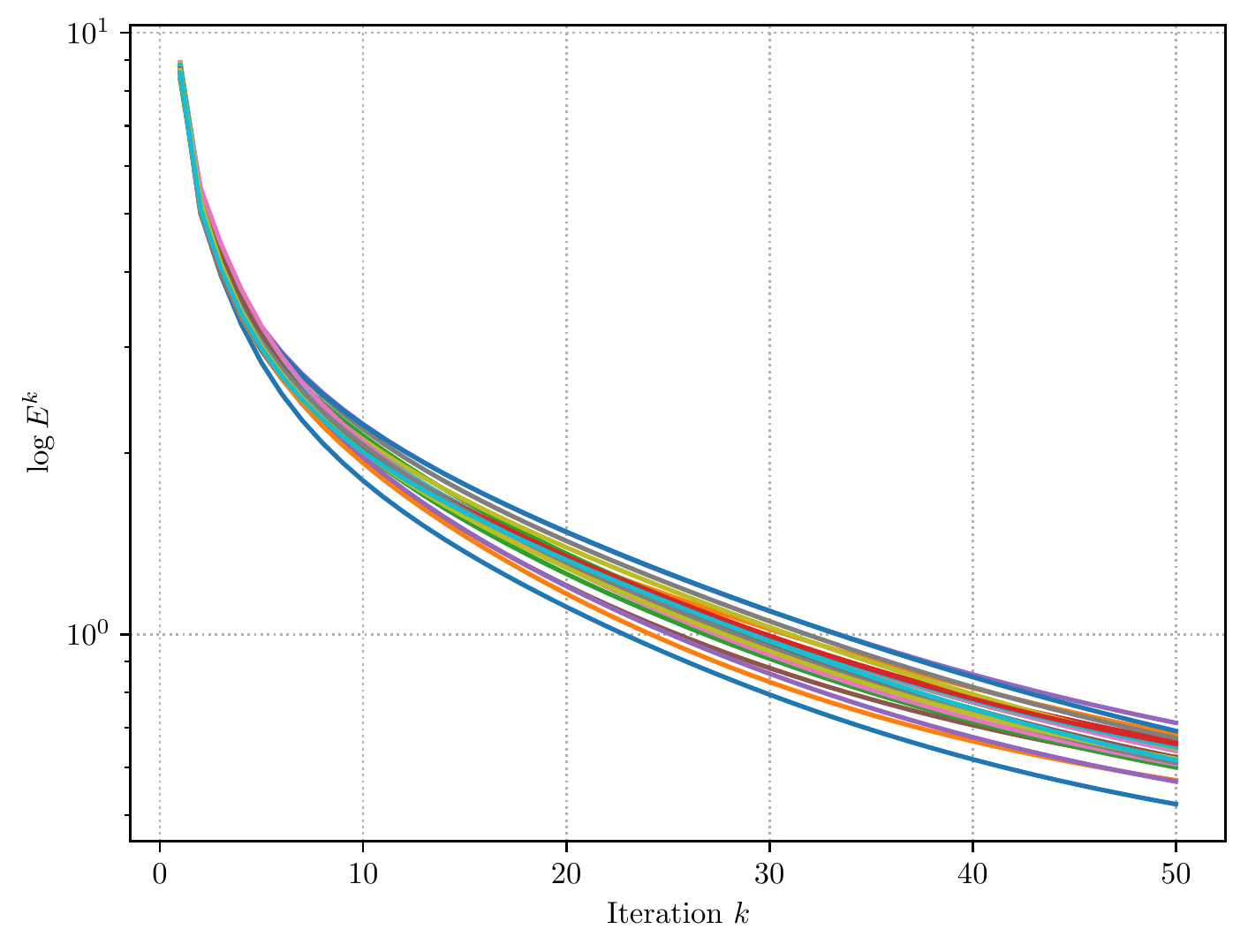}
\caption{Convergence of $E^k$ for $M=10$. Each row corresponds to a different target, with each column representing $N_E=10, 50, 100$. Each figure shows the log data misfit as a function of Kalman iteration for 20 random draws of initial ensemble.}
\label{fig:ex1}
\end{figure}

\begin{figure}[ht]\centering
  \includegraphics[width=.32\linewidth]{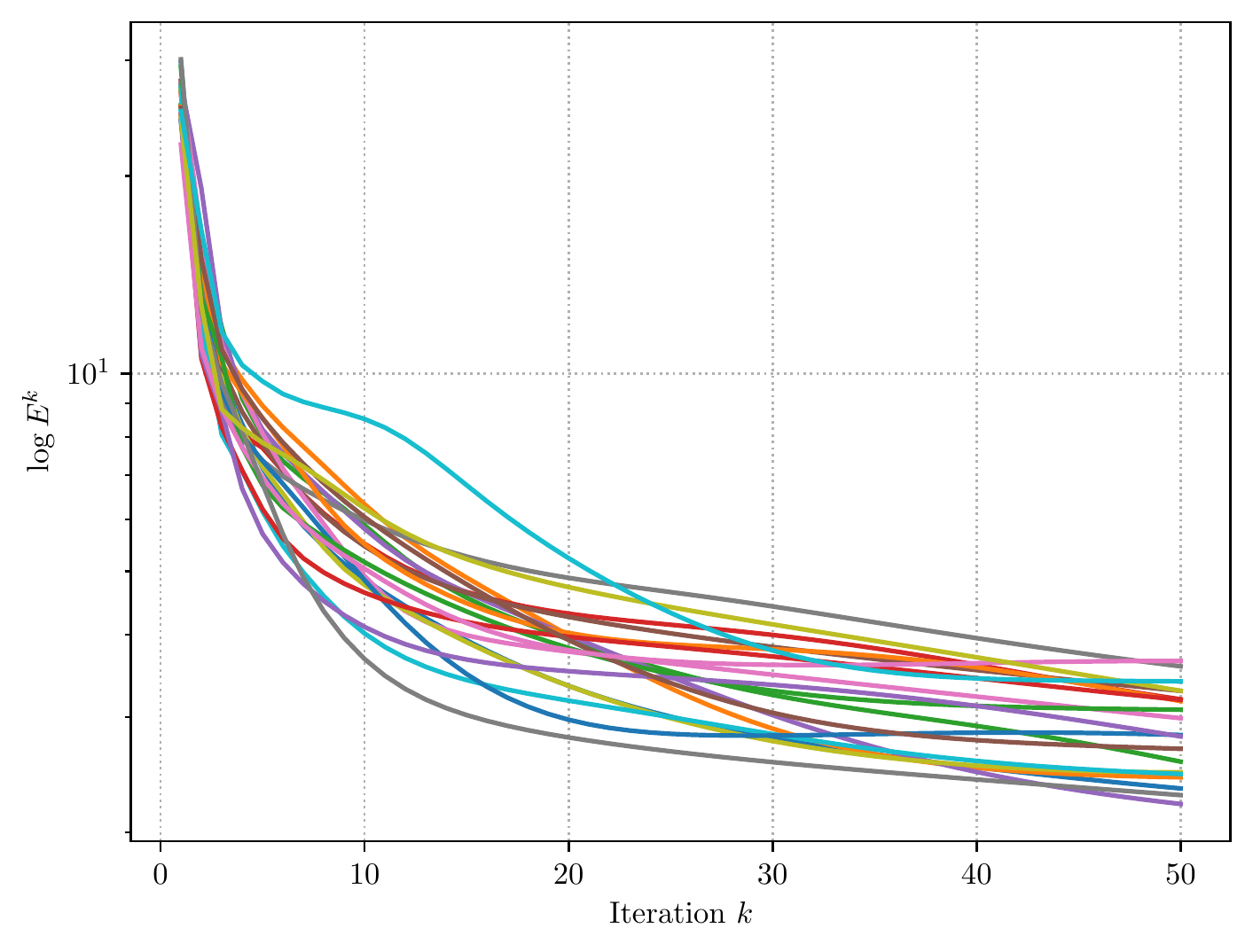}
  \includegraphics[width=.32\linewidth]{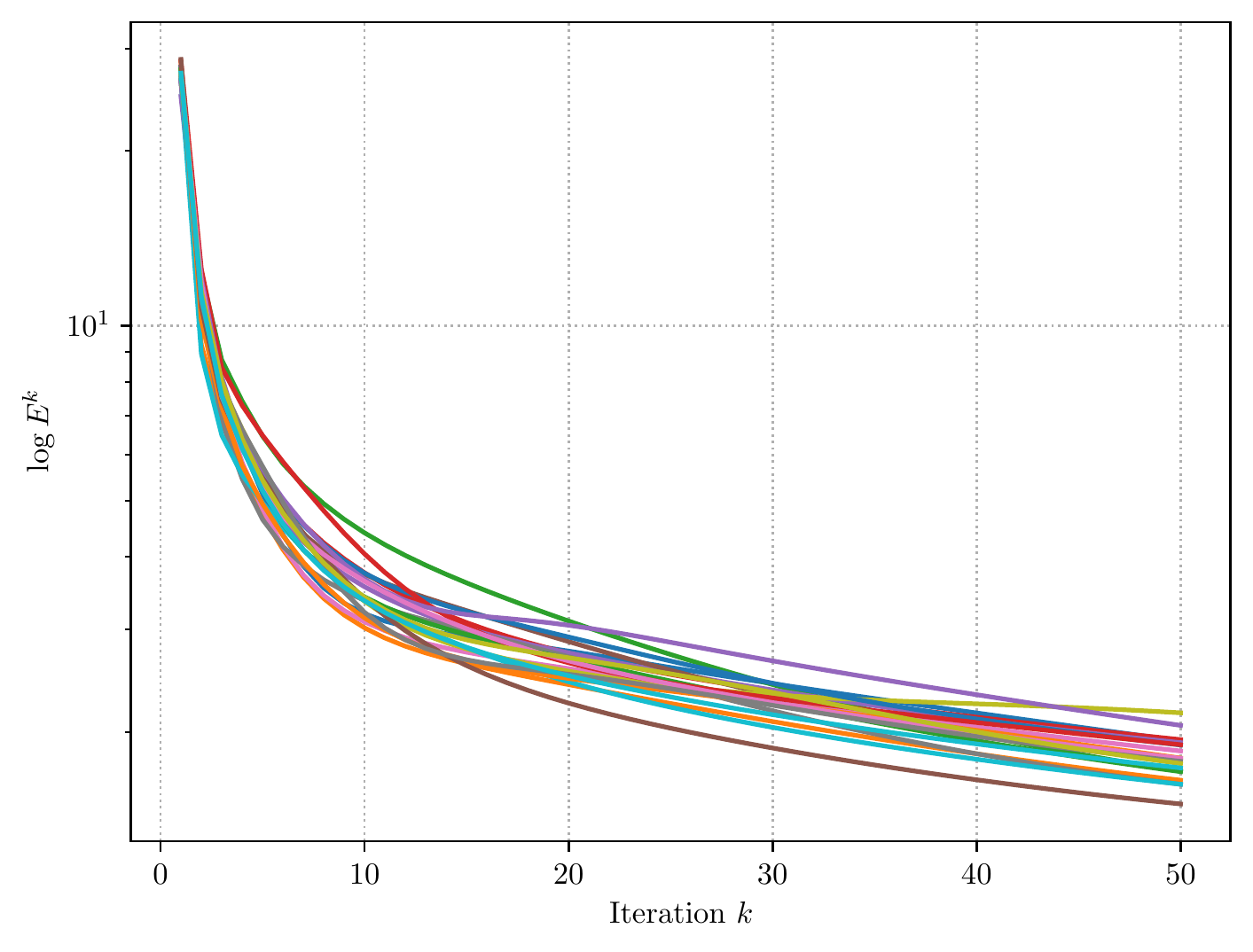}
  \includegraphics[width=.32\linewidth]{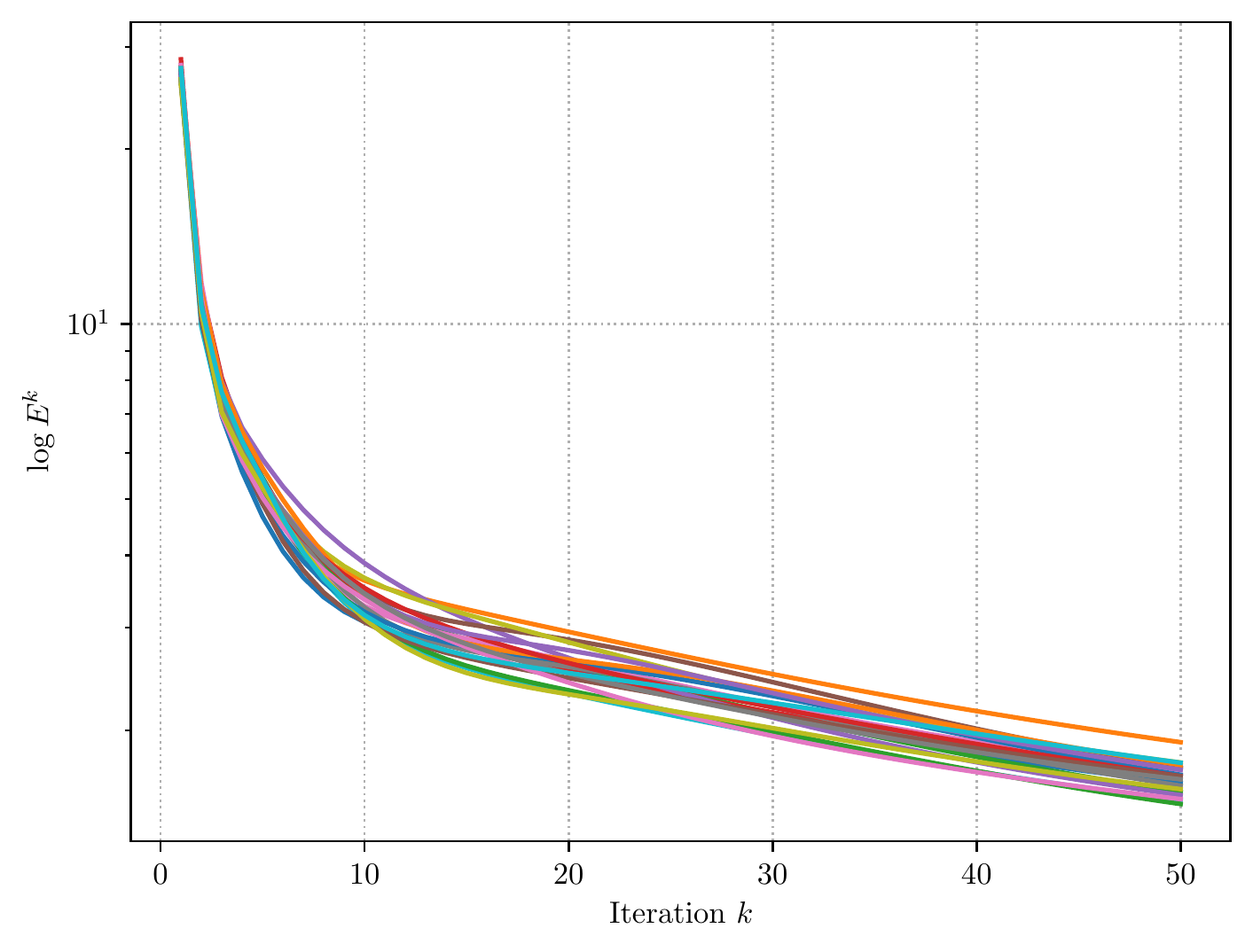}\\
  \includegraphics[width=.32\linewidth]{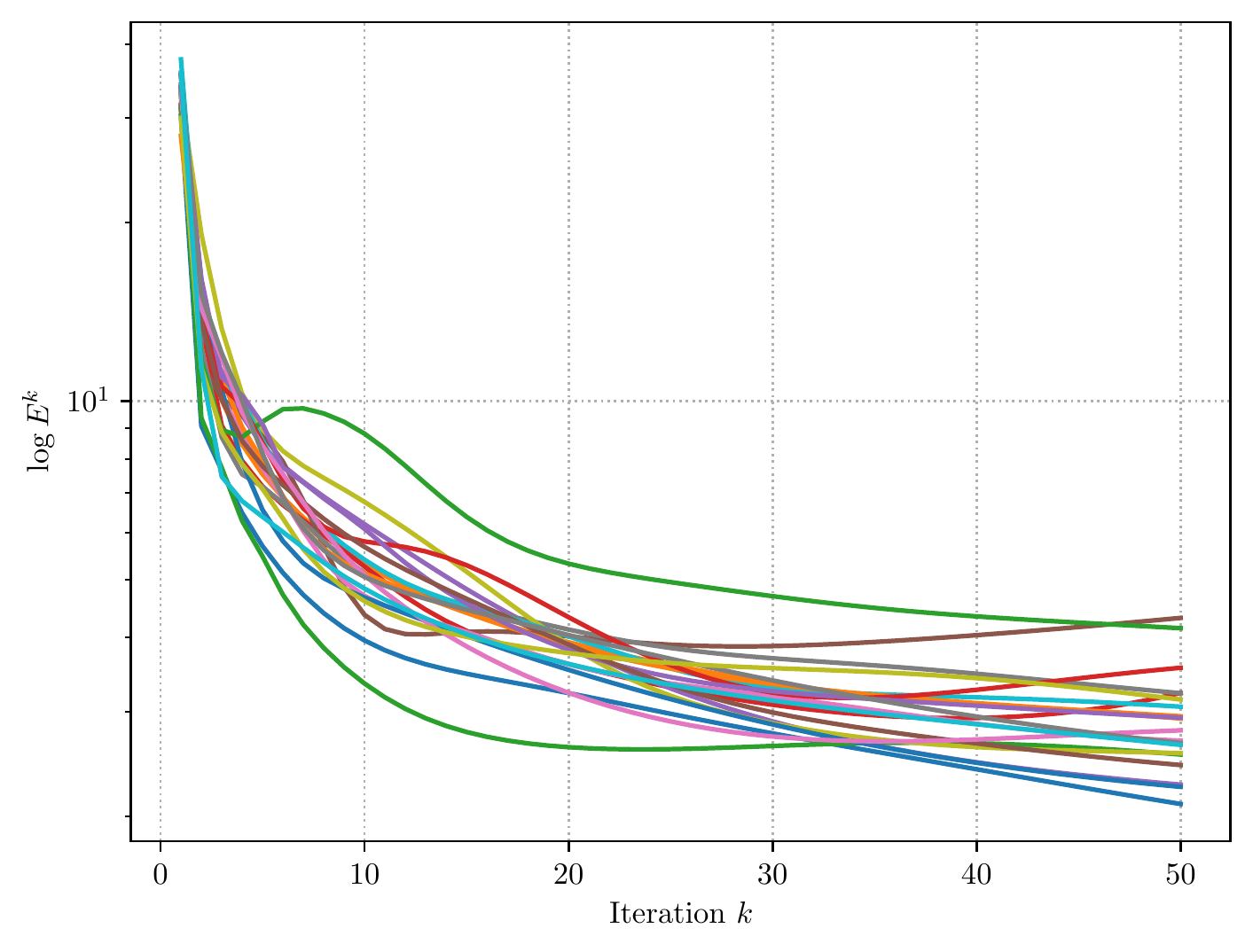}
  \includegraphics[width=.32\linewidth]{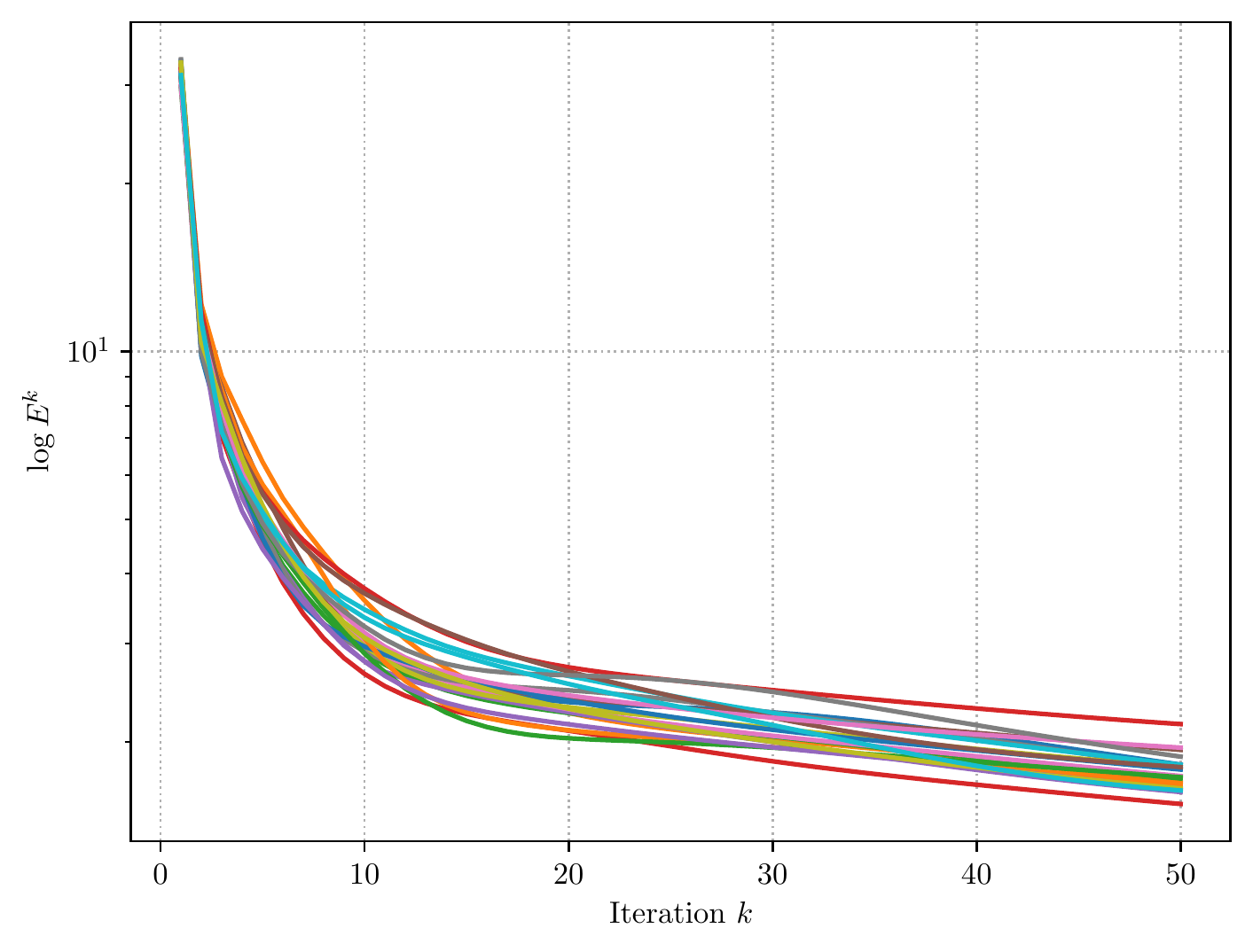}
  \includegraphics[width=.32\linewidth]{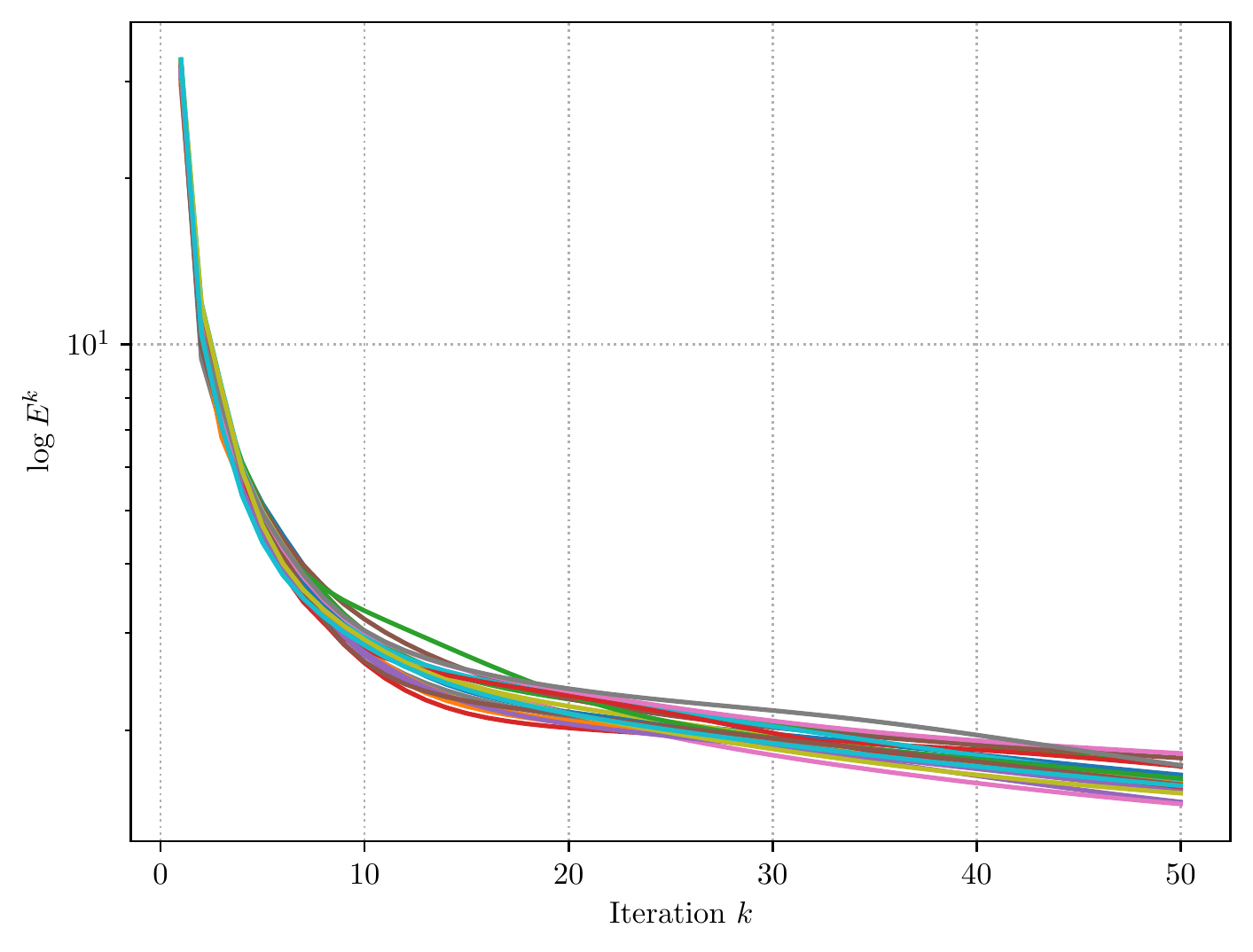}\\
  \includegraphics[width=.32\linewidth]{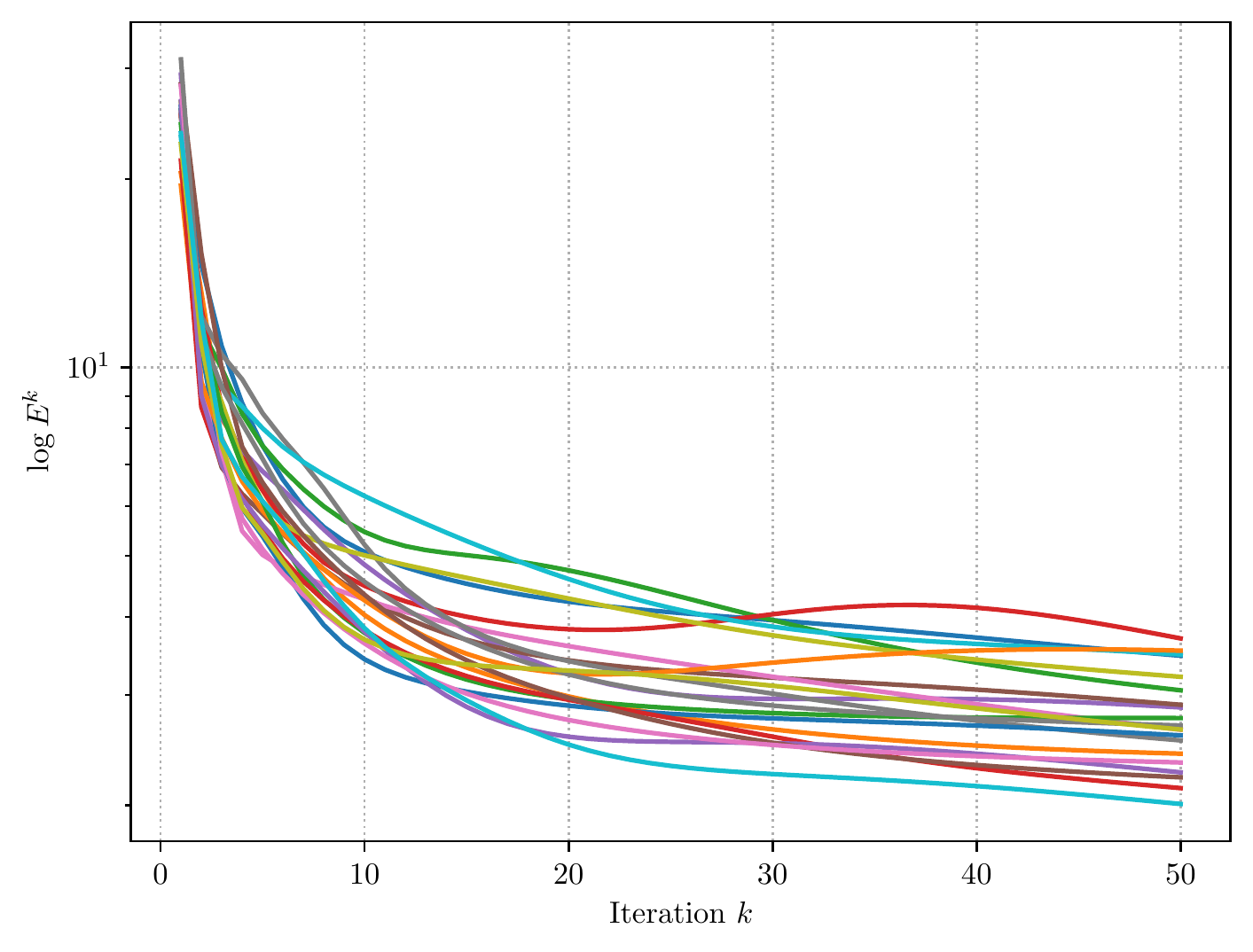}
  \includegraphics[width=.32\linewidth]{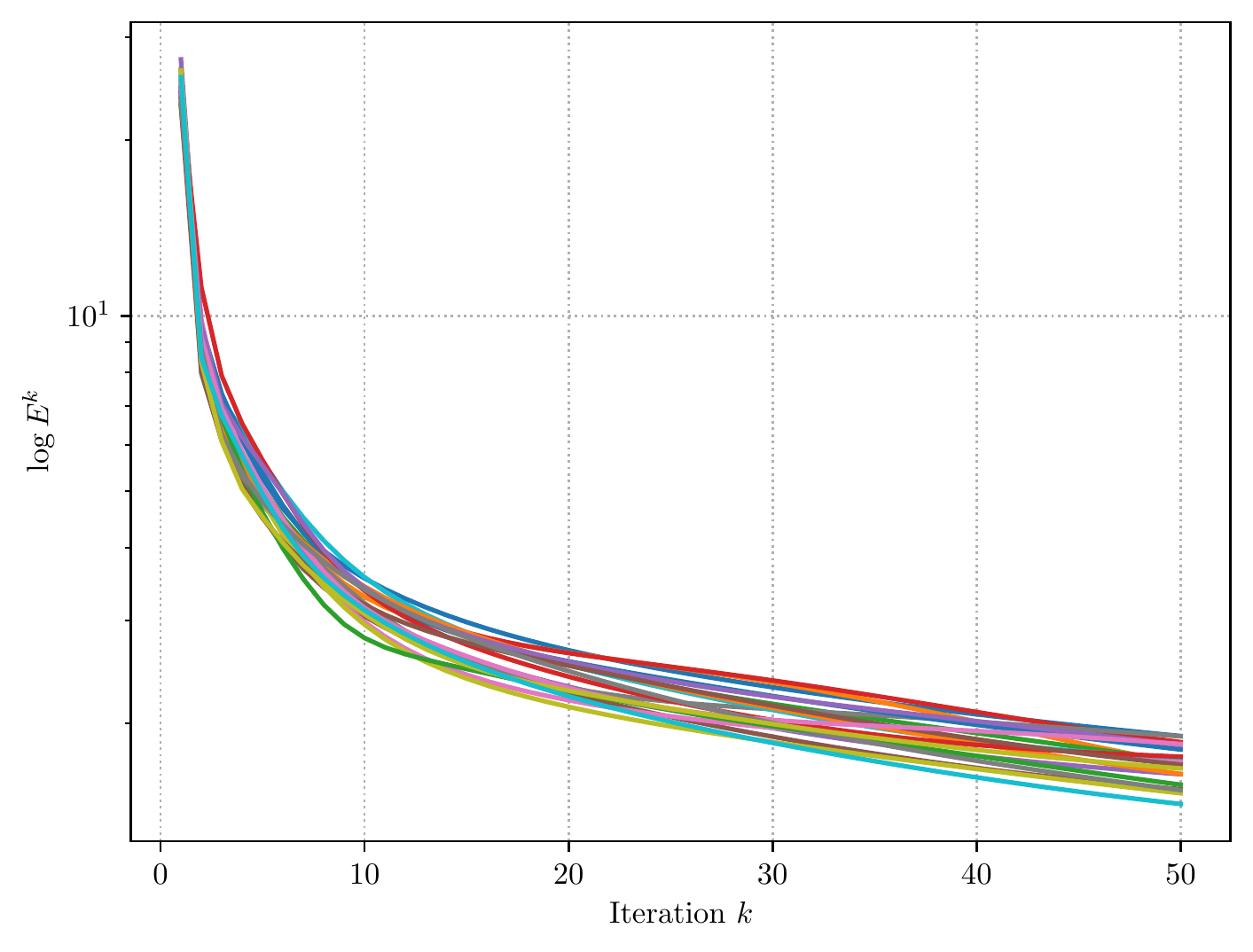}
  \includegraphics[width=.32\linewidth]{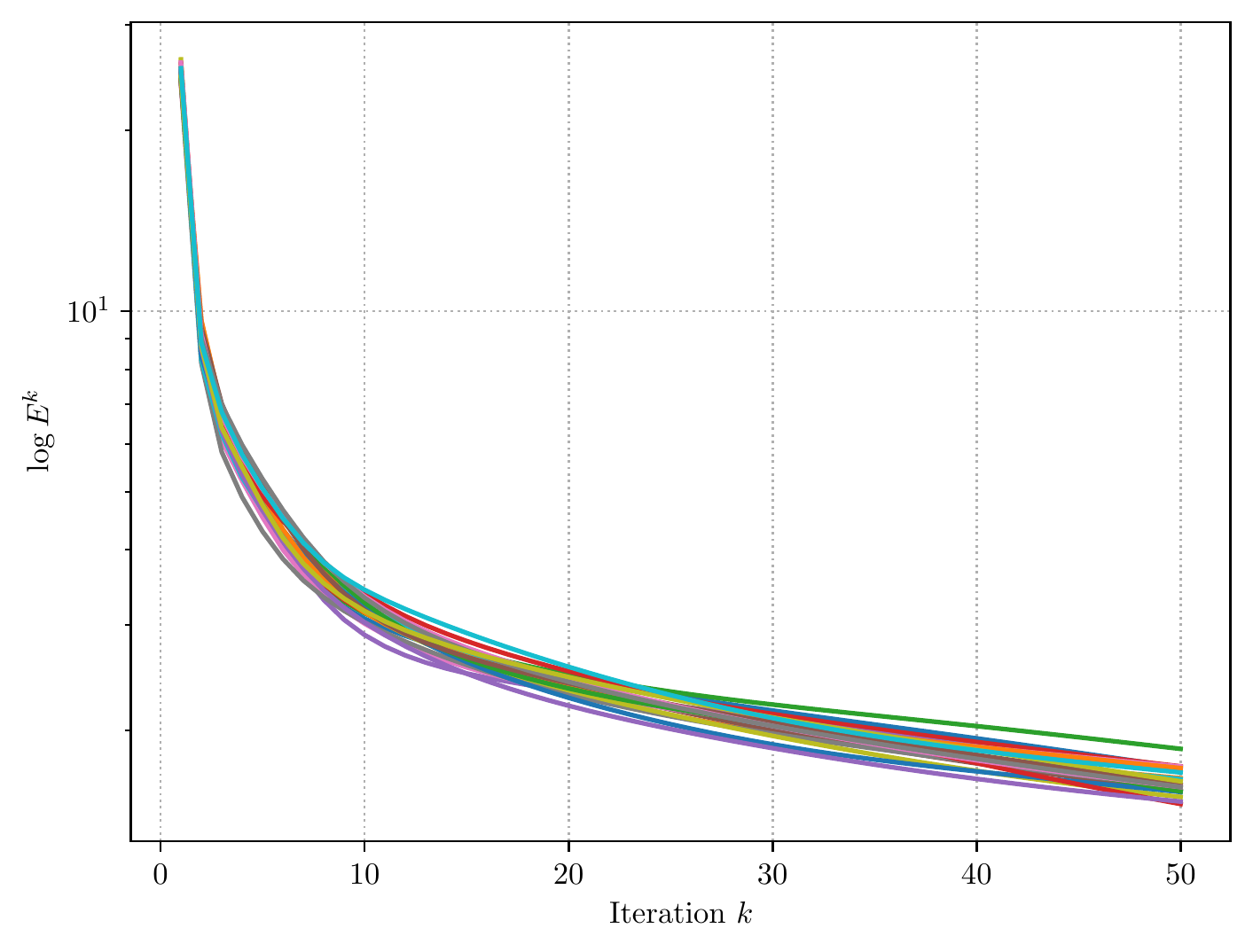}
\caption{Convergence of $E^k$ for $M=50$. Each row corresponds to a different target, with each column representing $N_E=10, 50, 100$. Each figure shows the log data misfit as a function of Kalman iteration for 20 random draws of initial ensemble.}
\label{fig:ex2}
\end{figure}

\begin{figure}[ht]\centering
  \includegraphics[width=.32\linewidth]{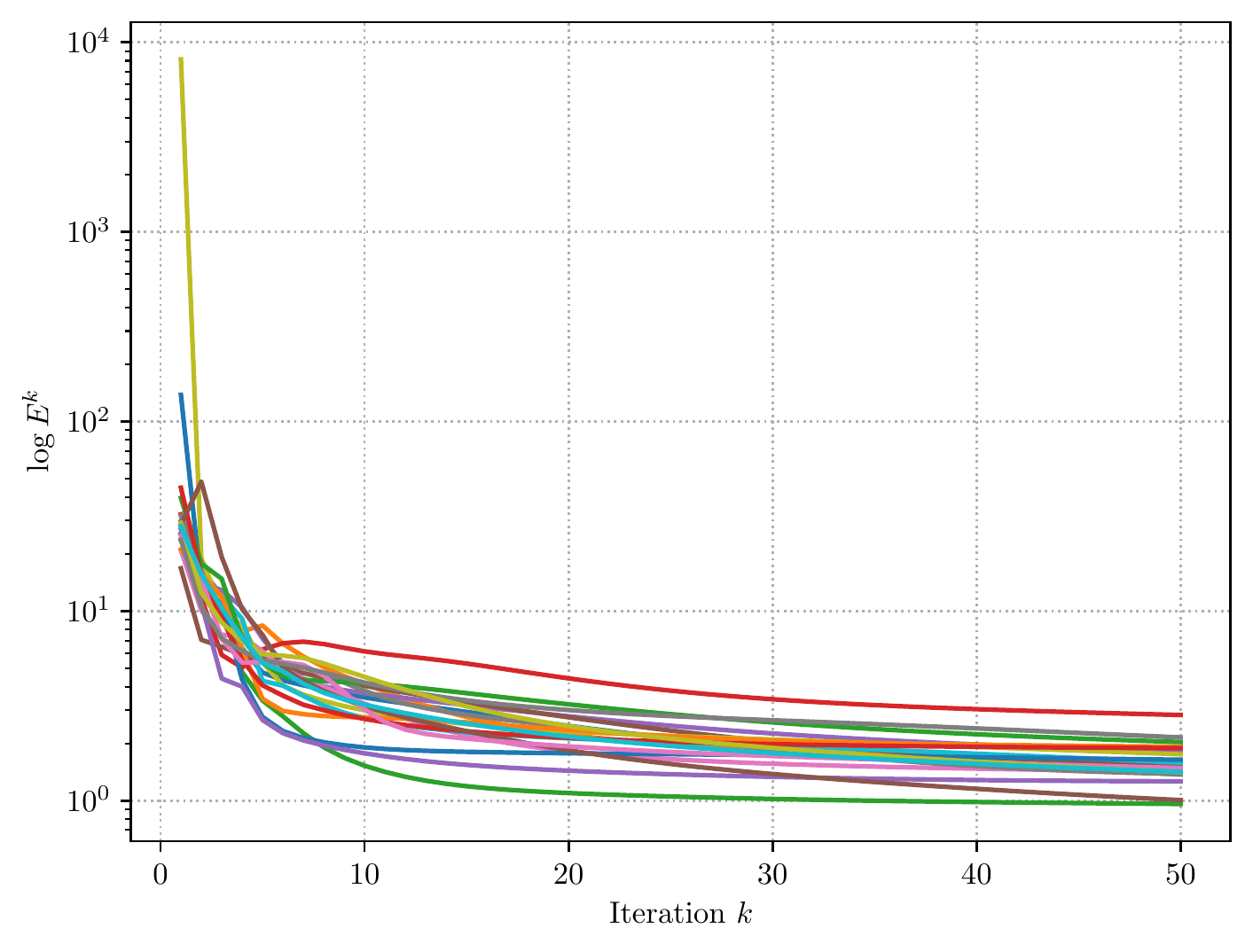}
  \includegraphics[width=.32\linewidth]{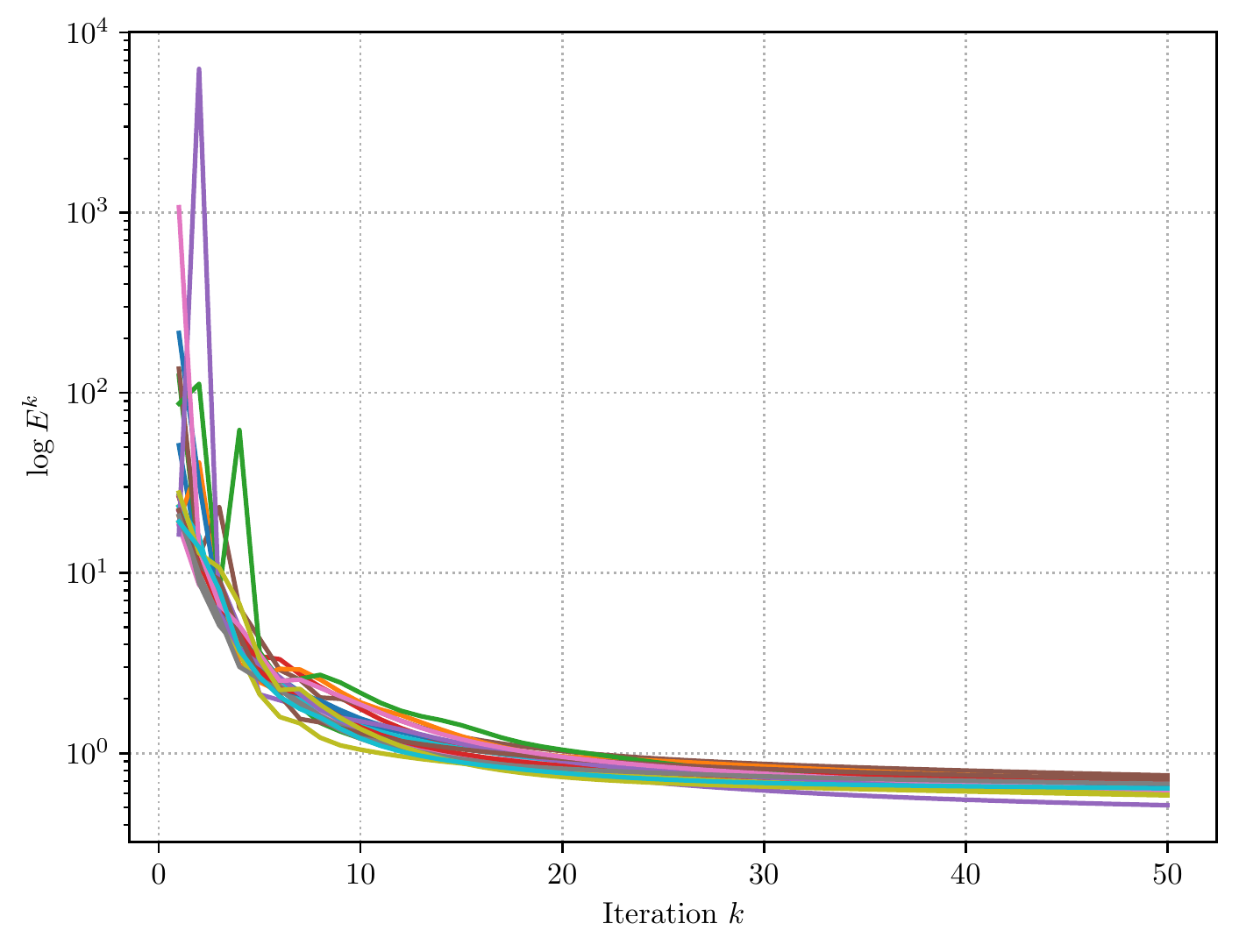}
  \includegraphics[width=.32\linewidth]{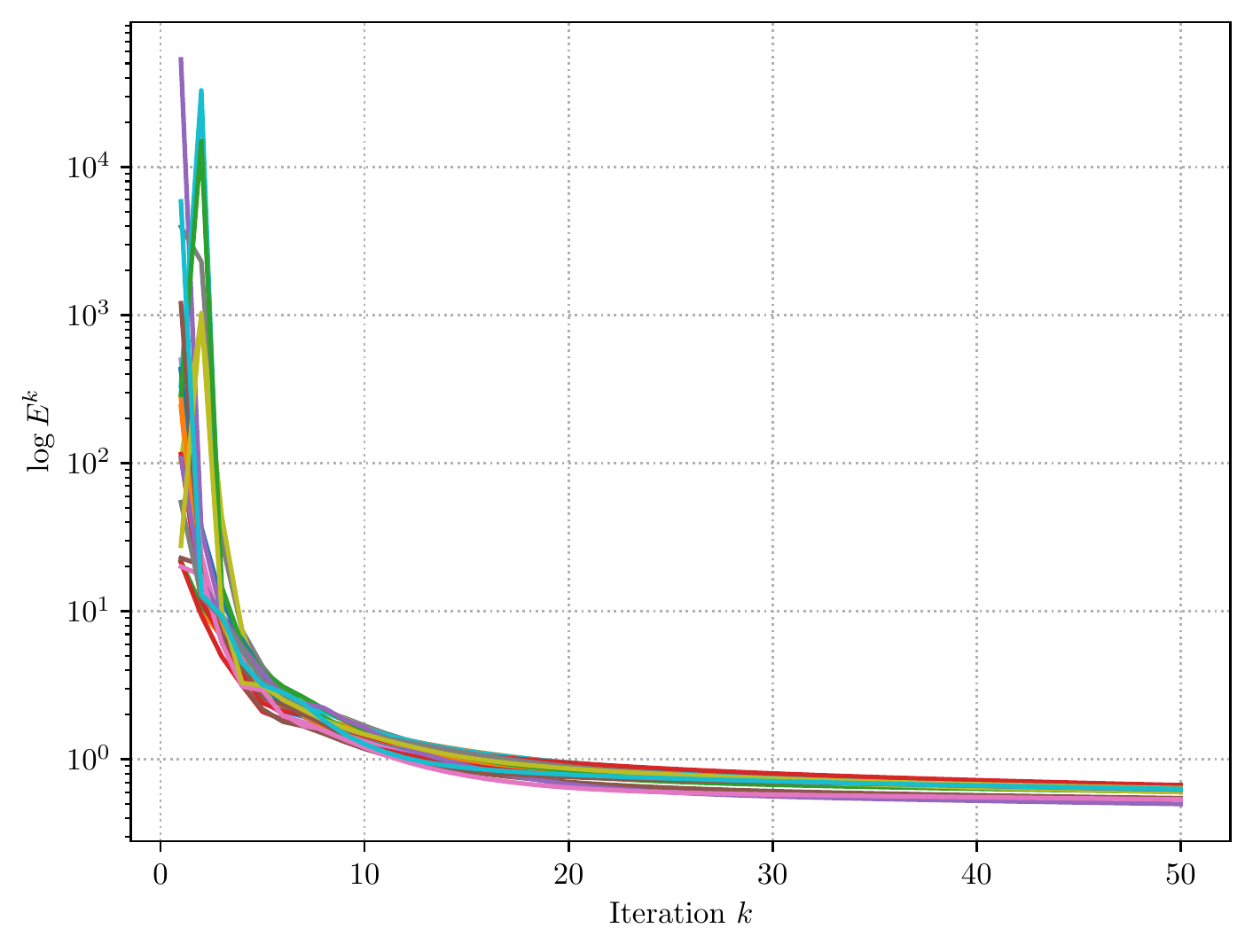}\\
  \includegraphics[width=.32\linewidth]{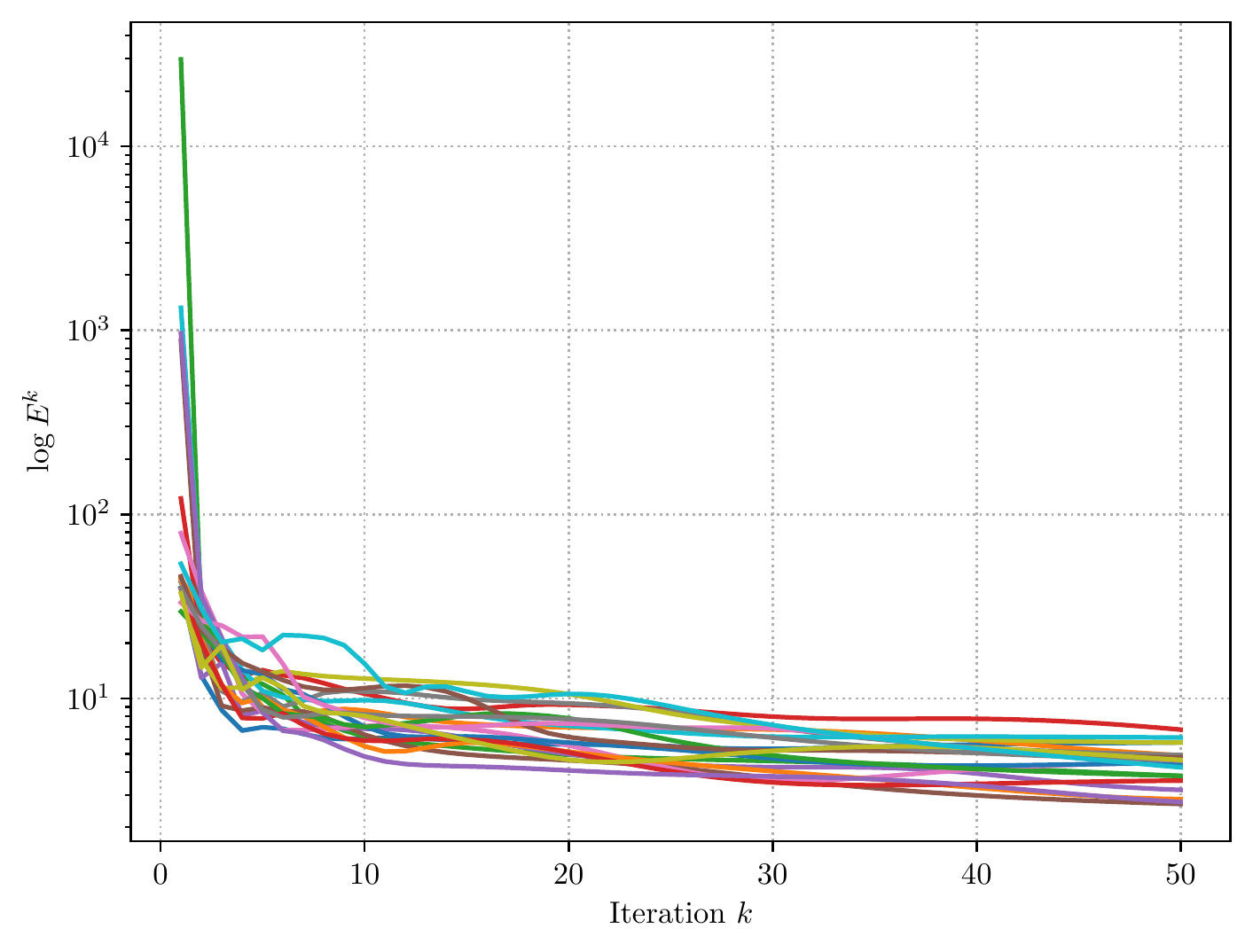}
  \includegraphics[width=.32\linewidth]{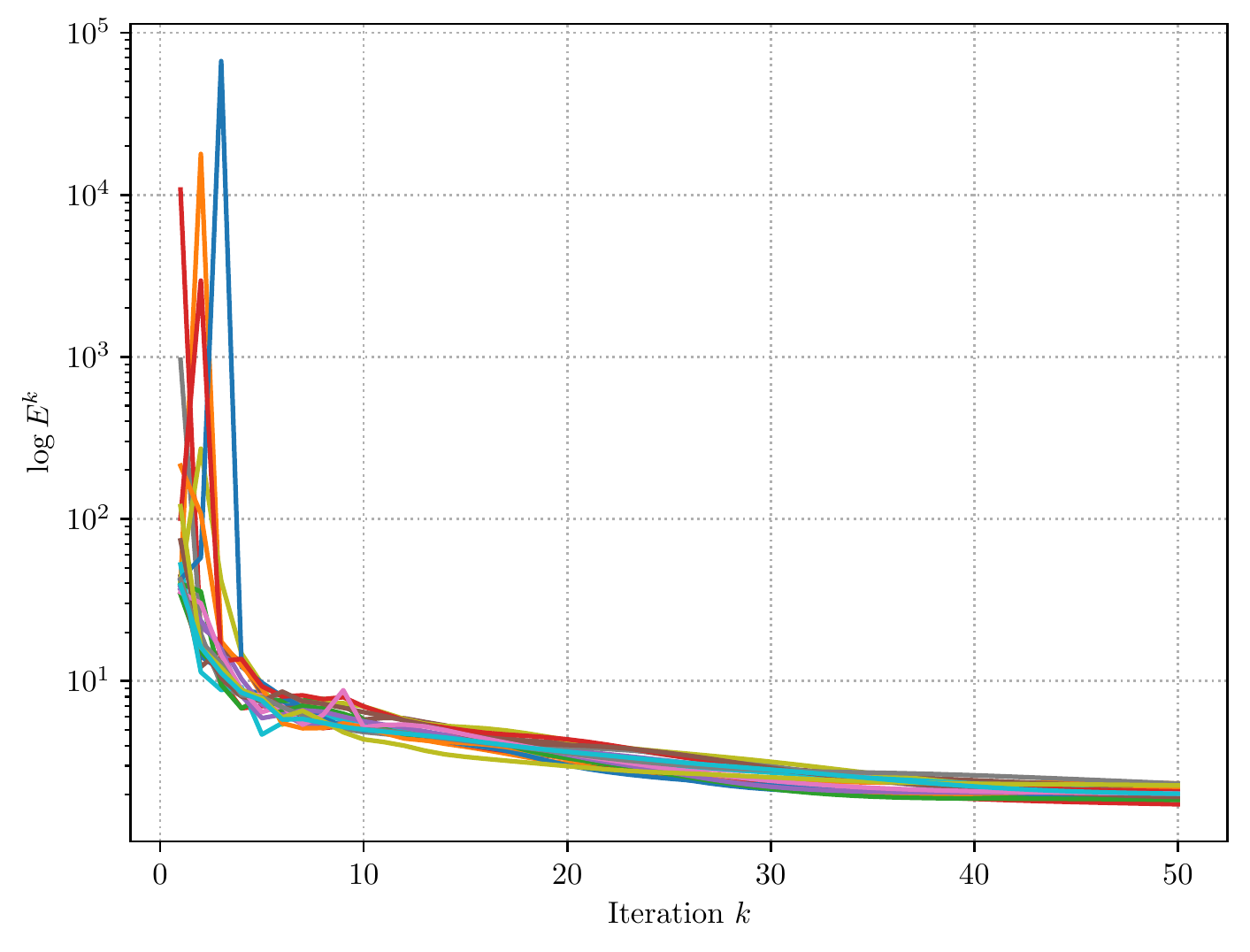}
  \includegraphics[width=.32\linewidth]{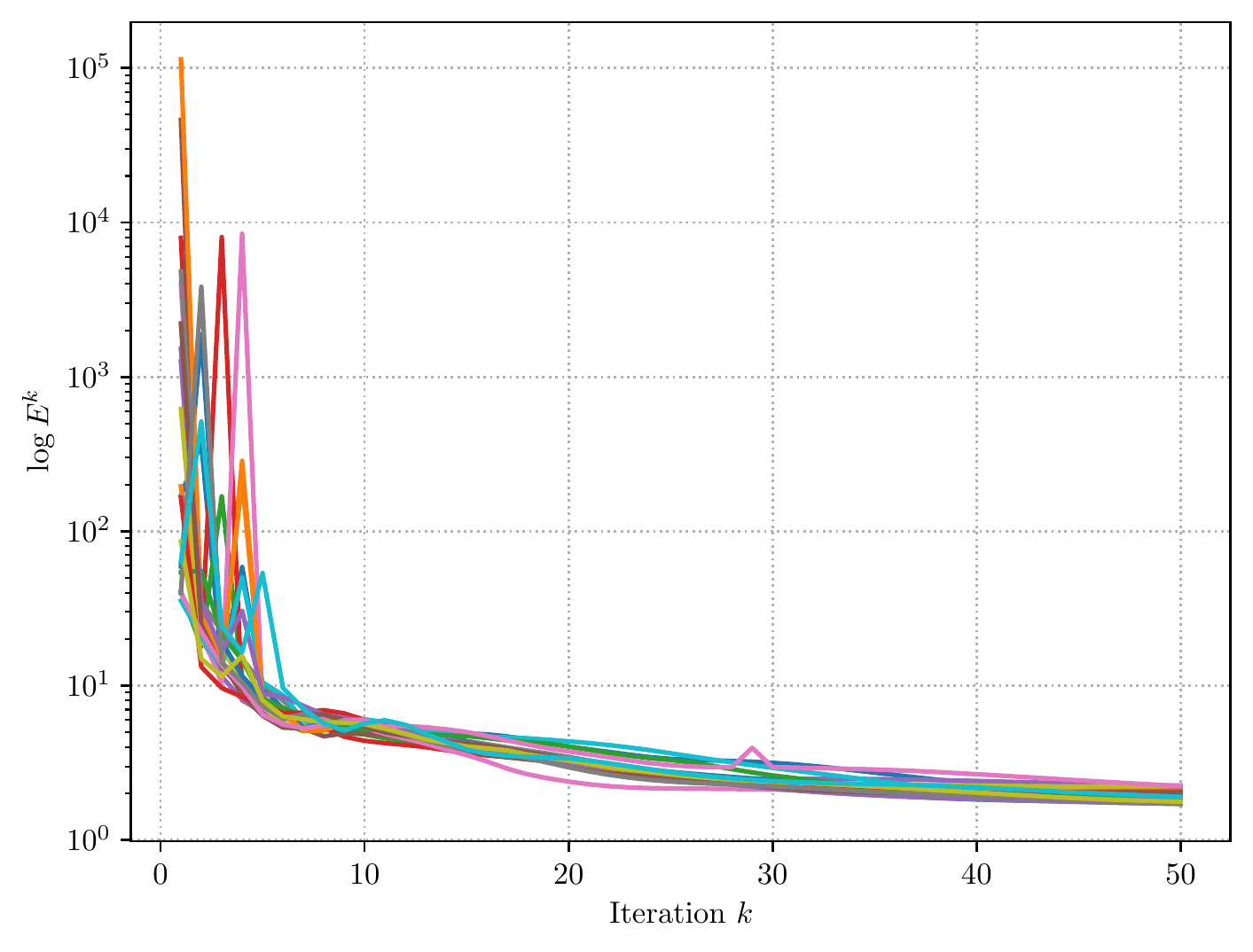}\\
  \includegraphics[width=.32\linewidth]{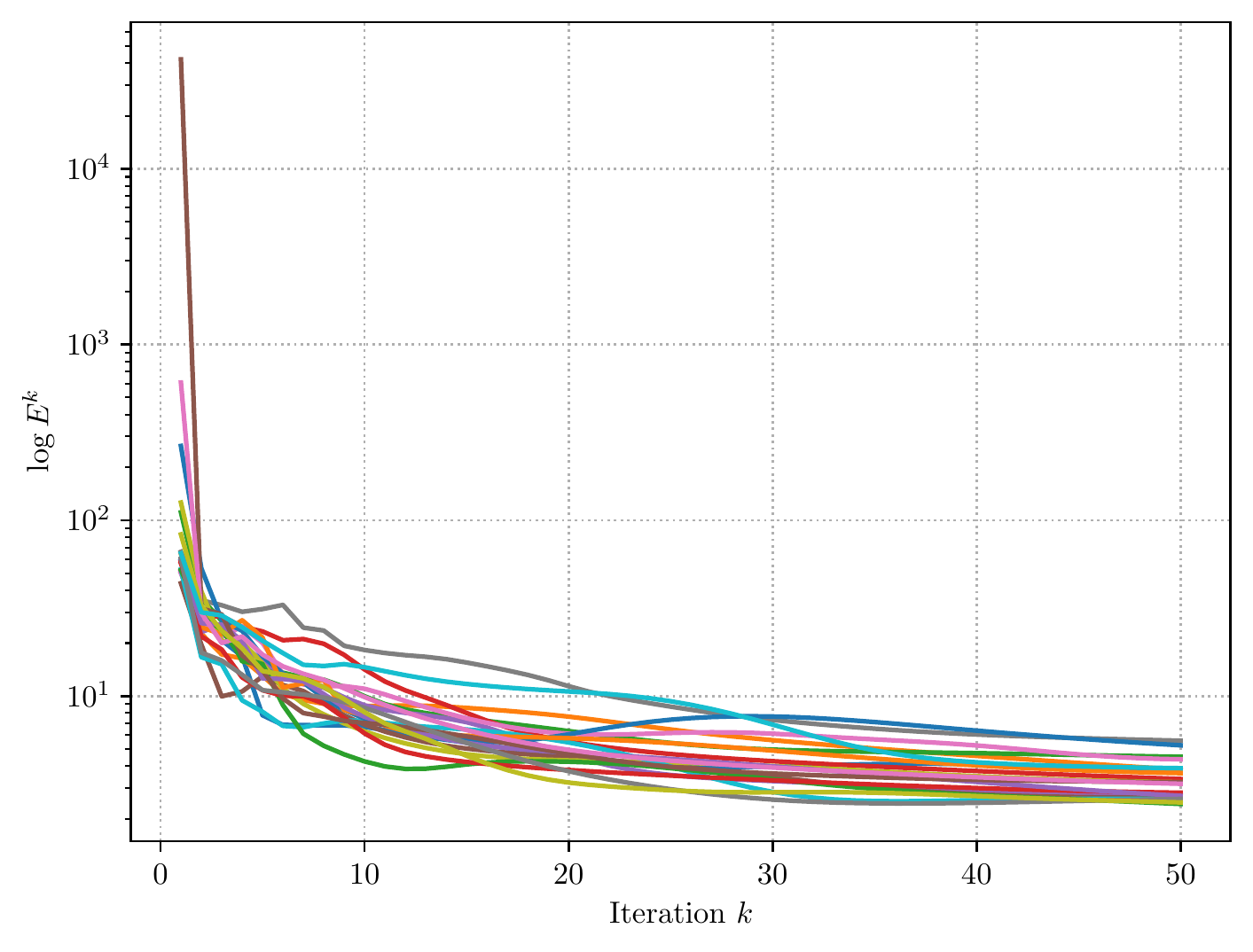}
  \includegraphics[width=.32\linewidth]{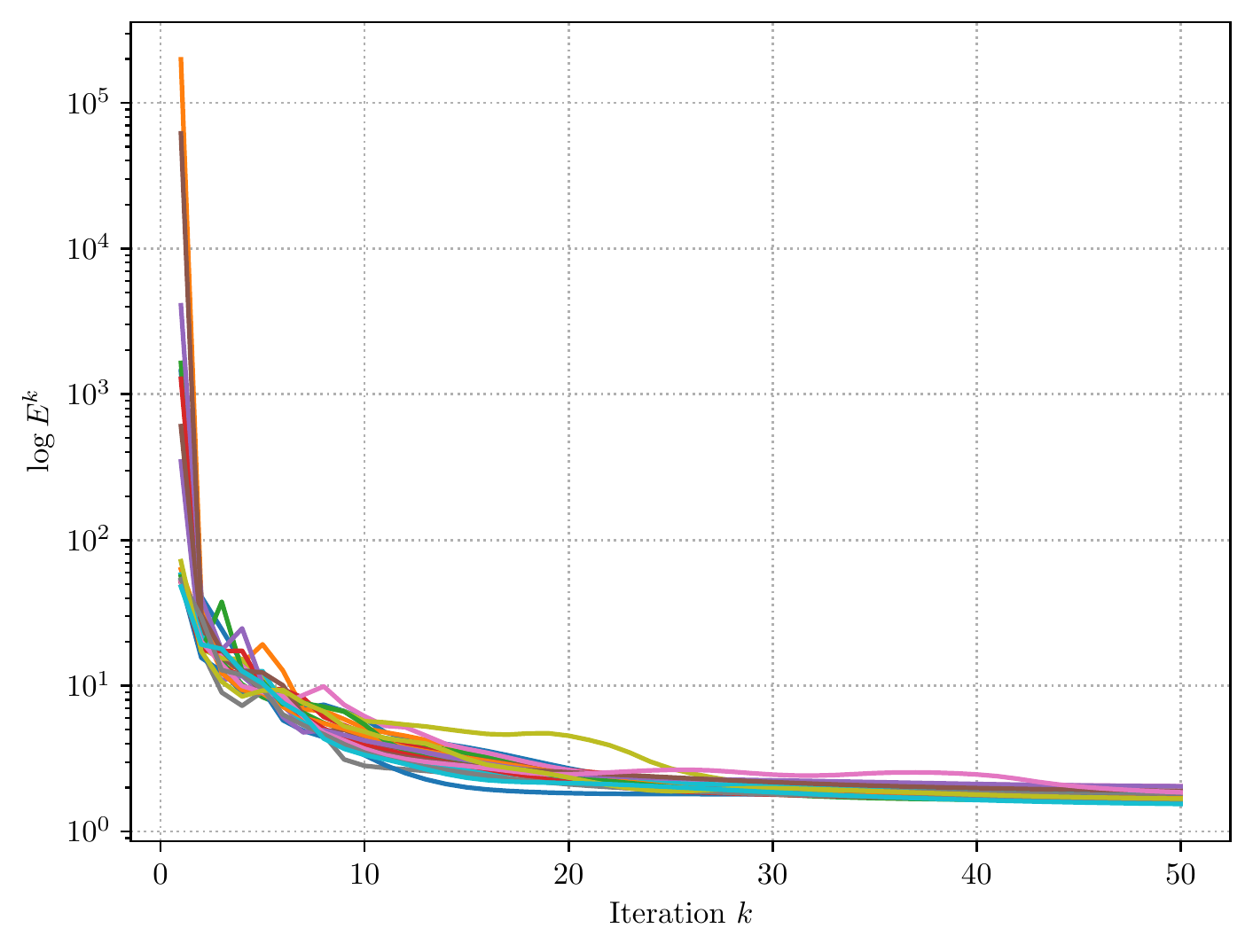}
  \includegraphics[width=.32\linewidth]{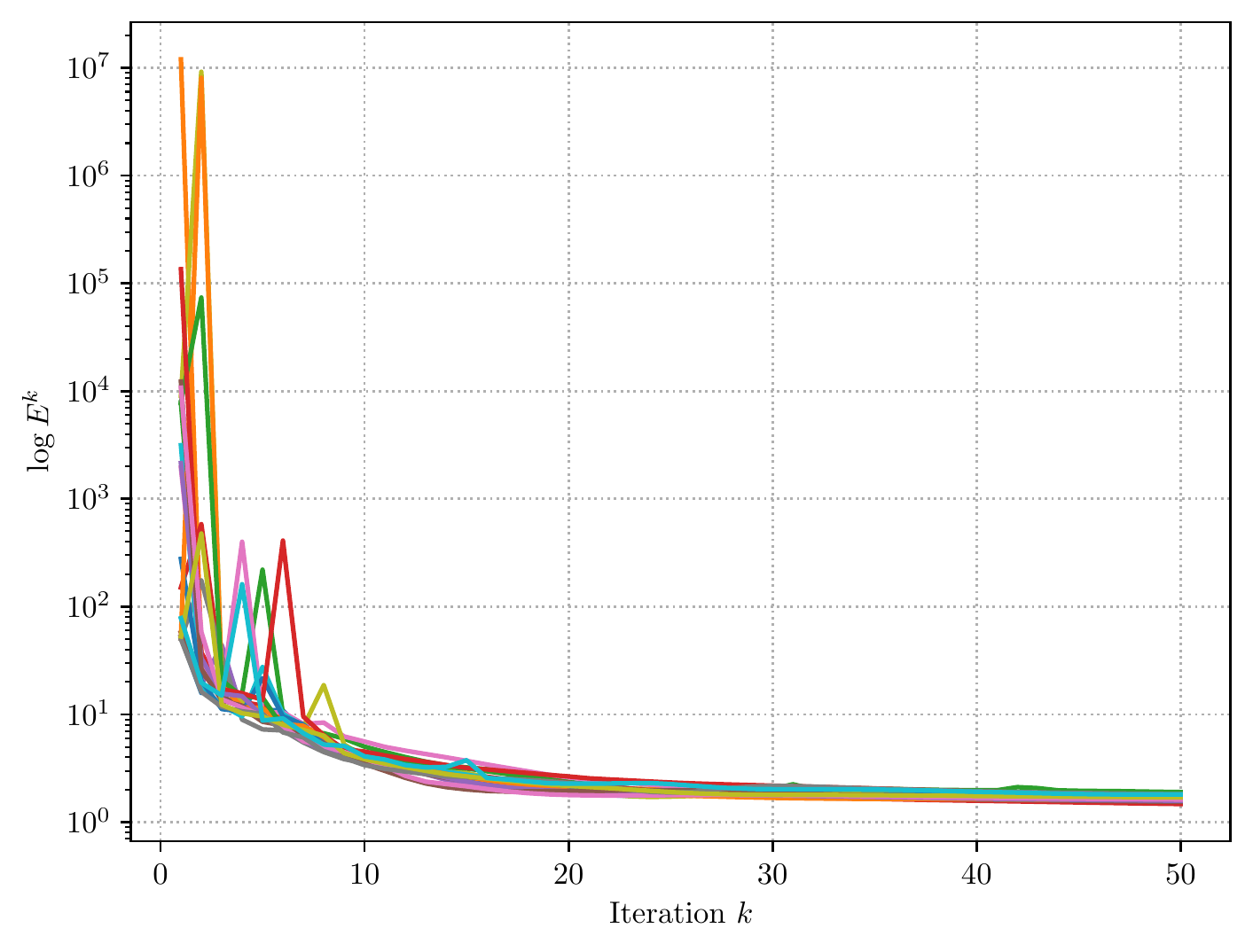}
\caption{Convergence of $E^k$ for $M=150$. Each row corresponds to a different target, with each column representing $N_E=10, 50, 100$. Each figure shows the log data misfit as a function of Kalman iteration for 20 random draws of initial ensemble.}
\label{fig:ex3}
\end{figure}

We first present applications of algorithm \ref{enkfdiffeo} to synthetic template-target pairs such as those presented in figure \ref{fig:templates_and_targets} for the $(M, N_E) \in \{(10, 10), (50, 50), (150, 100)\}$ in figures \ref{fig:L10}, \ref{fig:L50} and \ref{fig:L150}, respectively, at various iterations. Note that again we linearly interpolate between landmarks when plotting these figures. We ran these on a 2014 MacBook Pro with a 2.5GHz Intel Core i7 processor and 16GB of 1600MHz DDR3 RAM. Good matches are obtained in each case. While these figures only show the geodesics for one particular realisation of \eqref{eq:initial_momentum_N01}, we also have evidence that algorithm \ref{enkfdiffeo} shows some robustness with respect to the choice of initial momentum. Indeed, for $M=10$, figure \ref{fig:ex1} shows convergence of $E^k$ as a result of the Kalman iteration ($k$) for three different target configurations (corresponding to each row of figures) and values of $N_E =10, 50, 100$ (corresponding to each column) for 20 different draws of random initial ensembles distributed according to \eqref{eq:initial_momentum_N01}. Figures \ref{fig:ex2} and \ref{fig:ex3} show the same information but $M=50$ and $M=150$, respectively. We observe some similarities and differences in convergence depending on the choice of $M$. Overall we observe rapid convergence in the first 10 or so iterations, with stagnation in the residual $E^k$ obtained around $k\geq 30$. We see an improvement in the smoothness of the convergence as the ensemble size increases (the ensemble can simply span a larger space), most notably seen in the $M=10$ and $M=50$ cases. As we increase the ensemble size for $M=150$, we observe more consistent convergence across realisations of \eqref{eq:initial_momentum_N01} which aligns with our expectations. In particular for $M=150$ we observe some oscillations in the error in the early iterations of the filter. This can be attributed to spurious correlations between distant landmarks - we suggest ways in which this can be abated in section \ref{sec:conclusion}.

\section{Summary \& Outlook}\label{sec:conclusion}

In this paper we have presented a new robust approach to solving shape matching problems using a regularised derivative-free, massively parallel method. We have obtained high accuracy despite the global nature of algorithm \ref{enkfdiffeo}, paving the way for further investigation of Bayesian inversion techniques in the context of shape matching. The enKf we use here is agnostic to the forward model $f$ and can therefore be easily implemented alongside existing software packages or black box third-party implementations.\\

Several extensions present themselves. First, since the enKf is based on covariance matrices we may want to introduce some control over the way landmarks should influence each other. For instance, landmarks that are distant in space should not necessarily have a significant impact on the Kalman gain. The notion of \emph{covariance localisation} in enKf literature \cite{greybush2011balance} provides a useful tool in this case. Localisation means modifying the Kalman matrices by (Fr\"obenius) multiplication by a correlation matrix $L$ whose entries take values in $[0,1]$ using e.g. a loosely defined rule: ${L}_{ij} \approx 1$ if the information represented at $i$ is sufficiently close to affect the information at $j$, and vice versa; otherwise ${L}_{ij} \approx 0$. Total localisation i.e. ${L}_{ij} = \delta_{ij}$ where $\delta$ is the Kronecker delta may be used in the first instance to tune the trade-off between convergence rate and stability. For our applications localisation makes intuitive sense; parts of the shape that are far from each other do not affect each other in the Kalman update.\\

Other extensions are possible. A practical way of improving the accuracy of our matches is to implement a restart in analogy with e.g. the
generalised minimal residual method \cite{saad1986gmres} for linear systems. That is, if consensus has been reached in the ensemble for some tolerance i.e. at a certain iteration $k$, the quantity:
\[
\| P^{k, j} - \bar{P^k} \|_Q,
\]
is below a certain threshold for all $j=1,\ldots,N_E$, we argue that no more information can be extracted from the ensemble. In such a case, a \emph{restart} could be applied where we generate a whole new momentum ensemble generated from draws centered at the previous average momentum $\bar{P^k}$. This will then give a new ensemble with new directions in which to search. While this provides a way of controlling the information coming from the momentum, an adaptive regularisation strategy could also be investigated i.e. $\xi$ is a function of $k$.\\  

Finally, since we are dealing with a discretised version of the forward problem future work includes quantifying the error in the Bayesian inverse problem via this discretisation error drawing inspiration from the work from \cite{cotter2010approximation}. Future work also includes applying algorithm \ref{enkfdiffeo} to real data and rigorously treating the Bayesian inversion problem. Filters such as the unscented Kalman filter \cite{van2001unscented} or machine learning approaches could be explored.

\appendix
\section{Code}\label{appendix:a}

All the source code used to run the simulations presented in this paper is available from this repository \url{github.com/andreasbock/enkf\_landmarks}. Consult the \texttt{README.md} for details on how to reproduce the experiments presented here.
\clearpage
\printbibliography

@book{vogel2002computational,
  title={Computational methods for inverse problems},
  author={Vogel, Curtis R},
  year={2002},
  publisher={SIAM}
}

@book{reich2015probabilistic,
  title={Probabilistic forecasting and Bayesian data assimilation},
  author={Reich, Sebastian and Cotter, Colin},
  year={2015},
  publisher={Cambridge University Press}
}

@inproceedings{van2001unscented,
  title={The unscented particle filter},
  author={Van Der Merwe, Rudolph and Doucet, Arnaud and De Freitas, Nando and Wan, Eric A},
  booktitle={Advances in neural information processing systems},
  pages={584--590},
  year={2001}
}

@article{charlier2020kernel,
  title={Kernel operations on the {GPU}, with autodiff, without memory overflows},
  author={Charlier, Benjamin and Feydy, Jean and Glaun{\`e}s, Joan Alexis and Collin, Fran{\c{c}}ois-David and Durif, Ghislain},
  journal={arXiv preprint arXiv:2004.11127},
  year={2020}
}

@inproceedings{paszke2019pytorch,
  title={Pytorch: An imperative style, high-performance deep learning library},
  author={Paszke, Adam and Gross, Sam and Massa, Francisco and Lerer, Adam and Bradbury, James and Chanan, Gregory and Killeen, Trevor and Lin, Zeming and Gimelshein, Natalia and Antiga, Luca and others},
  booktitle={Advances in neural information processing systems},
  pages={8026--8037},
  year={2019}
}

@article{dupuis1998variational,
    title = {Variational problems on flows of diffeomorphisms for image matching},
    author = {Dupuis, Paul and Grenander, Ulf and Miller, Michael I},
    year = 1998,
    journal = {{Quarterly of Applied Mathematics}},
    publisher = {JSTOR},
    pages = {587--600}
}

@article{grenander1994representations,
    title = {Representations of knowledge in complex systems},
    author = {Grenander, Ulf and Miller, Michael I},
    year = 1994,
    journal = {Journal of the Royal Statistical Society. Series B (Methodological)},
    publisher = {JSTOR},
    pages = {549--603}
}

@article{grenander1998computational,
    title = {Computational anatomy: An emerging discipline},
    author = {Grenander, Ulf and Miller, Michael I},
    year = 1998,
    journal = {Quarterly of applied mathematics},
    volume = 56,
    number = 4,
    pages = {617--694}
}

@article{trouve1998diffeomorphisms,
    title = {Diffeomorphisms groups and pattern matching in image analysis},
    author = {Trouv{\'e}, Alain},
    year = 1998,
    journal = {International Journal of Computer Vision},
    publisher = {Springer},
    volume = 28,
    number = 3,
    pages = {213--221}
}

@article{trouve1995infinite,
    title = {An infinite dimensional group approach for physics based models in pattern recognition},
    author = {Trouv{\'e}, Alain},
    year = 1995,
    journal = {{Preprint}}
}

@article{beg2005computing,
    title = {Computing large deformation metric mappings via geodesic flows of diffeomorphisms},
    author = {Beg, M Faisal and Miller, Michael I and Trouv{\'e}, Alain and Younes, Laurent},
    year = 2005,
    journal = {{International Journal of Computer Vision}},
    publisher = {Springer},
    volume = 61,
    number = 2,
    pages = {139--157}
}

@article{vialard2012diffeomorphic,
    title = {Diffeomorphic 3D image registration via geodesic shooting using an efficient adjoint calculation},
    author = {Vialard, Fran{\c{c}}ois Xavier and Risser, Laurent and Rueckert, Daniel and Cotter, Colin J},
    year = 2012,
    journal = {International Journal of Computer Vision},
    publisher = {Springer},
    volume = 97,
    number = 2,
    pages = {229--241}
}

@article{holm2009euler,
    title = {{The Euler-Poincar{\'e} theory of metamorphosis}},
    author = {Holm, Darryl D and Trouv{\'e}, Alain and Younes, Laurent},
    year = 2009,
    journal = {Quarterly of Applied Mathematics},
    publisher = {JSTOR},
    pages = {661--685}
}

@article{trouve2005metamorphoses,
    title = {{Metamorphoses through Lie group action}},
    author = {Trouv{\'e}, Alain and Younes, Laurent},
    year = 2005,
    journal = {Foundations of Computational Mathematics},
    publisher = {Springer},
    volume = 5,
    number = 2,
    pages = {173--198}
}

@article{holm1998euler,
    title = {The Euler--Poincar{\'e} equations and semidirect products with applications to continuum theories},
    author = {Holm, Darryl D and Marsden, Jerrold E and Ratiu, Tudor S},
    year = 1998,
    journal = {Advances in Mathematics},
    publisher = {Elsevier},
    volume = 137,
    number = 1,
    pages = {1--81}
}

@book{younesshapes,
    title = {Shapes and diffeomorphisms},
    author = {Younes, Laurent},
    year = 2010,
    publisher = {Springer Science \& Business Media},
    volume = 171
}

@article{mumford2002pattern,
    title = {Pattern theory: the mathematics of perception},
    author = {Mumford, David},
    year = 2002,
    journal = {arXiv preprint math/0212400}
}

@article{younes2007jacobi,
    title = {Jacobi fields in groups of diffeomorphisms and applications},
    author = {Younes, Laurent},
    year = 2007,
    journal = {Quarterly of applied mathematics},
    publisher = {JSTOR},
    pages = {113--134}
}

@article{dashti2017bayesian,
    title = {The {B}ayesian approach to inverse problems},
    author = {Dashti, Masoumeh and Stuart, Andrew M},
    year = 2017,
    journal = {Handbook of Uncertainty Quantification},
    publisher = {Springer},
    pages = {311--428}
}

@article{kuhnel2017differential,
    title = {Differential geometry and stochastic dynamics with deep learning numerics},
    author = {K{\"u}hnel, Line and Arnaudon, Alexis and Sommer, Stefan},
    year = 2017,
    journal = {arXiv preprint arXiv:1712.08364}
}

@article{cotter2010approximation,
    title = {Approximation of {B}ayesian inverse problems for {PDE}s},
    author = {Cotter, Simon L and Dashti, Massoumeh and Stuart, Andrew M},
    year = 2010,
    journal = {SIAM Journal on Numerical Analysis},
    publisher = {SIAM},
    volume = 48,
    number = 1,
    pages = {322--345}
}

@article{cotter2013bayesian,
    title = {Bayesian data assimilation in shape registration},
    author = {Cotter, Colin John and Cotter, Simon L and Vialard, Fran{\c{c}}ois Xavier},
    year = 2013,
    journal = {Inverse Problems},
    publisher = {IOP Publishing},
    volume = 29,
    number = 4,
    pages = 45011
}

@article{iglesias2016regularizing,
    title = {A regularizing iterative ensemble Kalman method for {PDE}-constrained inverse problems},
    author = {Iglesias, Marco A},
    year = 2016,
    journal = {Inverse Problems},
    publisher = {IOP Publishing},
    volume = 32,
    number = 2,
    pages = 25002
}

@article{evensen1994sequential,
    title = {Sequential data assimilation with a nonlinear quasi-geostrophic model using {M}onte {C}arlo methods to forecast error statistics},
    author = {Evensen, Geir},
    year = 1994,
    journal = {Journal of Geophysical Research: Oceans},
    publisher = {Wiley Online Library},
    volume = 99,
    number = {C5},
    pages = {10143--10162}
}

@article{iglesias2013ensemble,
    title = {Ensemble Kalman methods for inverse problems},
    author = {Iglesias, Marco A and Law, Kody JH and Stuart, Andrew M},
    year = 2013,
    journal = {Inverse Problems},
    publisher = {IOP Publishing},
    volume = 29,
    number = 4,
    pages = 45001
}

@article{schneider2017earth,
    title = {Earth system modeling 2.0: A blueprint for models that learn from observations and targeted high-resolution simulations},
    author = {Schneider, Tapio and Lan, Shiwei and Stuart, Andrew and Teixeira, Jo{\~a}o},
    year = 2017,
    journal = {Geophysical Research Letters},
    publisher = {Wiley Online Library},
    volume = 44,
    number = 24,
    pages = {12--396}
}

@article{miller2006geodesic,
    title = {Geodesic shooting for {C}omputational {A}natomy},
    author = {Miller, Michael I and Trouv{\'e}, Alain and Younes, Laurent},
    year = 2006,
    journal = {{Journal of Mathematical Imaging and Vision}},
    publisher = {Springer},
    volume = 24,
    number = 2,
    pages = {209--228}
}

@article{greybush2011balance,
    title = {Balance and ensemble {K}alman filter localization techniques},
    author = {Greybush, Steven J and Kalnay, Eugenia and Miyoshi, Takemasa and Ide, Kayo and Hunt, Brian R},
    year = 2011,
    journal = {Monthly Weather Review},
    volume = 139,
    number = 2,
    pages = {511--522}
}

@inproceedings{li2007iterative,
    title = {An iterative ensemble {K}alman filter for data assimilation},
    author = {Li, Gaoming and Reynolds, Albert Coburn and others},
    year = 2007,
    booktitle = {SPE annual technical conference and exhibition},
    organization = {Society of Petroleum Engineers}
}

@article{stuart2010inverse,
    title = {Inverse problems: a {B}ayesian perspective},
    author = {Stuart, Andrew M},
    year = 2010,
    journal = {Acta numerica},
    publisher = {Cambridge University Press},
    volume = 19,
    pages = {451--559}
}

@article{younes2009evolutions,
    title = {Evolutions equations in computational anatomy},
    author = {Younes, Laurent and Arrate, Felipe and Miller, Michael I},
    year = 2009,
    journal = {NeuroImage},
    publisher = {Elsevier},
    volume = 45,
    number = 1,
    pages = {S40--S50}
}

@article{ma2008bayesian,
    title = {Bayesian template estimation in computational anatomy},
    author = {Ma, Jun and Miller, Michael I and Trouv{\'e}, Alain and Younes, Laurent},
    year = 2008,
    journal = {NeuroImage},
    publisher = {Elsevier},
    volume = 42,
    number = 1,
    pages = {252--261}
}

@article{ma2010bayesian,
    title = {A Bayesian generative model for surface template estimation},
    author = {Ma, Jun and Miller, Michael I and Younes, Laurent},
    year = 2010,
    journal = {International journal of biomedical imaging},
    publisher = {Hindawi},
    volume = 2010
}

@article{saad1986gmres,
    title = {GMRES: A generalized minimal residual algorithm for solving nonsymmetric linear systems},
    author = {Saad, Youcef and Schultz, Martin H},
    year = 1986,
    journal = {SIAM Journal on scientific and statistical computing},
    publisher = {SIAM},
    volume = 7,
    number = 3,
    pages = {856--869}
}

@article{kalman1960new,
    title = {A new approach to linear filtering and prediction problems},
    author = {Kalman, Rudolph Emil},
    year = 1960,
    journal = {Journal of basic Engineering},
    publisher = {American Society of Mechanical Engineers},
    volume = 82,
    number = 1,
    pages = {35--45}
}

@article{mandel2011convergence,
    title = {On the convergence of the ensemble Kalman filter},
    author = {Mandel, Jan and Cobb, Loren and Beezley, Jonathan D},
    year = 2011,
    journal = {Applications of Mathematics},
    publisher = {Springer},
    volume = 56,
    number = 6,
    pages = {533--541}
}

\end{document}